\documentclass{article}

\PassOptionsToPackage{numbers,compress}{natbib}

\usepackage[main, final]{neurips_2026}

\usepackage[utf8]{inputenc} %
\usepackage[T1]{fontenc}    %
\usepackage{url}            %
\usepackage{booktabs}       %
\usepackage{amsfonts}       %
\usepackage{amsmath}        %
\usepackage{nicefrac}       %
\usepackage{microtype}      %
\usepackage{xcolor}         %
\usepackage{graphicx}
\usepackage{multirow}
\usepackage{enumitem}       %
\usepackage{makecell}       %
\usepackage{tabularx}       %
\usepackage{xltabular}      %
\usepackage{wrapfig}        %

\usepackage{comment}
\usepackage[skip=0pt,font=small]{caption}
\newcommand{\wahib}[1]{#1}

\newcommand{\chen}[1]{#1}

\newcommand{\method}{\texttt{SGMA}}

\newcolumntype{Y}{>{\raggedright\arraybackslash}X}

\usepackage{subcaption}
\usepackage[pagebackref,breaklinks,colorlinks]{hyperref}
\usepackage{cleveref}       %

\title{Structure-Guided Masked Autoencoders for Ultra-High Resolution Scientific Image Understanding}

\author{%
  Enzhi Zhang$^{1}$ \quad Du Wu$^{2}$ \quad Rui Zhong$^{1}$ \quad Cong Ma$^{1}$ \quad Isaac Lyngaas$^{3}$ \quad Amir Koushyar Ziabari$^{3}$ \\
  \textbf{Xiao Wang$^{3}$ \quad Peng Chen$^{2}$ \quad Tao Luo$^{4}$ \quad Toshio Endo$^{5}$ \quad Fumiyoshi Shoji$^{2}$ \quad Kento Sato$^{2}$} \\
  \textbf{Kentaro Uesugi$^{6}$ \quad Takayuki Nonoyama$^{1}$ \quad Ryuji Kiyama$^{1}$ \quad Masahiro Yoshida$^{1}$ \quad Masaru Tezuka$^{1}$} \\
  \textbf{Tetsuya Ishikawa$^{7}$ \quad Satoshi Matsuoka$^{2}$ \quad Masaharu Munetomo$^{1}$ \quad Mohamed Wahib$^{2}$} \\[0.6em]
  \small $^{1}$Hokkaido University, Japan \quad
  $^{2}$RIKEN Center for Computational Science (R-CCS), Japan \\
  \small $^{3}$Oak Ridge National Laboratory, USA \quad
  $^{4}$A*STAR, Singapore \\
  \small $^{5}$Institute of Science Tokyo, Japan \quad
  $^{6}$Japan Synchrotron Radiation Research Institute (JASRI), Japan \\
  \small $^{7}$RIKEN SPring-8 Center, Japan
}

\begin{document}
\maketitle

\begin{abstract}

Self-supervised pre-training with Vision Transformers, including Masked Autoencoders (MAE), is difficult to apply to gigapixel scientific images. Random masking is poorly matched to the structured, multi-scale morphology of scientific data, while uniform tokenization produces prohibitively long sequences that make $O(N^2)$ attention impractical. We propose \method{}, a structure-guided masked autoencoding framework for ultra-high-resolution scientific images. \method{} couples two components: a content-adaptive quadtree tokenizer that compresses gigapixel images into a fixed-length sequence, and a structure-conditioned masking process that biases reconstruction toward spatially informative regions. To stabilize this process across scales, we introduce Damped Accumulation (DA), which aggregates signal-dependent responses across the tree into a structure canvas used to guide masking. The resulting pre-training task preserves fine microstructure while remaining compatible with standard ViT encoders and MAE-style reconstruction. Across electron microscopy, whole-slide optical microscopy, and X-ray CT datasets, \method{} consistently outperforms MAE baselines. It achieves 95.68\% Dice on the 8K${\times}$8K${\times}$28K SpringXCT dataset, improving over the same-architecture MAE baseline by +13.00 points, and 83.21\% Dice on the $32\text{K}^2$ WSI PAIP dataset, improving by +16.84 points, while providing up to a $24.8\times$ inference speedup.

\end{abstract}
    
\section{Introduction}
\label{sec:intro}

Benefiting from the immense success of large-scale pre-training paradigms, data-driven scientific discovery, particularly in the analysis of scientific image data, is achieving unprecedented depth and breadth. Ultra-High Resolution (UHR) images, such as those from high-throughput \wahib{optical and electron} microscopy \cite{Kiyama2022Nanoscale, noguchi2024real}, pathological scans ~\cite{KIM2021101854}, or X-ray Computed Tomography (XCT) \cite{cnudde2013high, shi2020application}, are being acquired at the TB/PB scale. These images contain exceptionally rich, fine-grained, multi-scale information and hierarchical structures. However, \wahib{directly using advanced models such as the Vision Transformer (ViT) to conduct vision analytics on such UHR images faces two major bottlenecks.}

First is the severe scarcity of labels. Although the acquisition of UHR images has become increasingly accessible, their annotation costs are prohibitively high. In terms of total pixels, annotating a single $64K^2$ resolution image is equivalent to annotating 16,384 images at 512 resolution ($512\times512$). More importantly, the precise annotation of scientific images (e.g. cell contours, material defects, or geographical features) relies heavily on time-consuming and costly domain expertise.
Second is the intractable computational complexity. The self-attention mechanism, the core of the Transformer model, exhibits quadratic computational and memory complexity with respect to the input sequence length $N$ (i.e., the number of \wahib{image} patches). For UHR images, this results in an astronomical sequence \wahib{size} (e.g. \chen{a $64K^2$ image with 16-pixel ($4\times4$) patches \wahib{yields} $16,384^2$ tokens}), making standard ViT training, fine-tuning, and even inference \wahib{intractable}.

\begin{figure*}[t]
\includegraphics[width=\textwidth]{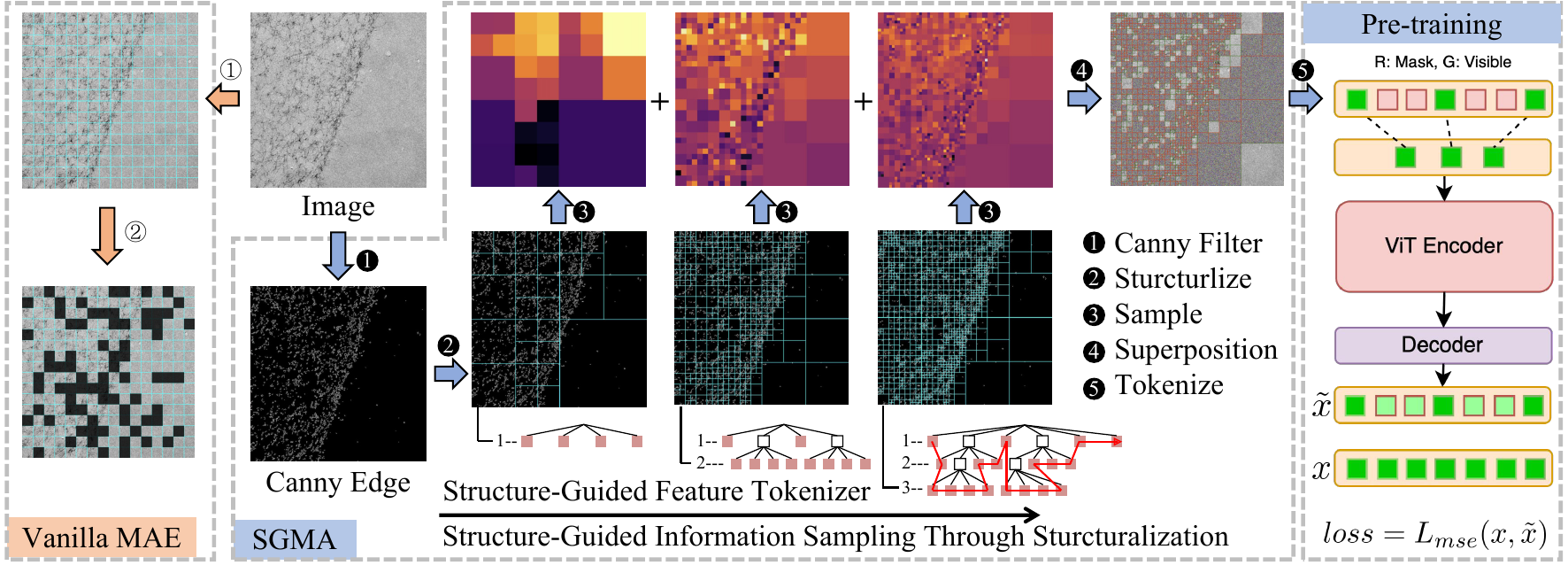}
    \caption{
    Overview of \method{}. A structure-guided quadtree decomposes the image and accumulates per-level spatial responses by superimposing scale-dependent Gaussian noise. The resulting structure-conditioned noise field determines where masks are placed on the image. When applied to fine-tuning SAM~\cite{kirillov2023segment} on HydrogelTEM-1K electron microscopy data, \method{} compresses the sequence length by a large margin while maintaining key spatial structures, enabling stronger fine-tuning performance.}
    \label{fig:hdme_overview}
\end{figure*}

To address the label scarcity problem, \wahib{unsupervised} pre-training methods like Masked Autoencoders (MAE) \cite{he2022masked}  are becoming increasingly used in vision models \cite{ravi2024sam2, kirillov2023segment}. MAE \wahib{enables models to learn} robust visual representations by reconstructing highly masked patches, leveraging information redundancy and contextual relationships between patches. However, \wahib{MAE faces challenges \cite{chen2023scaling, liu2021swin} with scientific UHR images}. \wahib{More specifically, MAE is not ideal for UHR scientific images because they do not incorporate knowledge of the hierarchical spatial microstructures present in such data. Further, downsampling the UHR image to fit in memory (e.g. to 256x256) destroys all fine-grained, micro-level details, which is often the most critical information.} \wahib{In addition}, even if a model is successfully pre-trained on UHR data, it must still process the complete, ultra-long patch sequence during fine-tuning or inference for downstream tasks (e.g. semantic segmentation). The quadratic complexity bottleneck persists.

Existing long-sequence solutions, such as hierarchical ViTs (e.g. Swin Transformer \cite{cao2022swin} or linear attention approximations \cite{pmlr-v162-dao22a, bo2023specformer, NEURIPS2021_b4fd1d2c}), while aimed at reducing computational load, introduce \wahib{complex model architectures and inter-layer load imbalance} (\wahib{such as} overlapped sliding windows) or \wahib{introduce custom models} with fundamental changes to the attention calculation. These modifications are often incompatible with pre-training paradigm of MAE, which relies on global, sparse patch reconstruction.
Therefore, a critical open question remains: \emph{can we design a framework that not only enables efficient pre-training on unlabeled UHR images \wahib{of different types of micro-structures}, yet simultaneously ensures that downstream task fine-tuning and deployment are computationally feasible?}

Inspired by the concept of spatial token compression using mixed-scale patches~\cite{wei2025deepseek, wang2025orbit, zhang2024adaptive, zhang2025shf}, we propose a novel pre-training framework, \method{} (\textit{Structure-Guided Masked AutoEncoder}, pronounced \emph{Sigma}). \wahib{First, we capture the spatial layout of fine micro-structures by applying a probabilistic, content-adaptive quadtree that recursively splits the image and extracts multi-level spatial information.
Next, we design a structural masking strategy that aligns the mask distribution with the micro-structural patterns and the multi-scale positions discovered by the quadtree. To generate this structure-guided mask, we model the sampling process as the iterative addition (superposition) of Gaussian noise across tree levels.} We then train an MAE-\wahib{based Vision Transformer} encoder to perform reconstruction. This design forces the model to learn to discern and reconstruct key structures and multi-scale representations in UHR images during the pre-training phase. Concurrently, the framework allows for efficient inference in downstream tasks. \wahib{Most importantly, the framework \method{} approach generalizes to images of very different micro-structures and resolutions: Transmission Electron Microscopy (TEM), optical WSI microscopy, and X-ray CT images at $<0.1\,\mathrm{nm}$, $12\,\mu\mathrm{m}$, and $200\,\mathrm{nm}$ image resolutions, respectively.
}
Our contributions are summarized as follows:
{
\setlength{\leftmargini}{15 pt}
\begin{itemize}
\item \emph{Structure-Guided Masked AutoEncoder}. A novel pre-training framework for unlabeled ultra-high resolution (UHR) images. \method{} learns representations through a task that jointly performs Structure-Guided Feature Tokenizer (SGFT) tokenization and Damped Accumulation (DA) structure-conditioned sampling. \wahib{This improves the performance of standard MAE since the model is geared to learn to reconstruct spatial regions with concentration of details}, while \wahib{mitigating} the critical downstream $O(N^2)$ computational bottleneck (in fine-tuning and inference) that standard MAE-trained models fail to address, enabling efficient end-to-end deployment.
    
\item \emph{Ultra-High Resolution Segmentation}. We conduct extensive segmentation experiments on three multi-scale UHR datasets (TEM: \emph{HydrogelTEM-1K}, X-ray CT: \emph{SpringXCT-8K}, and WSI microscopy: \emph{PAIP-32K}). Results show that \method{} significantly outperforms MAE-pretrain baselines in both accuracy and speed, achieving, for instance, a +16.84 Dice improvement and a $24.8\times$ inference speedup on the 32K PAIP dataset.
    
\item \emph{Downstream Task Analyze}. We validate the real-world efficacy of our method by applying our high-precision segmentation to a downstream scientific task: quantitative analysis of the pore network structure from the zero-shot SpringXCT 3D dataset and high-precision skeleton from HydrogelTEM dataset, showcasing its practical value in scientific discovery workflows.
\end{itemize}
}

\begin{figure*}
    \captionsetup[subfigure]{labelformat=empty}

    \begin{minipage}[c]{0.05\textwidth}
        \centering
        \rotatebox{90}{\shortstack[c]{\textbf{Electron Microscope}\\\footnotesize($<\!0.1$\,nm: atomic scale)}}
    \end{minipage}\hspace{4pt}%
    \begin{minipage}[c]{0.93\textwidth}
        \begin{subfigure}[b]{0.24\linewidth} 
            \caption{$1024^2@HydrogelTEM$}
            \includegraphics[width=\textwidth]{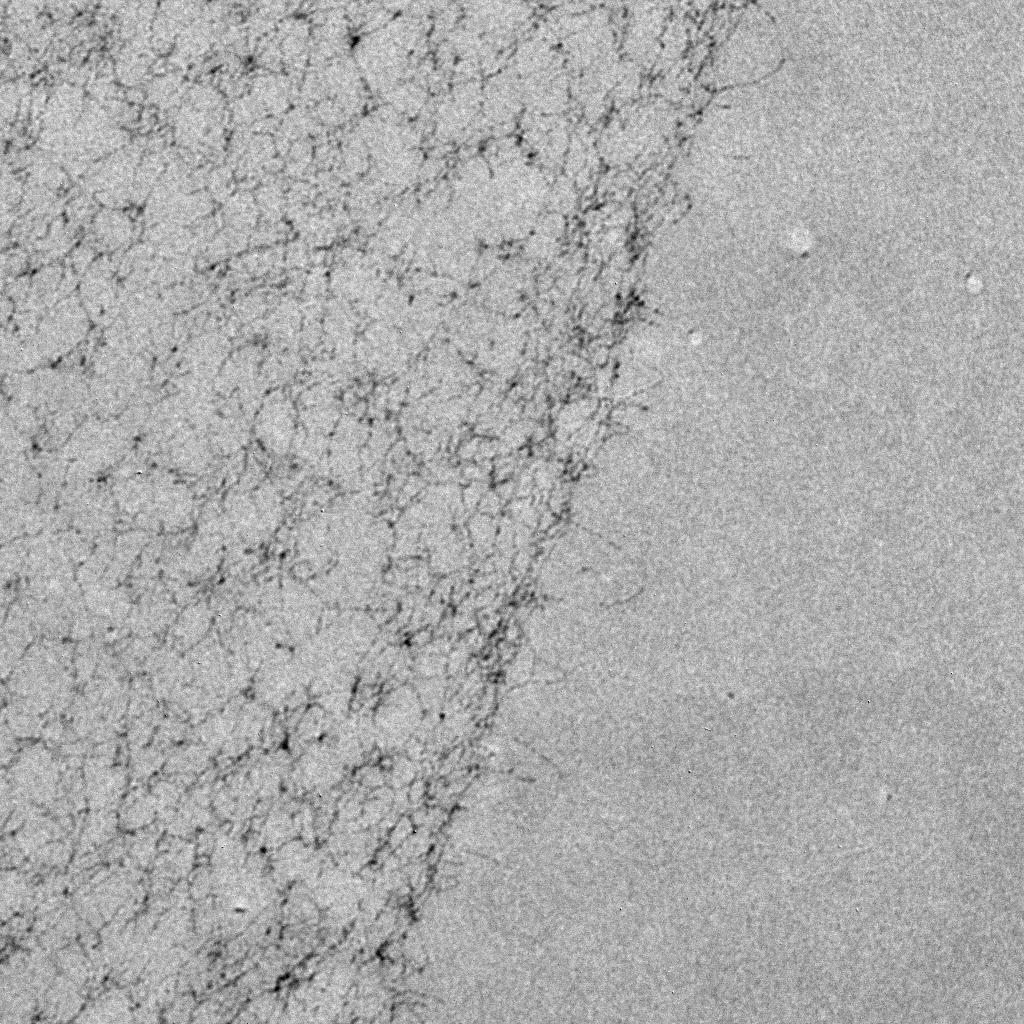}
        \end{subfigure}
        \hfill
        \begin{subfigure}[b]{0.24\linewidth}
            \caption{Seq. Len.: 1024}
            \includegraphics[width=\textwidth]{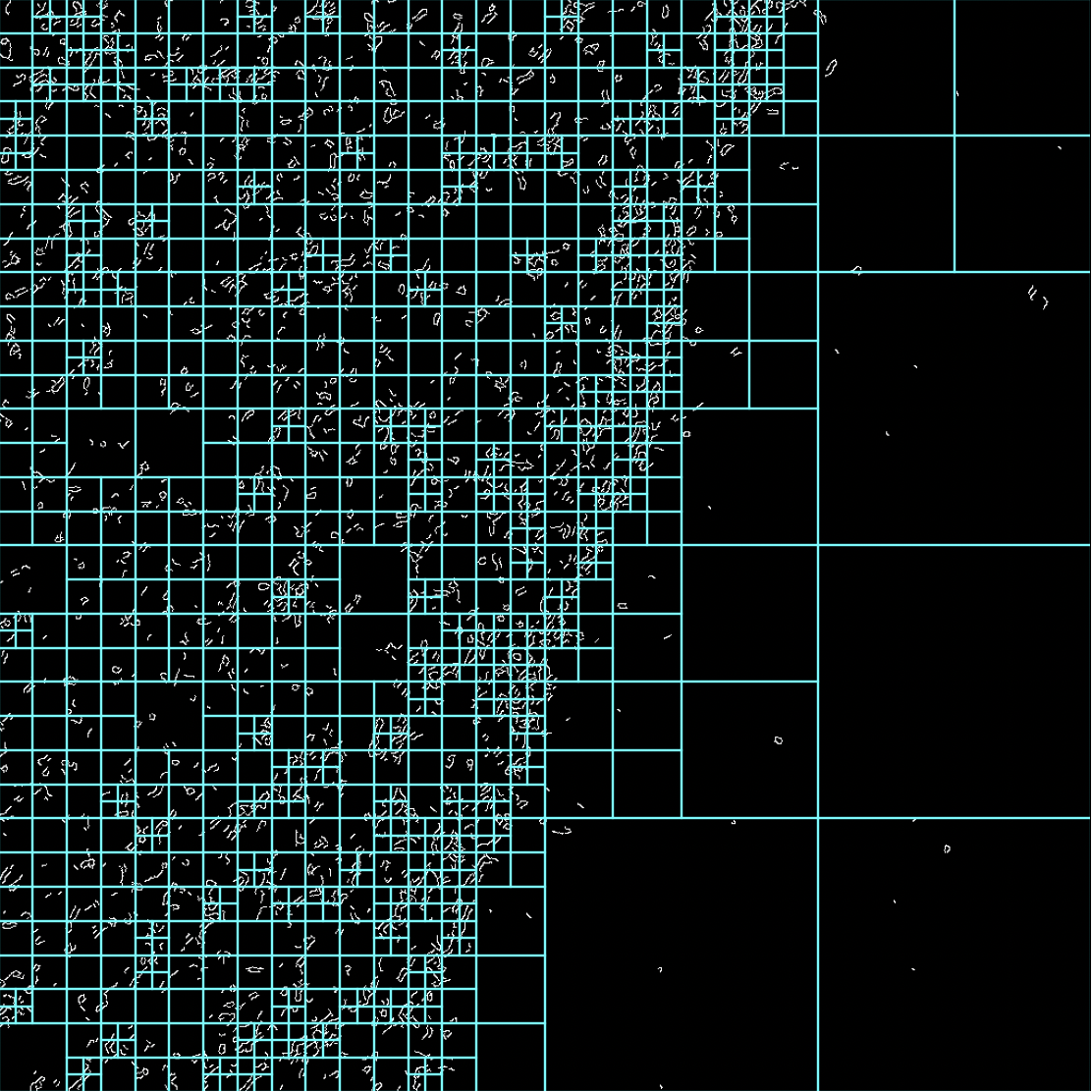}
        \end{subfigure}
        \hfill
        \begin{subfigure}[b]{0.24\linewidth}
            \caption{Hierarchy Depth: 9}
            \includegraphics[width=\textwidth]{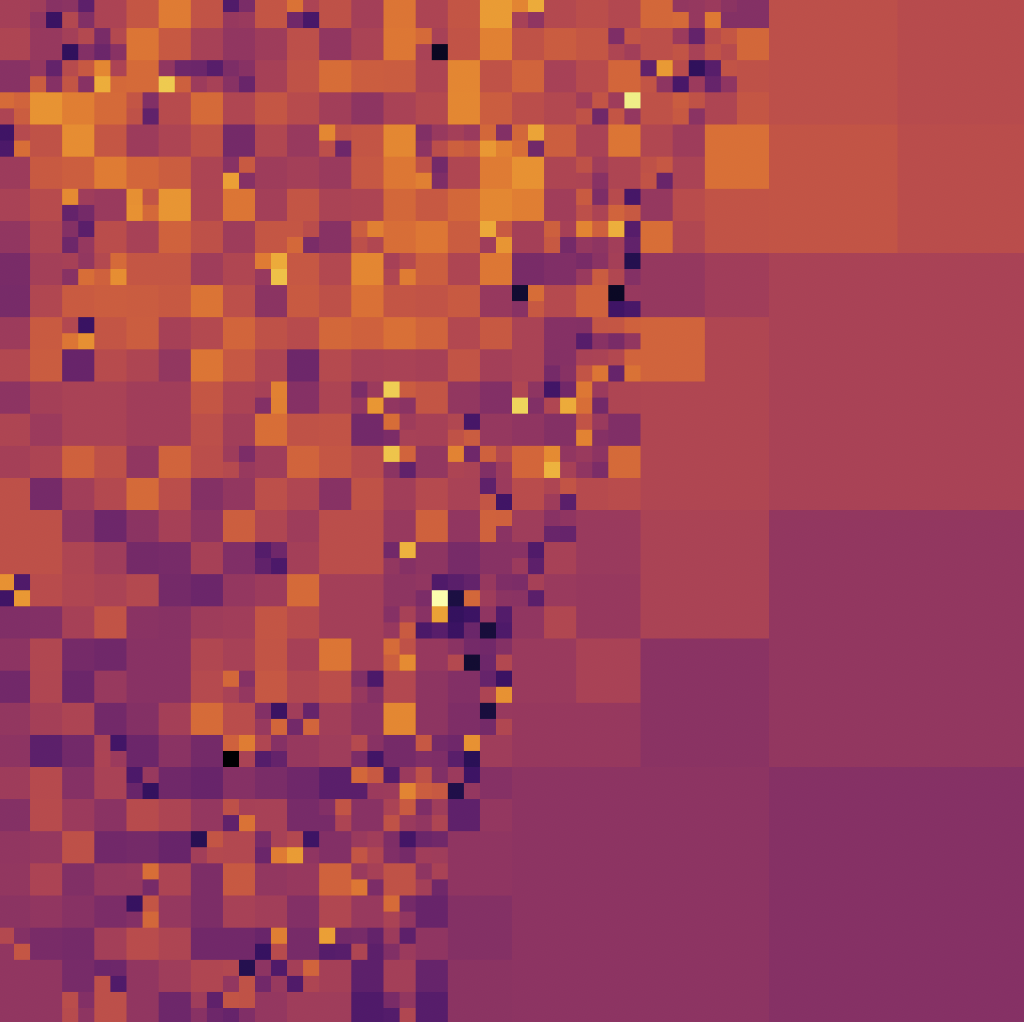}
        \end{subfigure}
        \hfill
        \begin{subfigure}[b]{0.24\linewidth}
            \caption{Input:Clean=G. Masked=R}
            \includegraphics[width=\textwidth]{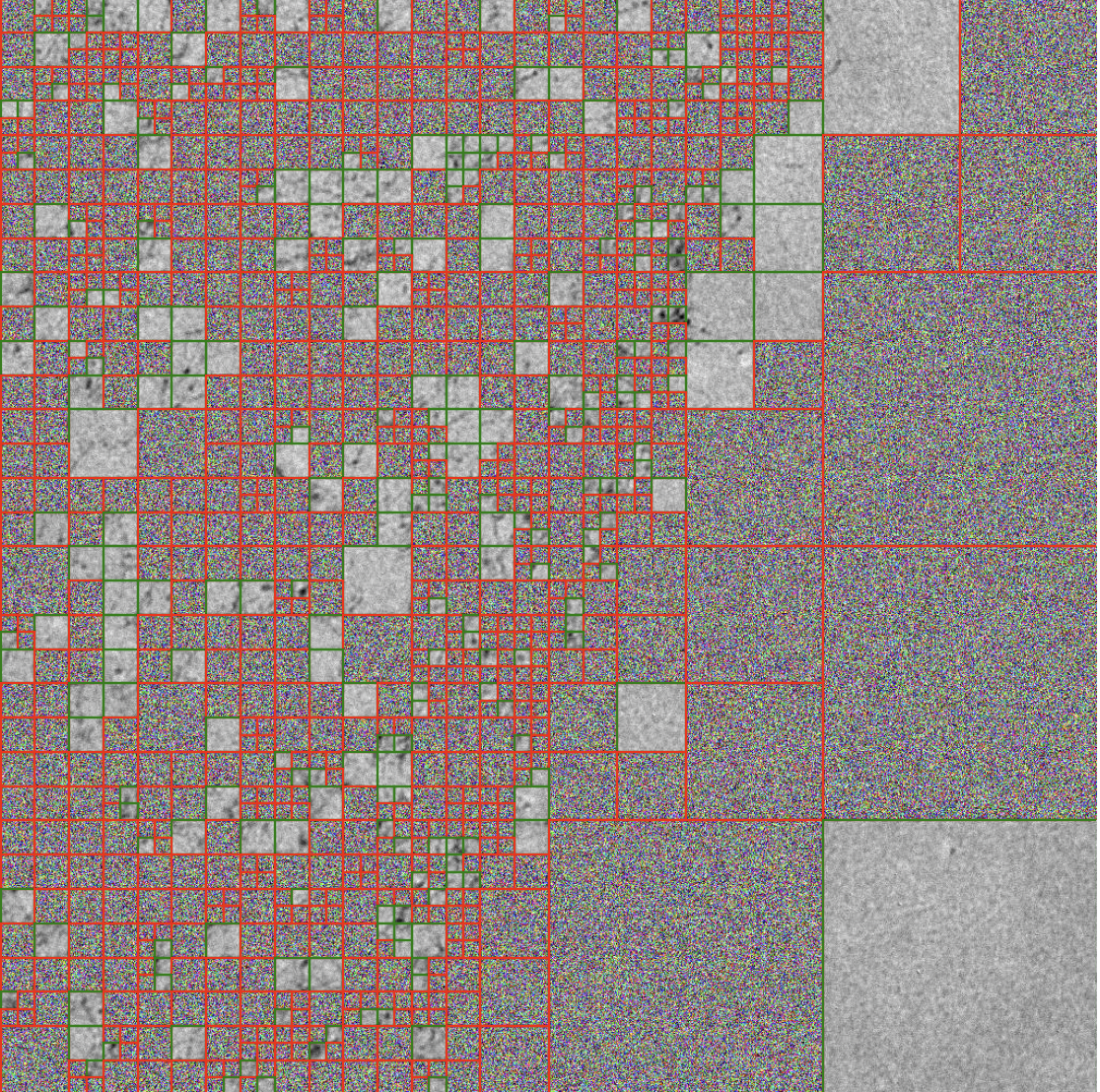}
        \end{subfigure}
    \end{minipage}

    \vfill %

    \begin{minipage}[c]{0.05\textwidth}
        \centering
        \rotatebox{90}{\shortstack[c]{\textbf{X-ray CT}\\\footnotesize(12\,$\mu$m; micro scale)}} %
    \end{minipage}\hspace{4pt}%
    \begin{minipage}[c]{0.93\textwidth}
        \begin{subfigure}[b]{0.24\linewidth}
            \caption{$8192^2@SpringXCT$}
            \includegraphics[width=\textwidth]{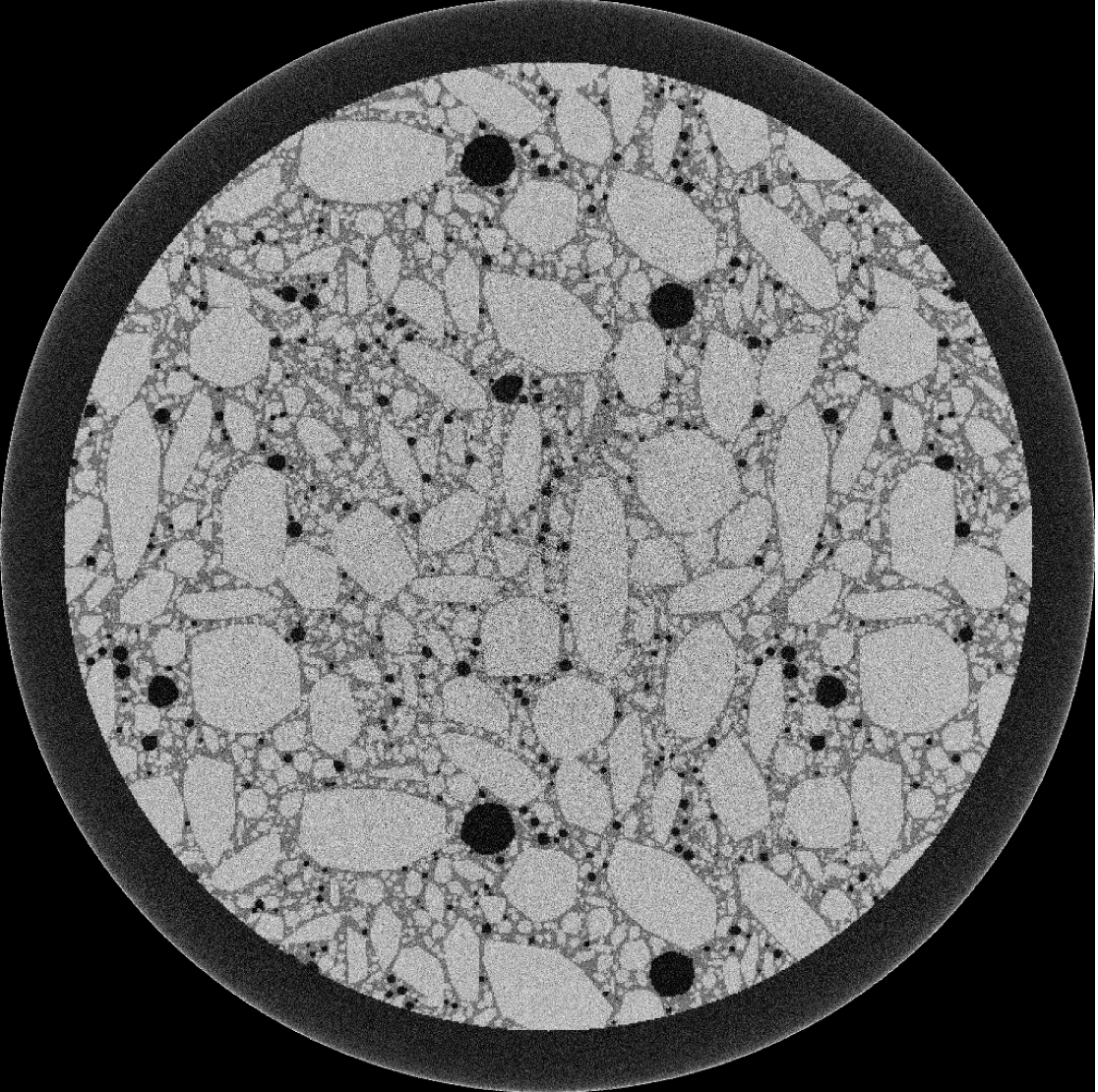}
        \end{subfigure}
        \hfill
        \begin{subfigure}[b]{0.24\linewidth}
            \caption{Seq. Len.: 8194}
            \includegraphics[width=\textwidth]{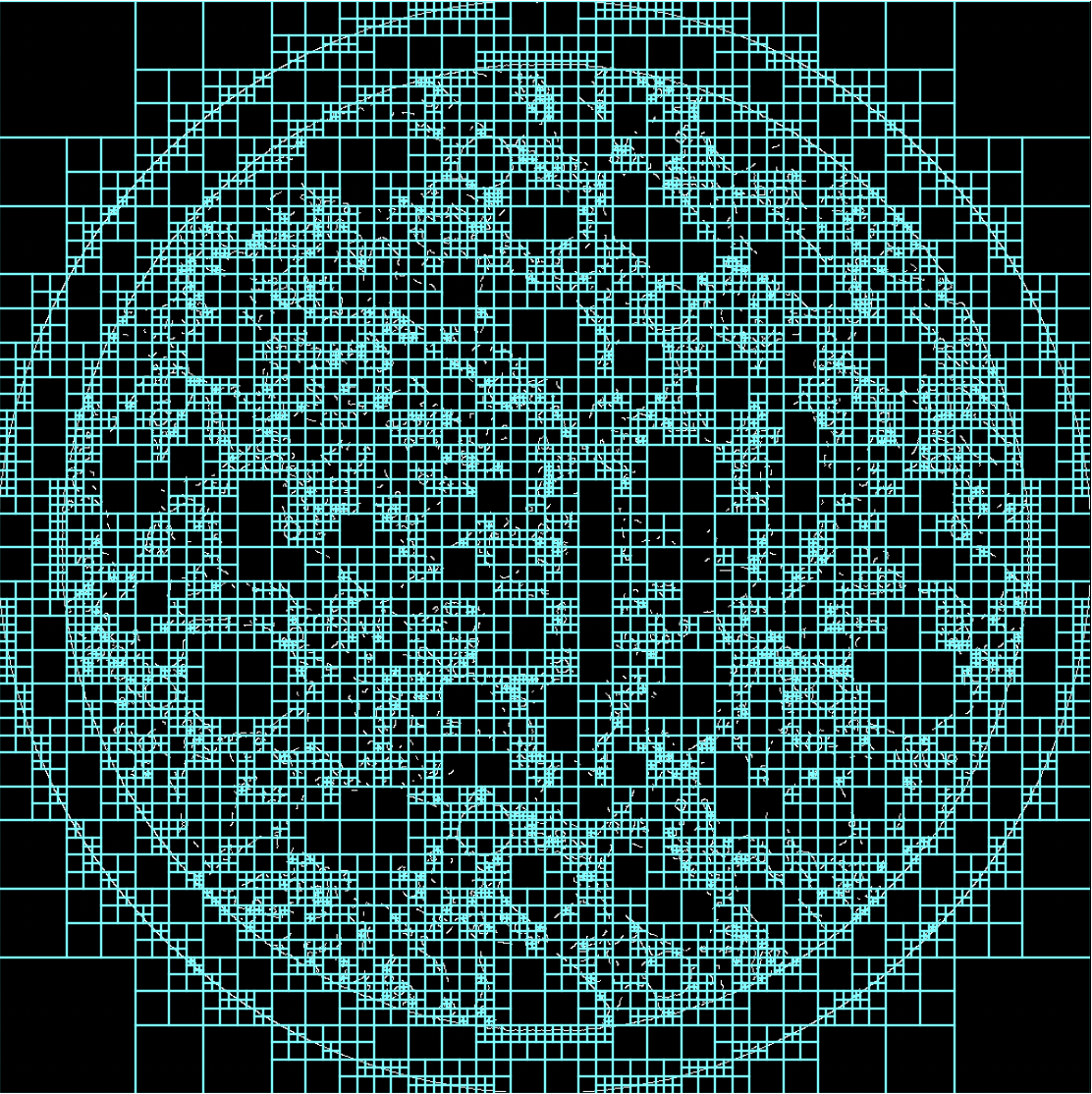}
        \end{subfigure}
        \hfill
        \begin{subfigure}[b]{0.24\linewidth}
            \caption{Hierarchy Depth: 12}
            \includegraphics[width=\textwidth]{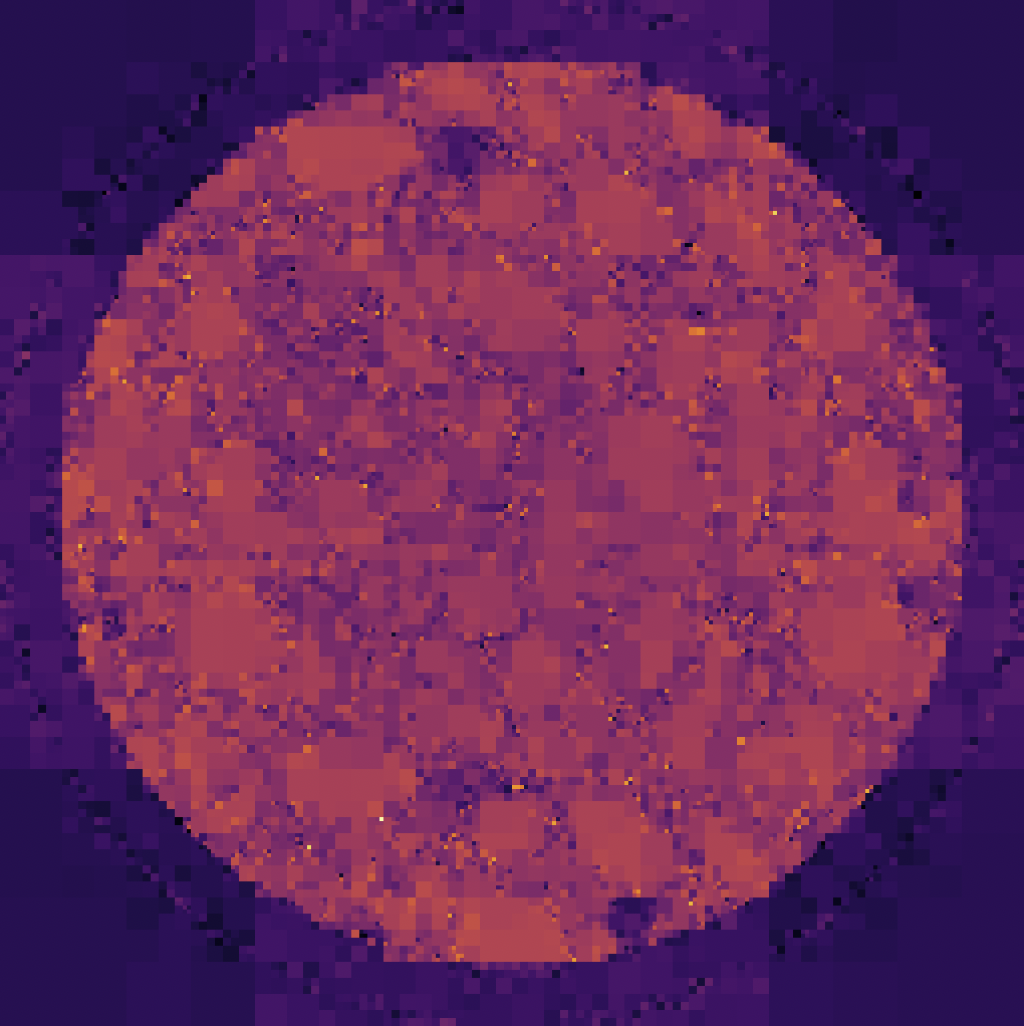}
        \end{subfigure}
        \hfill
        \begin{subfigure}[b]{0.24\linewidth}
            \caption{Input:Clean=G. Masked=R}
            \includegraphics[width=\textwidth]{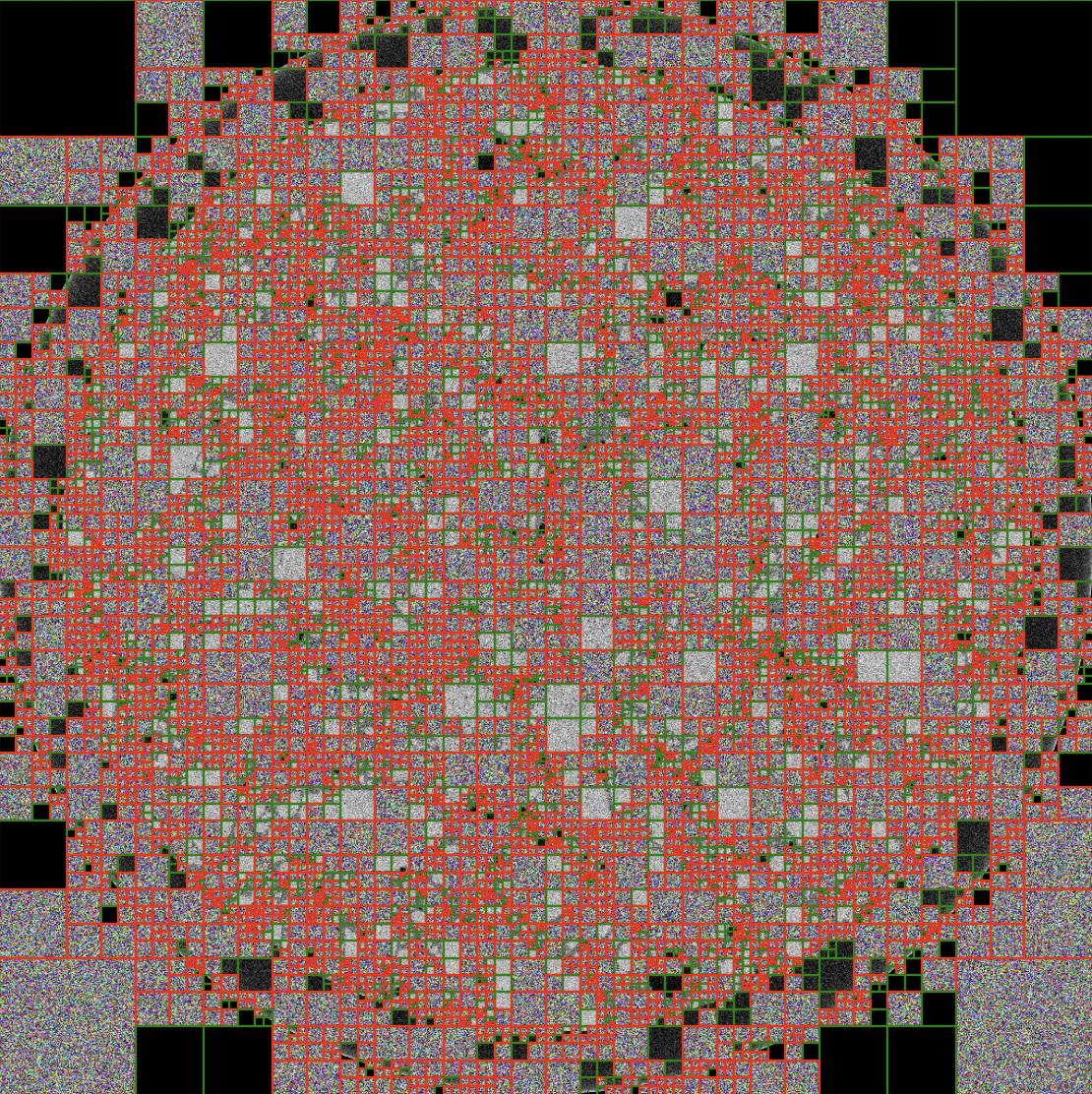}
        \end{subfigure}
    \end{minipage}

    \vfill %

    \begin{minipage}[c]{0.05\textwidth}
        \centering
        \rotatebox{90}{\shortstack[c]{\textbf{Optical Microscopy}\\\footnotesize(200\,nm; sub-micron scale)}} %
    \end{minipage}\hspace{4pt}%
    \begin{minipage}[c]{0.93\textwidth}
        \begin{subfigure}[b]{0.24\linewidth}
            \caption{$32768^2@PAIP$}
            \includegraphics[width=\textwidth]{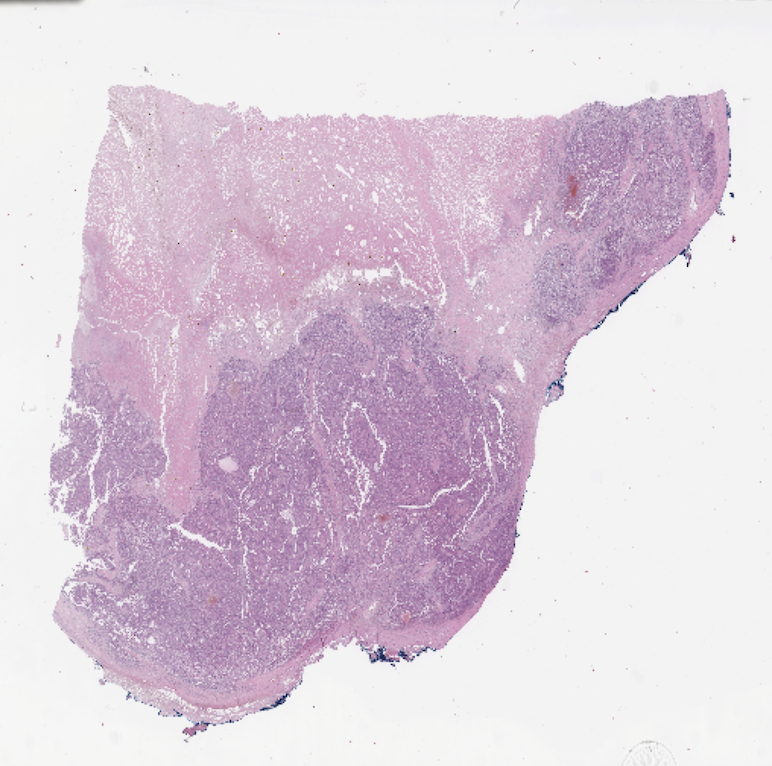}
            \caption{\textbf{Input High-Res Image}}
        \end{subfigure}
        \hfill
        \begin{subfigure}[b]{0.24\linewidth}
            \caption{Seq. Len.: 16384}
            \includegraphics[width=\textwidth]{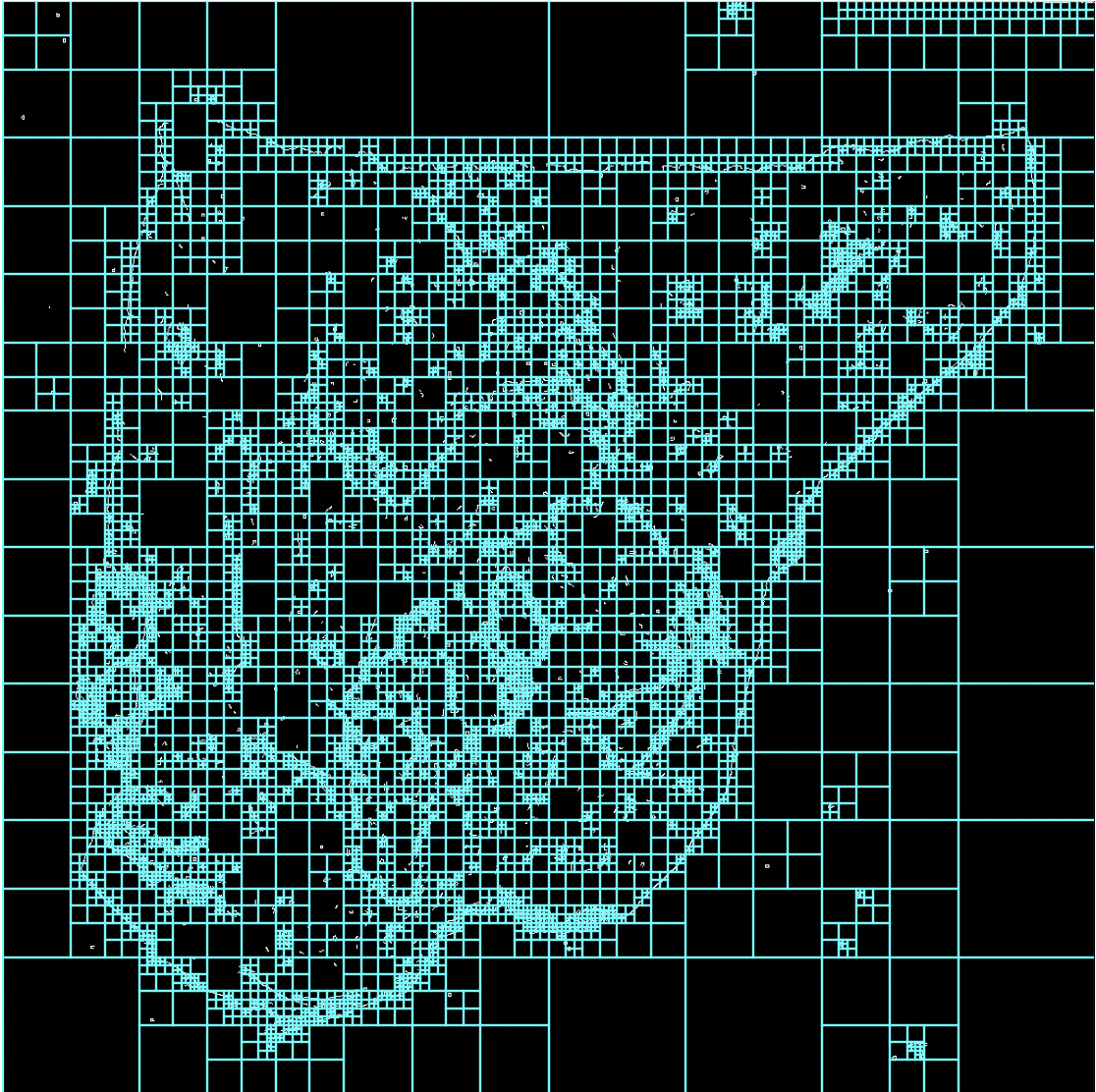}
            \caption{\textbf{Structural Patches}}
        \end{subfigure}
            \hfill
        \begin{subfigure}[b]{0.24\linewidth}
            \caption{Hierarchy Depth: 14}
            \includegraphics[width=\textwidth]{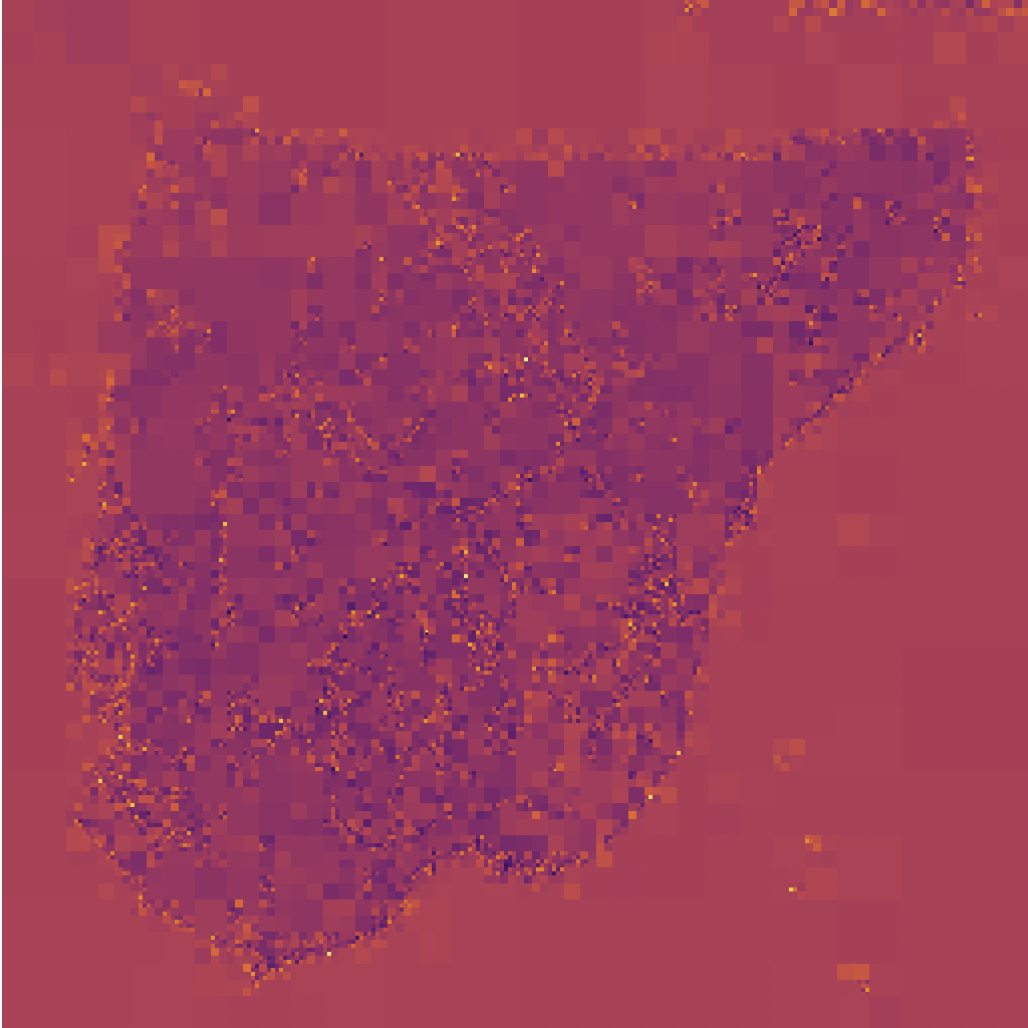}
            \caption{\textbf{Structure Noise}}
        \end{subfigure}
        \hfill
        \begin{subfigure}[b]{0.24\linewidth}
            \caption{Input:Clean=G. Masked=R}
            \includegraphics[width=\textwidth]{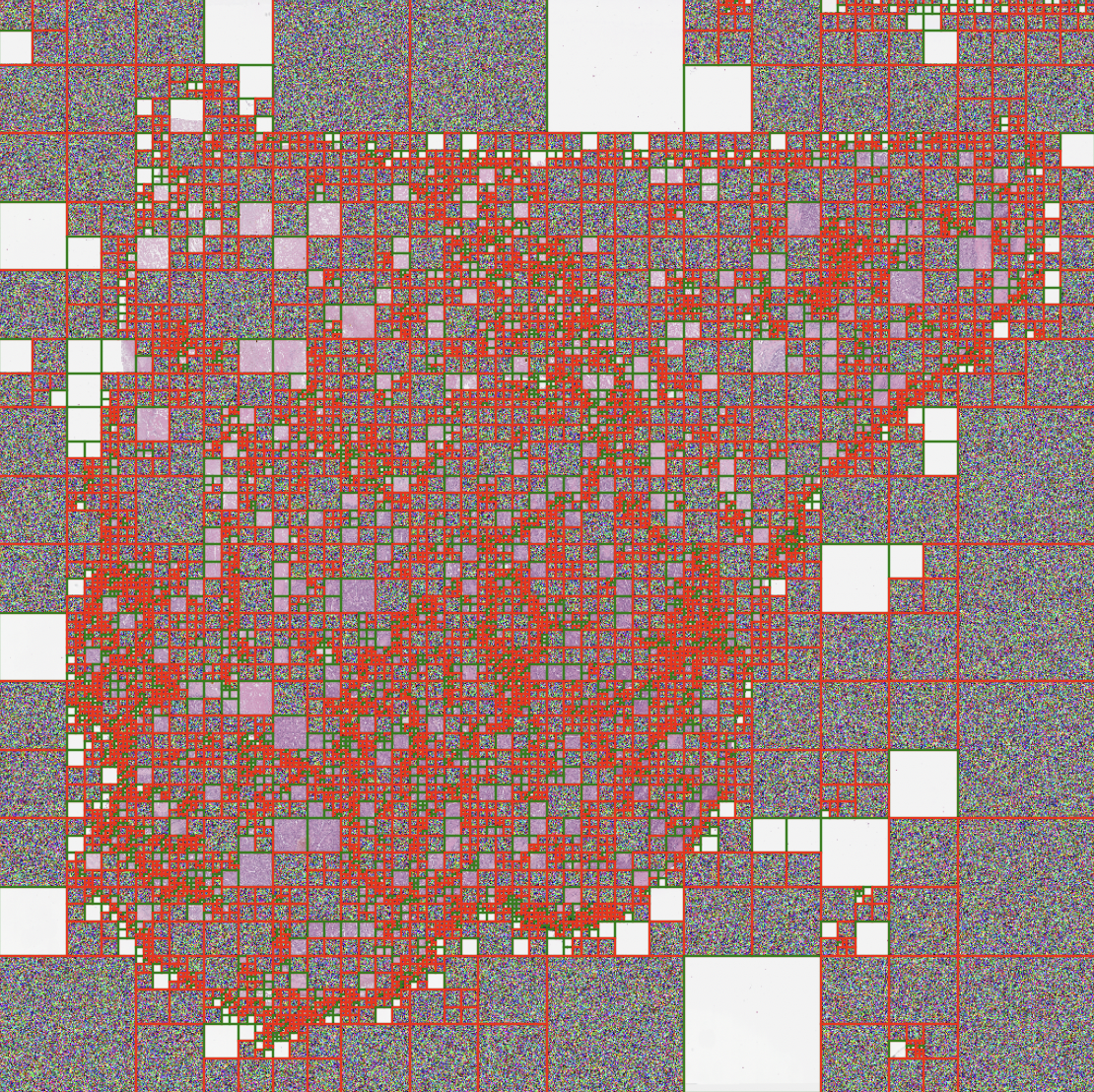}
            \caption{\textbf{Structure-Guided Mask}}
        \end{subfigure}
    \end{minipage}

    \caption{
    Adaptive structural patches and spatial information from transmission electron microscopy: HydrogelTEM-1K~\cite{Kiyama2022Nanoscale}, X-ray CT: SpringXCT-8K~\cite{van2016fast}, and WSI microscopy: PAIP-32K~\cite{KIM2021101854} datasets. These datasets span centimeter to nanometer scales and include biomedical, macroscopic, and microscopic samples. \method{} adapts to the multi-scale micro-structure of each domain for MAE pre-training.
    }
    \label{fig:seg_res} 
\end{figure*}

\section{Related Work}
\label{sec:related_work}

\begin{figure*}
    \captionsetup[subfigure]{labelformat=empty}
    \begin{subfigure}[b]{0.24\textwidth}
        \caption{HydrogelTEM-1K}
        \includegraphics[width=\textwidth]{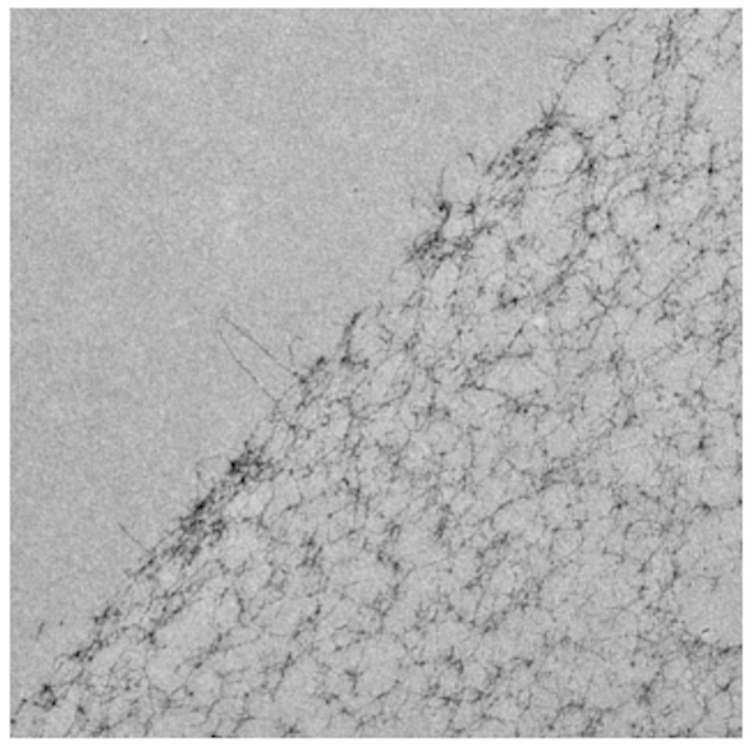}
    \end{subfigure}
    \hfill
    \begin{subfigure}[b]{0.24\textwidth}
        \caption{Dice Score:100\%}
        \includegraphics[width=\textwidth]{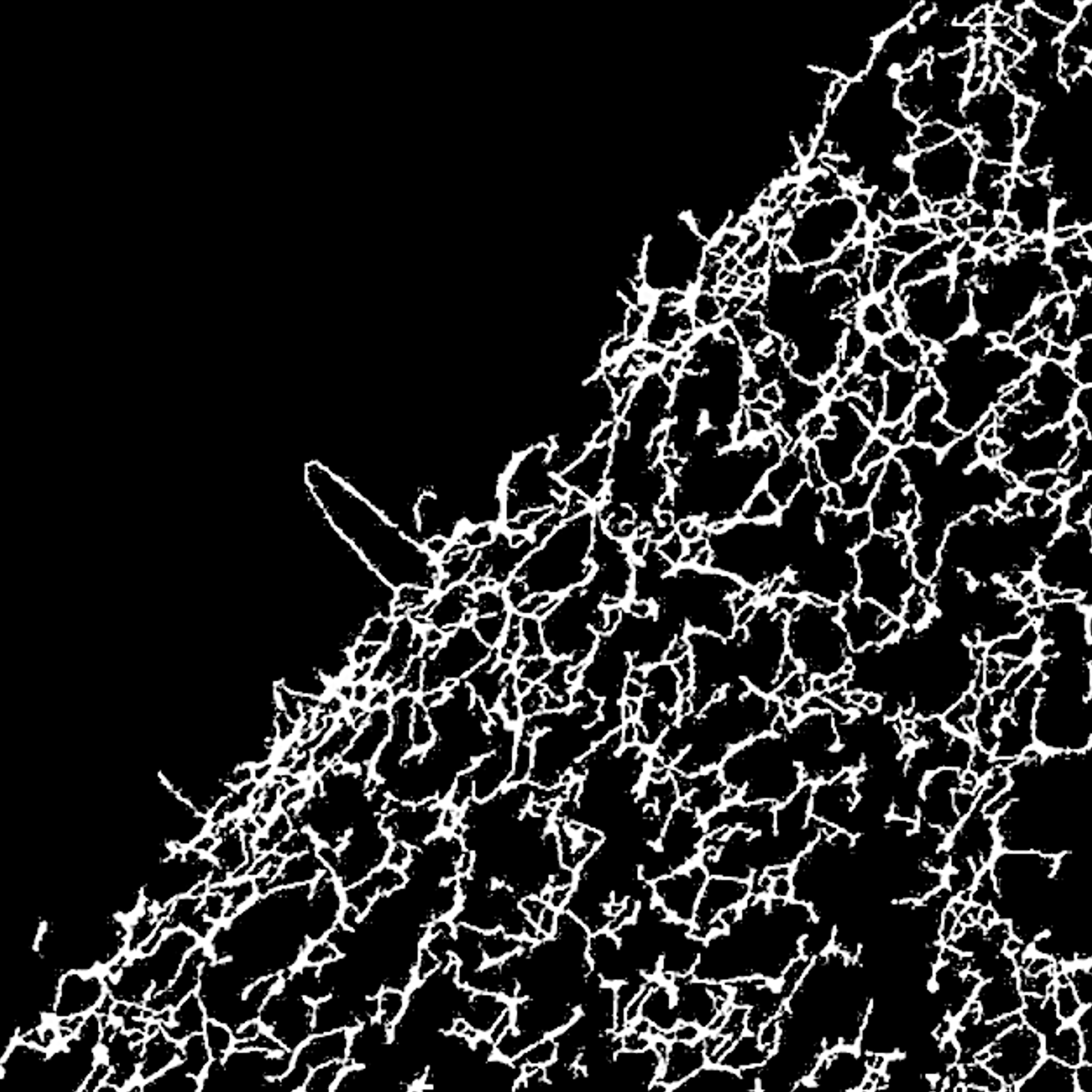}
    \end{subfigure}
    \hfill
    \begin{subfigure}[b]{0.24\textwidth}
        \caption{Dice Score:72.31\%}
        \includegraphics[width=\textwidth]{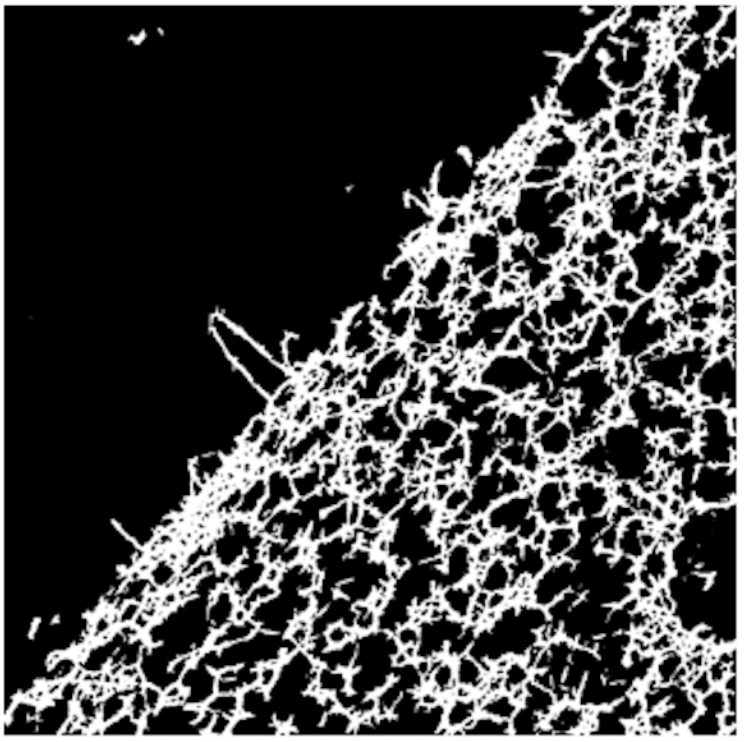}
    \end{subfigure}
    \hfill
    \begin{subfigure}[b]{0.24\textwidth}
        \caption{Dice Score:73.47\%}
        \includegraphics[width=\textwidth]{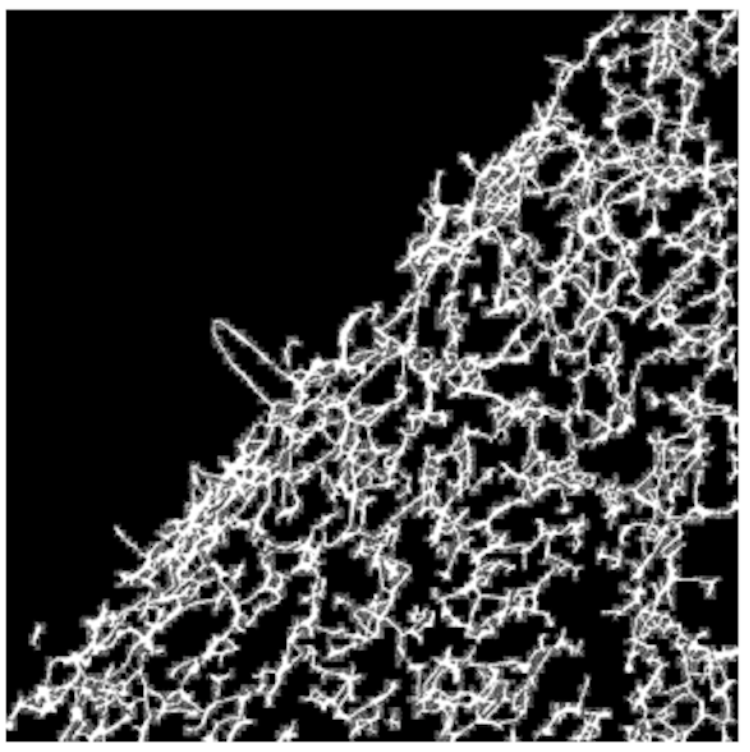}
    \end{subfigure}
        \hfill
    \begin{subfigure}[b]{0.24\textwidth}
        \caption{SpringXCT-8K}
        \includegraphics[width=\textwidth]{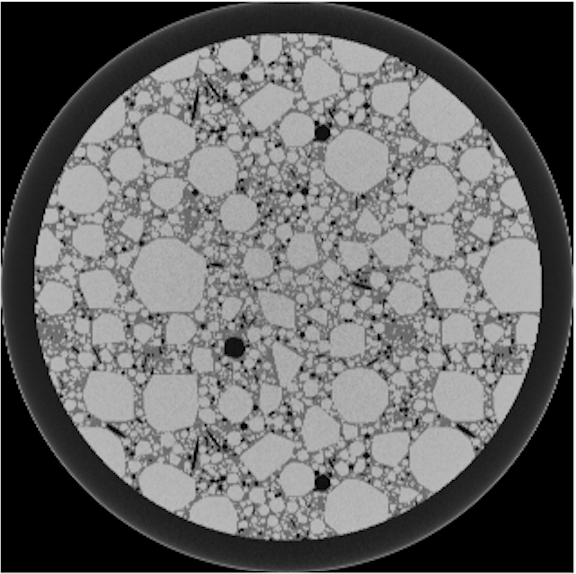}
    \end{subfigure}
    \hfill
    \begin{subfigure}[b]{0.24\textwidth}
        \caption{Dice Score:100\%}
        \includegraphics[width=\textwidth]{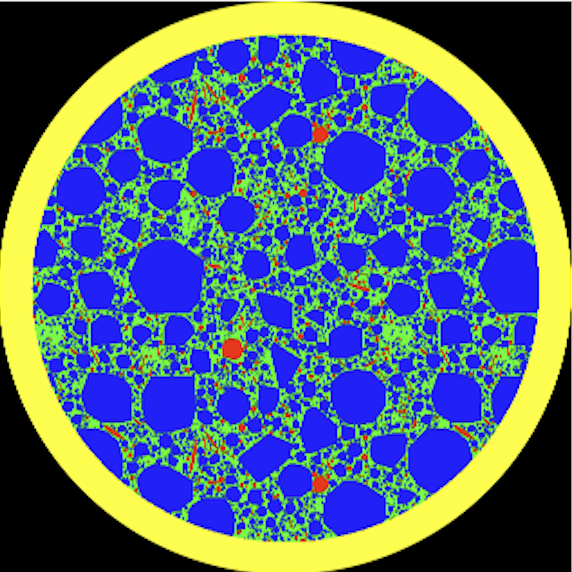}
    \end{subfigure}
    \hfill
    \begin{subfigure}[b]{0.24\textwidth}
        \caption{Dice Score:82.68\%}
        \includegraphics[width=\textwidth]{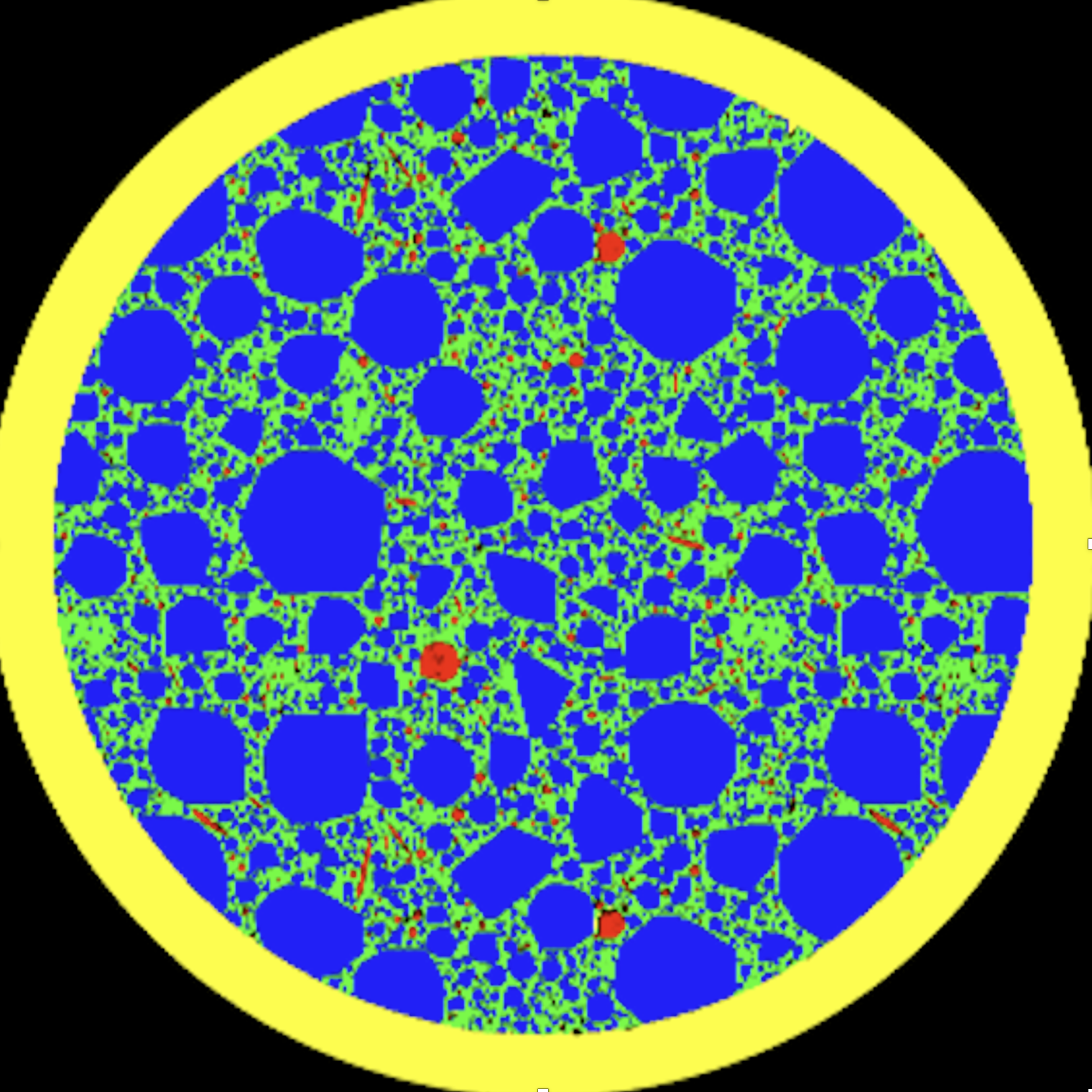}
    \end{subfigure}
    \hfill
    \begin{subfigure}[b]{0.24\textwidth}
        \caption{Dice Score:95.68\%}
        \includegraphics[width=\textwidth]{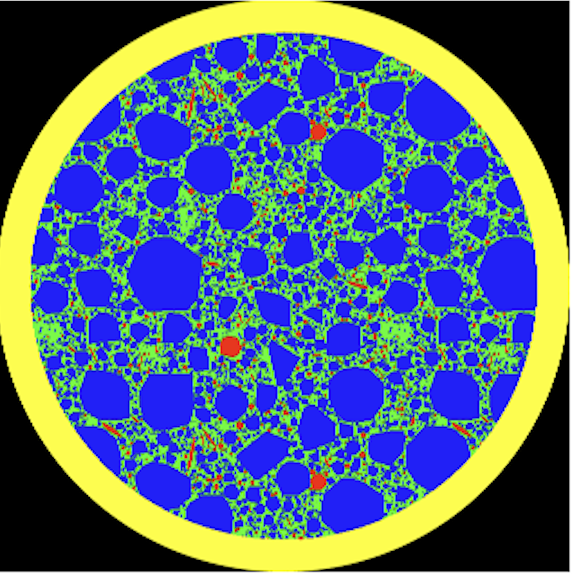}
    \end{subfigure}
    \hfill
    \begin{subfigure}[b]{0.24\textwidth}
        \caption{PAIP-32K}
        \includegraphics[width=\textwidth]{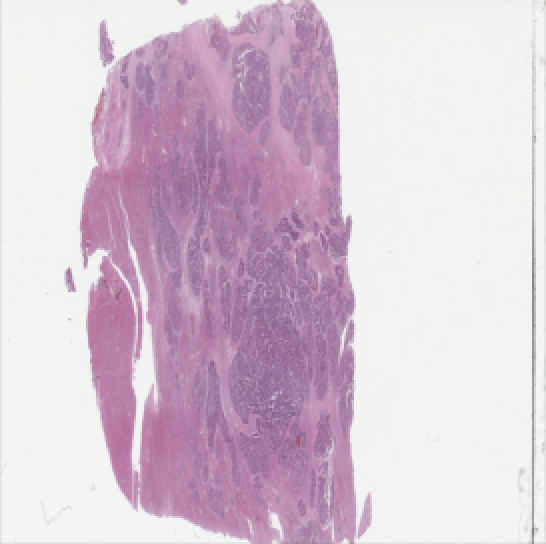}
        \caption{\textbf{Input High-Res Image}}
    \end{subfigure}
    \hfill
    \begin{subfigure}[b]{0.24\textwidth}
        \caption{Dice Score:100\%}
        \includegraphics[width=\textwidth]{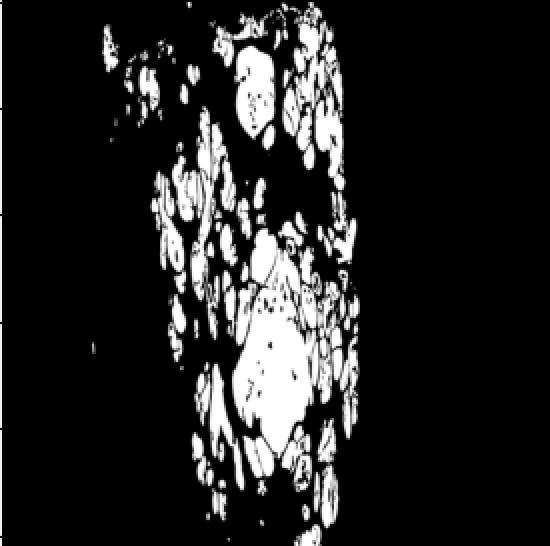}
        \caption{\textbf{Ground truth}}
    \end{subfigure}
        \hfill
    \begin{subfigure}[b]{0.24\textwidth}
        \caption{Dice Score:65.78\%}
        \includegraphics[width=\textwidth]{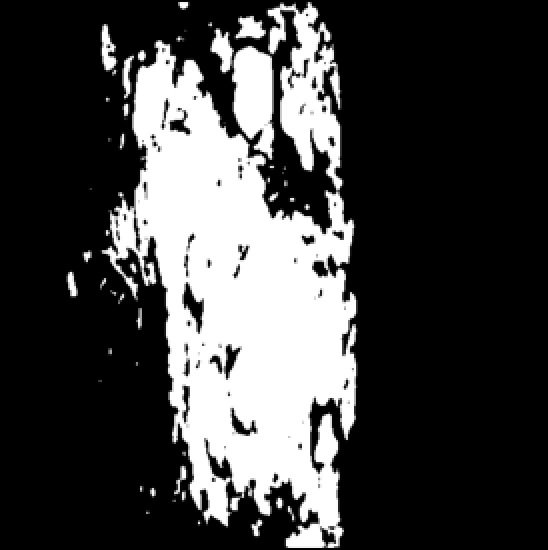}
        \caption{\textbf{MAE-SAM}}
    \end{subfigure}
    \hfill
    \begin{subfigure}[b]{0.24\textwidth}
        \caption{Dice Score:82.11\%}
        \includegraphics[width=\textwidth]{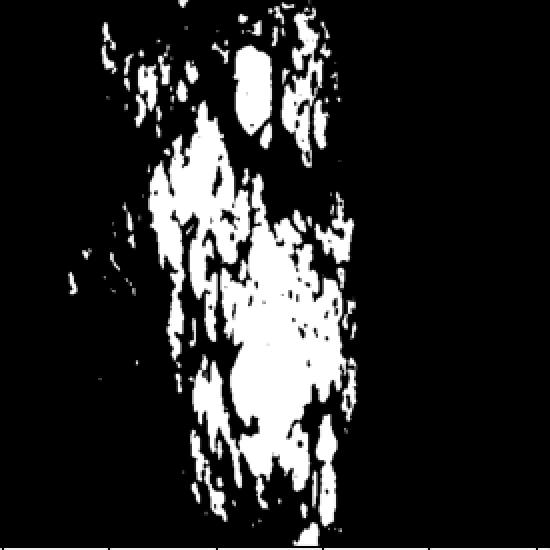}
        \caption{\textbf{\method{}-SAM}}
    \end{subfigure}
    \caption{We compared the segmentation difference between MAE-SAM~\cite{kirillov2023segment} and \method{}-SAM on TEM HydrogelTEM-1K, X-ray CT SpringXCT-8K, and WSI PAIP-32K datasets. At the same GPU compute budgets, \method{}-SAM can go down to patch size of $8\times8$ (vs. $1,024\times1,024$ at best for SAM before going OOM). As a result, \method{}-SAM can extract and express mask details better than MAE-SAM, with the gap in accuracy favoring \method{}-SAM as the resolution gets higher.}
    \label{fig:seg_comparison}
\end{figure*}

Our \method{} framework lies at the intersection of three core research directions: self-supervised learning through masked image modeling, efficient Transformer architectures for long-sequence, and high-resolution scientific imaging~\cite{Kiyama2022Nanoscale, noguchi2024real, KIM2021101854, cnudde2013high, shi2020application}.

\emph{Masked Image Modeling}. Self-supervised learning(SSL) aims to learn rich data representations without human-provided labels. While early methods focused on contrastive learning \cite{chen2020simple, he2020momentum}, Masked Image Modeling (MIM) has recently emerged as a highly effective and scalable approach, inspired by the success of Masked Language Modeling (e.g. BERT \cite{devlin2019bert}) in NLP. Masked Autoencoder (MAE) \cite{he2022masked} is a prominent example, operating by randomly masking a large portion (e.g. 75\%) of input image patches and tasking a model to reconstruct the original, uncorrupted pixels in the masked regions. Its asymmetric encoder-decoder architecture, where the encoder processes only a small subset of visible patches, makes pre-training highly efficient. However, MAE's design presents two fundamental challenges for our target domain of giga-pixel scientific images. First, its \emph{uniform patching} scheme becomes computationally intractable at giga-pixel resolutions. Second, its \emph{random masking} task is designed for semantic in-painting, which may be suboptimal for scientific images where the pre-training goal should be learning fine-grained textures, cell morphologies, and subtle, physically-grounded anomalies. Our work addresses both limitations.

\emph{Efficient Transformers for Long Sequences}. A standard ViT's self-attention mechanism has a computational and memory cost of $O(N^2)$ with respect to the sequence length $N$, which increases as the patch size decreases. For giga-pixel images, $N$ can easily exceed $10^6$ patches, making standard self-attention infeasible. To mitigate this, several strategies have been developed. \emph{Sequence parallel methods} and efficient kernels, including Deep-Speed Ulysses~\cite{jacobs2023deepspeed}, LightSeq~\cite{li2023lightseq}, RingAttention~\cite{liu2023ring}, LLS~\cite{wang2023ultralong}, and FlashAttention~\cite{FlashAttention22, dao2023flashattention2}, optimize the computation. \emph{Linear approximation methods} reduce complexity through techniques like spectral attention~\cite{pmlr-v162-dao22a, bo2023specformer, NEURIPS2021_b4fd1d2c}, low-rank approximation~\cite{choromanski2020rethinking, Katharopoulos20}, sparse attention matrix sampling~\cite{sparse-transformer19, reformer20, Roy21, beltagy2020longformer, bigbird20}, or infrequent self-attention updates~\cite{ying2021lazyformer, rabe2022selfattention}. These methods reduce the computational load, but excessive reduction can lead to performance loss~\cite{Shi21}. \emph{Hierarchical training of ViTs}, where multiple transformers are trained at different resolution levels~\cite{Si21, Chen22, chen2021crossvit, yu2023megabyte}, is another approach, but using multiple transformers increases training time and complexity. Recently, \emph{quadtrees} have been used in image segmentation to reduce attention cost, e.g. quadtree/octree attention or patch pre-processing~\cite{tang2022quadtree, ibing2023octree, zhang2024adaptive}. Both of those approaches employ quadtrees, but involve additional model complexity or need expert knowledge for hyperparameters. Our Structure-Guided Feature Tokenizer (SGFT) mechanism differs fundamentally by using the quadtree purely as an \emph{adaptive input tokenizer} to generate a fixed-length sequence for a standard ViT, and most importantly, we \emph{interleave the tree construction with our structure-conditioned sampling}. This joint formulation of a multi-scale data structure and a physics-inspired pre-training objective is the core novelty of our \method{} framework. 
A more detailed side-by-side comparison with representative long-sequence and hierarchical Transformer models is provided in Appendix~\ref{sec:supp_longseq} (Table~\ref{tab:long_sequence_summary}), offering clearer context for our \method{}’s positioning.

\section{Methodology}
\label{sec:method}

\method{} consists of three parts, as shown in Figure~\ref{fig:hdme_overview}: preprocessing \cite{rong2014improved}, multi-scale tree construction, and MAE pre-training. The key component is structure-conditioned modeling, realized through Structure-Guided Feature Tokenizer (SGFT) construction and information sampling.

\subsection{Structure-Guided Feature Tokenizer}
\label{sec:hierarchical_patching}

We construct a multi-scale extractor over the image 
$x$ by recursively partitioning its domain.
However, the standard greedy selection, $n^*=\arg\max V(n_j)$, is inherently sequential and non-parallelizable, creating a computational bottleneck as the number of active $k$ increases. To address this, we "soften" the selection by reformulating the splitting criteria as a probabilistic sampling process. We define a probability distribution $\pi$ over the set of active tokens $\mathcal{A}_k$, where the probability of selecting a token $n_j$ is proportional to its salience score. The transition from current token set $\mathcal{A}_k$ to next set $\mathcal{A}_{k+s-1}$ is defined recursively:
\begin{align}\scriptsize
\label{eq:ahp_split}
\mathcal{A}_{k+s-1} = \begin{cases} 
\mathcal{A}_k, & \text{if $k \geq N$} \\
(\mathcal{A}_k \setminus \{n^*\}) \cup \text{Split}(n^*), &  n^* \sim \pi(\cdot | \mathcal{A}_k, \tau)
\end{cases}
\end{align}
where $N$ is the target fixed sequence length for the model. The probability distribution $\pi$ is defined using the softmax function with a temperature parameter $\tau$:
\begin{align}\scriptsize
\label{eq:softmax_sampling}
\pi(n_j | \mathcal{A}_k, \tau) = \frac{\exp(V(n_j) / \tau)}{\sum_{n_i \in \mathcal{A}_k} \exp(V(n_i) / \tau)}
\end{align}
Here, $V(n_j)$ is the scoring criterion (e.g. the sum of edge map values) for token $n_j$. The temperature $\tau$ controls the "softness" of the selection: as $\tau \to 0$, $\pi$ approaches the original deterministic `argmax` selection, while as $\tau \to \infty$, it approaches a uniform random selection. Beyond removing the sequential bottleneck, this softmax-batched formulation is also substantially faster in wall-clock terms than the SHF/AP-style $\arg\max$ tokenizer (up to $20\times$ at $N{=}10^6$ tokens) at matched tokenization quality; we report a full per-resolution comparison in Appendix~\ref{sec:supp_hat_construction} (Table~\ref{tab:hat_construction}).

\subsection{Structure-Guided Information Sampling}
\label{sec:noise_generation}
We define the state of the joint process at step $k$ (representing a tree with $k$ tokens) as the tuple $X_k = (\mathcal{A}_k, C_N^{(k)})$, where $\mathcal{A}_k$ is the set of active tokens and $C_N^{(k)}$ is the full-resolution cumulative structure canvas. The transition $X_k \to X_{k+s-1}$ is defined recursively:
\begin{align}\scriptsize
\label{eq:joint_recursive_update}
X_{k+s-1} = \begin{cases} 
X_k, & \text{if $k \geq N$} \\
(\mathcal{A}_{k+s-1}, C_N^{(k+s-1)}), & \text{otherwise}
\end{cases}
\end{align}
where $N$ is the target sequence length. The process is initialized at $k=1$ with the root state $X_1 = (\mathcal{A}_1, C_N^{(1)})$. This state consists of the initial token set $\mathcal{A}_1 = \{n_0\}$, which contains only the root node $n_0$ (at depth $j=0$), and the initial structure canvas $C_N^{(1)}$. The canvas is populated with a single "root information" layer $\epsilon_0 \sim \mathcal{N}(0, \sigma_0^2 x)$, where the global variance $\sigma_0^2 = V_T(x) / \sqrt{0+1} = V_T(x)$.

For any state $X_k$ where $k < N$, the transition $X_k \to X_{k+s-1}$ is executed. This joint update consists of two simultaneous actions. First, a structural update determines the new token set $\mathcal{A}_{k+s-1}$ by sampling a parent node $n^*$ (at depth $j-1$) $\sim \pi(\cdot | \mathcal{A}_k, \tau)$ and replacing it with its $s$ children, as defined in \cref{eq:ahp_split}. Second, a concurrent information update forms the new canvas $C_N^{(k+s-1)} = C_N^{(k)} + \Delta C_N^{(k+1)}$. The sparse update tensor $\Delta C_N^{(k+1)}$ adds a new, localized noise component $\epsilon_{\text{new}}^{(c)}$ for each new child node $n^{(c)}$ (now at depth $j$). The variance of this new information is governed by
$
\text{Var}(\epsilon_{\text{new}}^{(c)}) = \frac{\sigma_T^2(c)}{\sqrt{j+1}}
$
where $\sigma_T^2(c) = V_T(p^{(c)})$ is the local target variance from the child's patch. This process repeats until the tree contains $N$ tokens.

This generative process defines a sequence of nested information additions along the ancestral path $\mathcal{P}(n_i) = (n_0, \dots, n_d)$ for any token $n_i$. Due to the additivity of Gaussian distribution, the final distribution $N_i$ is the sum of all independent components $\epsilon_j$ added along this path: $N_i = \sum_{j=0}^{d} \epsilon_j$. The final variance $\sigma_{\text{final}, i}^2$ is therefore the sum of the variances of these components. 
Thus, the total variance is just the sum of depth-scaled local variances:
\begin{equation}\scriptsize 
\label{eq:final_variance_precise}
\sigma_{\text{final}, i}^2 = \sum_{j=0}^{d} \text{Var}(\epsilon_j) = \sum_{j=0}^{d} \frac{\sigma_T^2(j)}{\sqrt{j+1}} = \sum_{j=0}^{d} \frac{1}{\sqrt{j+1}\,|S_j|} \sum_{(h, w) \in S_j} x(h, w)
\end{equation}
where $p_j$ is the patch of ancestor node $n_j$ at depth $j$, and $S_j$ its pixel coordinates. The accumulation grows sub-linearly with depth, avoiding variance explosion.

\subsection{SGMA Pre-training Framework}
\label{sec:pretraining}

\noindent
\textbf{Image Encoder and Input}.
After completing adaptive structural patching (\cref{sec:hierarchical_patching}) and structure-conditioned sampling distribution generation, we obtain the adaptive structural patch sequence $\mathcal{P} = \{p_i\}_{i=1}^{N_h}$ and the final structure-conditioned sampling distribution $N_i$ for each patch. We randomly select a subset of indices $\mathcal{I}_{\text{masked}} \subset \{1, \dots, N_h\}$ (e.g. 75\% suggested in \cite{he2022masked}) to be corrupted. We then form a visible input sequence $\mathcal{P}' = \mathcal{P} -  \mathcal{I}_{\text{masked}}$. We use a standard ViT encoder $f_\theta$ \cite{dosovitskiy2021image}. Following the Masked Autoencoder (MAE) paradigm, the encoder $f_\theta$ processes the \textit{structure-aware}, clean sequence $\mathcal{P}'$.

\noindent
\textbf{Reconstruction}.
A lightweight decoder $g_\phi$ is attached to the encoder $f_\theta$. The decoder receives the full sequence of encoded representations $Z=f_\theta(\mathcal{P}')$ and is tasked with reconstructing the \textit{original, clean} version of \textit{only the noised patches}. The model learns to leverage the clean patches ($p_j{\in}\mathcal{P}'$) as context to reconstruct corrupted patches ($i{\in}\mathcal{I}_{\text{masked}}$). This pre-training objective function $\mathcal{L}_{\text{SGMA}}$ is:
\begin{equation}\scriptsize 
\label{eqn:hdme-obj}
\min_{\theta, \phi}\; \mathbb{E}_{I\sim\mathcal{D},\, \mathcal{I}_{\text{masked}}\sim\text{Mask}} \Bigg[ \sum_{i \in \mathcal{I}_{\text{masked}}} \mathcal{L}_{\text{MSE}}\Big( g_\phi\big(f_\theta(\mathcal{P}'), i\big),\ p_i \Big) \Bigg]
\end{equation}
Where $I$ is an image from dataset $\mathcal{D}$, $\mathcal{P}'$ is the corrupted input sequence, $p_i$ is the clean patch, $g_\phi(\cdot, i)$ is the decoder prediction for the $i$-th patch, and $\mathcal{L}_{\text{MSE}}$ is the L2 loss.

\subsection{Fine-tuning}
\label{sec:fine-tuning}

\paragraph{Fine-tuning Encoder}. After pre-training with $\mathcal{L}_{\text{SGMA}}$, the reconstruction decoder $g_\phi$ is discarded. We load the weights of our pre-trained encoder $f_\theta$ and adapt it for downstream tasks, such as segmentation. We retrofit our \method{} scheme into a segmentation framework, minimally adapting the encoder $f_\theta$ to process high-resolution inputs.

\noindent \textbf{Fine-tuning Objective}. During the fine-tuning phase, we do not directly generate full-resolution masks $Y$ but instead generate encoded mask logits $Y' = \{y'_i\}_{i=1}^N$ at the patch level. This saves memory and reduces backpropagation costs. We summarize the fine-tuning objective function:
\begin{align}\scriptsize 
\label{eqn:hft-obj}
\min_\theta \mathbb{E}_{\substack{(I,Y)\sim\mathcal{D}}}
\Bigg[ \sum_{i=1}^N \mathcal{L}_{\text{seg}}\Big( h_\theta(Z_i),\ T_{\mathrm{tree}}(Y)_i \Big) \Bigg]
\end{align}
Where $N$ is the total number of patches, $\theta$ represents the trainable weights of the encoder and a simple linear head $h_\theta$, $\mathcal{L}_{\text{seg}}$ denotes a segmentation loss (e.g. Dice and Cross-Entropy), $Z_i = f_\theta(\mathcal{P})_i$ is the encoder output for patch $i$, and $T_{\mathrm{tree}}(Y)_i$ is the $i$-th patch from the ground-truth mask $Y$, generated using the same hierarchical tree operator $T_{\mathrm{tree}}$ applied to the input $I$.

\section{Experiments}
\label{sec:exp}

\subsection{High Resolution Scientific Image Datasets}
\textbf{HydrogelTEM-1K:} is a nanoscale hydrogel network dataset  \cite{Kiyama2022Nanoscale, noguchi2024real} visualized using transmission electron microscopy (TEM). To preserve structure, the target PAMPS network was interpenetrated with a PDMAAm "skeleton network". A novel mineralization technique using Fe³+ ions (forming AFO nanoparticles) provided the necessary electron contrast. The resin-substituted samples were cut into 100 nm thick specimens and observed by Hitachi and JEOL TEM/STEM systems (100-200 kV).

\noindent 
\textbf{SpringXCT-8K:} A high-resolution XCT dataset of 46 concrete and asphalt specimens scanned at SPring-8 (200 keV)~\cite{10.1117/12.3027913}. Each 3D volume has size $8{,}192 \times 8{,}192 \times 28{,}800$. For \textit{pre-training}, we use $308{,}000$ slices of size $8{,}192 \times 8{,}192$.
For \textit{fine-tuning}, we build a simulation pipeline to generate ultrahigh-resolution data. We start from virtual microstructures~\cite{ziabari2025pycmg}, resample them to 12.02 $\mu$m resolution, and obtain $1{,}500$ slices of size $8{,}192 \times 8{,}192$. We then simulate the imaging process with realistic attenuation, and add degradations such as sparse angles, Poisson/Gaussian noise, and ring artifacts. The sinograms are reconstructed using FBP~\cite{shepp1974fourier, buzug2011computed}. This yields 60 synthetic volumes with ground-truth labels and corresponding reconstructions (see appendix) for training.

\noindent \textbf{PAIP-32K:}~\cite{KIM2021101854} is a high-resolution, real-world liver cancer pathology dataset, with sample resolutions up to $64K^2$, significantly surpassing those of conventional image datasets. PAIP contains $2,457$ Whole-Slide Images (WSIs). When lower resolutions are needed, we downscale the images to uniform sizes of $[16,384, 32,768, 65,536]$ square pixels. For training, we randomly select $70\%$ of samples, $10\%$ for validation, and $20\%$ for testing. All datasets are shuffled and normalized to $[0.0,1.0]$ as input for the model.

\subsection{Evaluating Models: Baselines \& Proposed}
\noindent \textbf{Baseline Models:} Our primary baseline is the standard Masked AutoEncoder (MAE) \cite{he2022masked} pre-training strategy. We evaluate this on the \wahib{widely-used} segmentation model SAM \cite{kirillov2023segment}, which uses a ViT encoder (ViT-Base, Large, or Huge) pre-trained with MAE; we refer to this configuration as MAE-SAM. We also benchmark against U-Net-style models \cite{ronneberger2015u}, which use a contraction-expansion path with skip connections. Our main baseline in this category is the UNETR \cite{hatamizadeh2022unetr}, for which we create an MAE-UNETR variant by pre-training its ViT encoder by the standard MAE objective.

\noindent \textbf{Proposed Models:}
We replace MAE pre-training with the proposed \method{} objective (Eq. ~\ref{eqn:hdme-obj}) and apply it to the same architectures, yielding \method{}-SAM and \method{}-UNETR. 
For SAM-based models, we use symmetrical depatching~\cite{zhang2025shf}, eliminating the need for a heavy decoder. For SpringXCT-3D data, we apply the SAM~2~\cite{ravi2024sam2} encoder to 2D slices and stack the outputs for final prediction. 
For comparison, we include U-Net~\cite{ronneberger2015u} (ImageNet-pretrained via \texttt{timm}~\cite{jain2022hugging}), UNETR~\cite{hatamizadeh2022unetr}, and Swin-UNet~\cite{cao2022swin}.

\begin{table}[t]
\caption{Quantitative segmentation results (Dice Score \% $\uparrow$) on all datasets. Our \method{} pre-training method is compared against standard baselines and MAE pre-training on identical architectures. The "Improvement" column shows the Dice Score gain of our \method{} model over its equivalent MAE-trained baseline (e.g. FT-SGMA-UNETR vs. FT-MAE-UNETR).
}
\centering
\label{tab:combined_results}
\resizebox{\linewidth}{!}{%
\renewcommand{\arraystretch}{0.8}%
\begin{tabular}{c l c c c l}
\toprule
\textbf{Dataset} & \textbf{Model} & \textbf{Patch Size} & \textbf{Sequence Length} & \textbf{Dice Score (\%)} & \textbf{Performance Improvement} \\
\midrule
\multirow{10}{*}{\rotatebox[origin=c]{90}{\textbf{HydrogelTEM-1K}}} & SAM~2~\cite{ravi2024sam2} & 8 & 16384 & 69.87 & - \\
 & UNet~\cite{ronneberger2015u} & - & - & 71.03 & - \\
 & UNETR~\cite{hatamizadeh2022unetr} & 16 & 4096 & 70.59 & - \\
\cmidrule(lr){2-6}
 & FT-MAE-SAM & 8 & 16384 & 72.31 & (Baseline) \\
 & FT-MAE-SAM~2 & 8 & 16384 & 73.12 & (Baseline) \\
 & FT-MAE-UNETR & 16 & 4096 & 74.56 & (Baseline) \\
\cmidrule(lr){2-6}
 & \textbf{FT-SGMA-SAM (ours)} & 2 & 16384 & 73.48 & +1.17 (vs. FT-MAE-SAM) \\
 & \textbf{FT-SGMA-SAM~2 (ours)} & 2 & 16384 & 74.23 & +1.11 (vs. FT-MAE-SAM~2) \\
 & \textbf{FT-SGMA-UNETR (ours)} & 2 & 16384 & \textbf{75.33} & \textbf{+0.77} (vs. FT-MAE-UNETR) \\
\midrule
\multirow{10}{*}{\rotatebox[origin=c]{90}{\textbf{SpringXCT-8K}}} & SAM~2~\cite{ravi2024sam2} & 128 & 4096 & 83.81 & - \\
 & UNet~\cite{ronneberger2015u} & - & - & 78.73 & - \\
 & UNETR~\cite{hatamizadeh2022unetr} & 128 & 4096 & 82.23 & - \\
 & Swin-UNETR~\cite{DBLP:conf/cvpr/TangY0RLXNH22} & - & - & 83.96 & - \\
\cmidrule(lr){2-6}
 & FT-MAE-SAM & 128 & 4096 & 82.68 & (Baseline) \\
 & FT-MAE-SAM~2 & 128 & 4096 & 85.98 & (Baseline) \\
 & FT-MAE-UNETR & 128 & 4096 & 86.12 & (Baseline) \\
\cmidrule(lr){2-6}
 & \textbf{FT-SGMA-SAM (ours) } & 2 & 16384 & \textbf{95.68} & \textbf{+13.00} (vs. FT-MAE-SAM) \\
 & \textbf{FT-SGMA-SAM~2 (ours) } & 2 & 16384 & 93.77 & +7.79 (vs. FT-MAE-SAM~2) \\
 & \textbf{FT-SGMA-UNETR (ours)} & 4 & 8194 & 91.93 & +5.81 (vs. FT-MAE-UNETR) \\
\midrule
\multirow{10}{*}{\rotatebox[origin=c]{90}{\textbf{PAIP-32K}}} & SAM~\cite{kirillov2023segment} & 1024 & 1024 & 62.34 & - \\
 & UNet~\cite{ronneberger2015u} & - & - & 61.38 & - \\
 & UNETR~\cite{hatamizadeh2022unetr} & 1024 & 1024 & 74.96 & - \\
 & TransUNet~\cite{chen2021transunet} & - & - & 69.88 & - \\
\cmidrule(lr){2-6}
 & FT-MAE-SAM & 1024 & 1024 & 65.78 & (Baseline) \\
 & FT-MAE-SAM~2 & 1024 & 1024 & 66.37 & (Baseline) \\
 & FT-MAE-UNETR & 1024 & 1024 & 77.24 & (Baseline) \\
\cmidrule(lr){2-6}
 & \textbf{FT-SGMA-SAM (ours)} & 4 & 16384 & 82.11 & +16.33 (vs. FT-MAE-SAM) \\
 & \textbf{FT-SGMA-SAM~2 (ours)} & 4 & 16384 & \textbf{83.21} & \textbf{+16.84} (vs. MAE-SAM~2) \\
 & \textbf{FT-SGMA-UNETR (ours)} & 8 & 8194 & 81.23 & +3.99 (vs. FT-MAE-UNETR) \\
\bottomrule
\end{tabular}%
}

\end{table}

\section{Evaluation and Results}

\subsection{High-Precision Image Segmentation}
\label{sec:seg_performance}

\noindent\textbf{Quantitative Results.} As shown in Table \ref{tab:combined_results}, our \method{} pre-training method consistently outperforms standard baselines and MAE-based pre-trained models across all three diverse datasets. The performance improvement correlates significantly with image resolution. On the \textbf{HydrogelTEM-1K} dataset, the improvement is steady but modest; our FT-SGMA-UNETR achieves the top score of 75.33\%, a \textbf{+0.77} point increase over its corresponding FT-MAE-UNETR (74.56\%). On the ultra-high resolution datasets, the advantage of \method{} becomes exceptionally pronounced. On the \textbf{SpringXCT-8K} dataset, our FT-SGMA-SAM model achieves a state-of-the-art Dice score of 95.68\%, a massive \textbf{+13.00} point improvement over the FT-MAE-SAM baseline (82.68\%). This improvement is most dramatic on the \textbf{PAIP-32K} dataset. Here, our FT-SGMA-SAM~2 model (83.21\%) surpasses the equivalent FT-MAE-SAM~2 model (66.37\%) by \textbf{+16.84} points. This demonstrates that while standard MAE performs poorly when handling ultra-high resolution inputs, our \method{} pre-training, which leverages adaptive structural patching, provides a far more effective representation for capturing both global context and micro-level details.

\noindent\textbf{Segmentation Results.} Figure \ref{fig:seg_res} provides a visual comparison on the HydrogelTEM-1K, SpringXCT-8K, and PAIP-32K datasets. Baselines like UNet produce blurry and incomplete boundaries. MAE-UNETR (which was forced to use large patch sizes during pre-training) fails to resolve individual structures, merging them into a single mass. In contrast, our SGMA-UNETR, having been pre-trained on adaptive-scale patches via SGFT, successfully resolves fine, nm-scale structures, producing a segmentation mask that is crisp and topologically accurate. This demonstrates our model's ability to learn meaningful multi-scale features. Although the main focus of the paper is ultra-high-resolution 2D imaging, the long-sequence problem also appears in 3D volumetric data; on standard 3D medical benchmarks (BTCV, KiTS19) \method{}-SAM~2 matches or marginally exceeds strong specialized baselines (Swin UNETR-V2, U-Mamba, nnFormer) on Dice while being $2.58\times$--$7.15\times$ faster thanks to SGFT-based sequence compression (Appendix~\ref{sec:supp_extra_baselines}, Table~\ref{tab:btcv_kits}). 
We also report a PAIP-16K classification comparison against ViT, HIPT, and SHF-ViT, together with a Damped Accumulation ablation in Appendix~\ref{sec:supp_cls_ablation} (Table~\ref{tab:classification_paip16k}). \method{}-ViT achieves the best performance in both Top-1 accuracy and Dice. The NoDA ablation further indicates that the impact of DA is task-dependent, with a marginal gain for classification ($+0.88$ Top-1) but a notable improvement for segmentation ($+3.00$ Dice), suggesting that DA primarily benefits dense prediction by introducing multi-scale spatial structure.

\subsection{Compute Efficiency: Training and Inference}
\label{sec:efficiency}

\noindent\textbf{The Baseline Bottleneck.}
Standard MAE-based models (MAE-UNETR, MAE-SAM1/2) must use a uniform grid. On a $32,768 \times 32,768$ image, even a large $32 \times 32$ patch size yields a sequence of $N=(1024)^2 \approx 1M$ patches, which is computationally intractable and causes Out-of-Memory (OOM) errors on any available hardware. 
To avoid OOM, baselines use very large patches (e.g. $128 \times 128$ or $256 \times 256$), losing the fine-grained details needed for our tasks.

\noindent\textbf{SGMA Efficiency.}
\method{} compresses the input into a \textit{fixed-length} sequence $K$ (e.g. $K=1024$ or $K=4096$) \textit{regardless} of the input resolution. This adaptive compression allows us to process the 32K image with the same computational budget as the 1K image. As shown in Table \ref{tab:speed-hmae}, our \method{} pre-training framework achieves a dramatic improvement in both training and inference speed compared to a MAE-based baseline (which must operate on large patches to avoid OOM). Our method is significantly faster, enabling rapid experimentation and pre-training on massive scientific datasets that were previously infeasible. Worth mentioning that, the setup of Table~\ref{tab:speed-hmae} is intentionally biased \emph{against} our method on both axes. On the MAE side, the chosen sequence length is the largest the GPU budget can fit before going OOM, i.e.\ MAE is reported at its hardware ceiling. On the \method{} side, we deliberately use a sub-optimal short sequence (e.g.\ $N{=}2{,}048$ on PAIP-32K, far below the $N{=}16{,}384$ used by our best-Dice configuration in Table~\ref{tab:combined_results}), so that compute matches and we can isolate the speedup. Even with this handicap, \method{} still beats MAE on Dice (e.g.\ $76.08$ vs.\ $62.34$ on PAIP-32K, a $+13.7$-point margin). 
Beyond raw segmentation accuracy, we also evaluate \method{} in a zero-shot setting on real samples and use the resulting masks for downstream scientific analysis (pore-network extraction on SpringXCT-8K and hydrogel skeletonization on HydrogelTEM-1K). Full quantitative and qualitative results are deferred to Appendix~\ref{sec:supp_zeroshot} due to space constraints. Full hyperparameter settings (masking ratio, sequence length, model size) are reported in Appendix~\ref{sec:supp_hparams}.

\begin{table}[t]
\centering
\caption{Inference efficiency under the \emph{same compute budget} (identical GPU count per row) for MAE-SAM and our SGMA-SAM. The MAE row uses the longest sequence that fits within its GPU budget before OOM, while \method{} uses a deliberately shorter sequence (e.g. $N{=}2{,}048$ on PAIP-32K vs.\ $N{=}16{,}384$ in Table~\ref{tab:combined_results}) to match wall-clock compute. This enables a clear comparison of SGFT speedup. SGMA still achieves higher Dice and faster inference, so the reported speedups (up to 24.8$\times$) are conservative lower bounds.}

\resizebox{\linewidth}{!}
{
\begin{tabular}{l c lcccc}
\toprule
\textbf{Datasets} & \textbf{GPUs} & \textbf{Model} & {\textbf{Time (s/img)}} & {\textbf{Seq. Len.}} & {\textbf{Dice (\%)}} & {\textbf{Speedup ($\times$)}} \\
\midrule
\multirow{2}{*}{\textbf{HydrogelTEM-1K} }
& 4& MAE-SAM-8~\cite{he2022masked,kirillov2023segment} & 0.38261 & 16384 & 72.31 & {1$\times$} \\
& 4& \textbf{SGMA-SAM} (ours) & \textbf{0.09913} & \textbf{8194} & \textbf{72.56} & \textbf{3.86$\times$} \\
\midrule
\multirow{2}{*}{\textbf{SpringXCT-8K}}
& 128& MAE-SAM-128~\cite{he2022masked,kirillov2023segment} & 2.5168 & 4096 & 82.68 & {1$\times$} \\
& 128& \textbf{SGMA-SAM} (ours) & \textbf{0.3512} & \textbf{1024} & \textbf{89.37} & \textbf{7.17$\times$} \\
\midrule
\multirow{2}{*}{\textbf{PAIP-32K}}
& 512& MAE-SAM-1024~\cite{he2022masked,kirillov2023segment} & 8.9812 & 16384 & 62.34 & {1$\times$} \\
& 512& \textbf{SGMA-SAM} (ours) & \textbf{0.3663} & \textbf{2048} & \textbf{76.08} & \textbf{24.8$\times$} \\
\bottomrule
\end{tabular}%
}
\label{tab:speed-hmae}
\end{table}

\subsection{Limitations and Effect of Hyperparameters}
\label{sec:limitations}
We summarize how the three main hyperparameters of \method{} affect performance, and which of these effects translate into practical limitations. The full visualizations are in Appendix~\ref{sec:supp_hparams} (Figures~\ref{fig:supp_ratio}--\ref{fig:supp_model}).

\noindent\textbf{Masking ratio.} $r$ (Figure~\ref{fig:supp_ratio}). Reconstruction quality degrades gracefully as $r$ grows from $0.25$ to $0.75$, and the per-quadtree-level sampling preserves structurally important regions across ratios. We did not observe a clear failure mode in this range, so the choice of $r$ is \emph{not} a practical limitation. \emph{Limitation:} \textbf{Sequence length.} $N$ (Figure~\ref{fig:supp_seq}). The fixed SGFT token budget trades coverage for detail; under-sized $N$ on highly heterogeneous inputs (e.g.\ PAIP-32K with $N{=}1024$) leaves fine micro-structures coarsely tokenized and produces a measurable drop in Dice. Choosing $N$ therefore requires a per-dataset budget tuned to the input resolution and structural density. \noindent\textbf{Model size} (Figure~\ref{fig:supp_model}). Reconstruction quality improves monotonically from ViT-L to ViT-XL on the same input under identical masking. This is the expected scaling behavior, so we do not regard model size as a limitation \emph{per se}, although the largest variants inherit the standard ViT pre-training cost.

\noindent\textbf{Beyond hyperparameters: Canny-edge dependence.} The SGFT salience score is computed from a Canny edge map, so the quality of the tree decomposition depends on whether edges faithfully reflect task-relevant micro-structure. On images with weak gradients, dominant noise, or texture-defined (rather than edge-defined) regions, the tree may over-tokenize uninformative areas and under-tokenize the true regions of interest, which would bound \method{}'s gains over uniform patching.

\section{Conclusion}
\chen{
We presented \method{}, a novel and effective framework that makes MAE-style pre-training feasible for ultra-high resolution scientific images. \chen{Using structure-guided masked autoencoders,} \method{} avoids the $O(N^2)$ bottleneck and preserves important fine-scale structures. Across TEM, XCT, and WSI datasets, \method{} achieves consistently better segmentation accuracy than MAE, including gains of +16.84\% Dice on 32K pathology and up to $24.8\times$ faster inference. \method{} also enables reliable zero-shot analysis, such as pore-network extraction and hydrogel skeletonization.
}

\begin{ack}
This work was supported by JSPS Program for Forming Japan's Peak Research Universities (J-PEAKS) Grant Number JPJS00420230001.
\end{ack}

{
    \small
    \bibliographystyle{plainnat}
    \bibliography{main}

\begin{thebibliography}{69}
\providecommand{\natexlab}[1]{#1}
\providecommand{\url}[1]{\texttt{#1}}
\expandafter\ifx\csname urlstyle\endcsname\relax
  \providecommand{\doi}[1]{doi: #1}\else
  \providecommand{\doi}{doi: \begingroup \urlstyle{rm}\Url}\fi

\bibitem[Ainslie et~al.(2020)Ainslie, Ontanon, Alberti, Cvicek, Fisher, Pham,
  Ravula, Sanghai, Wang, and Yang]{ainslie2020etc}
Joshua Ainslie, Santiago Ontanon, Chris Alberti, Vaclav Cvicek, Zachary Fisher,
  Philip Pham, Anirudh Ravula, Sumit Sanghai, Qifan Wang, and Li~Yang.
\newblock Etc: Encoding long and structured inputs in transformers.
\newblock \emph{arXiv preprint arXiv:2004.08483}, 2020.

\bibitem[Beltagy et~al.(2020)Beltagy, Peters, and Cohan]{beltagy2020longformer}
Iz~Beltagy, Matthew~E Peters, and Arman Cohan.
\newblock Longformer: The long-document transformer.
\newblock \emph{arXiv preprint arXiv:2004.05150}, 2020.

\bibitem[Bo et~al.(2023)Bo, Shi, Wang, and Liao]{bo2023specformer}
Deyu Bo, Chuan Shi, Lele Wang, and Renjie Liao.
\newblock Specformer: Spectral graph neural networks meet transformers.
\newblock In \emph{The Eleventh International Conference on Learning
  Representations}, 2023.
\newblock URL \url{https://openreview.net/forum?id=0pdSt3oyJa1}.

\bibitem[Buzug(2011)]{buzug2011computed}
Thorsten~M Buzug.
\newblock Computed tomography.
\newblock In \emph{Springer handbook of medical technology}, pages 311--342.
  Springer, 2011.

\bibitem[Cao et~al.(2022)Cao, Wang, Chen, Jiang, Zhang, Tian, and
  Wang]{cao2022swin}
Hu~Cao, Yueyue Wang, Joy Chen, Dongsheng Jiang, Xiaopeng Zhang, Qi~Tian, and
  Manning Wang.
\newblock Swin-unet: Unet-like pure transformer for medical image segmentation.
\newblock In \emph{European conference on computer vision}, pages 205--218.
  Springer, 2022.

\bibitem[Chen et~al.(2021{\natexlab{a}})Chen, Fan, and Panda]{chen2021crossvit}
Chun-Fu~Richard Chen, Quanfu Fan, and Rameswar Panda.
\newblock Crossvit: Cross-attention multi-scale vision transformer for image
  classification.
\newblock In \emph{Proceedings of the IEEE/CVF international conference on
  computer vision}, pages 357--366, New York, NY, USA, 2021{\natexlab{a}}.
  IEEE.

\bibitem[Chen et~al.(2021{\natexlab{b}})Chen, Lu, Yu, Luo, Adeli, Wang, Lu,
  Yuille, and Zhou]{chen2021transunet}
Jieneng Chen, Yongyi Lu, Qihang Yu, Xiangde Luo, Ehsan Adeli, Yan Wang, Le~Lu,
  Alan~L Yuille, and Yuyin Zhou.
\newblock Transunet: Transformers make strong encoders for medical image
  segmentation.
\newblock \emph{arXiv preprint arXiv:2102.04306}, 2021{\natexlab{b}}.

\bibitem[Chen et~al.(2022{\natexlab{a}})Chen, Chen, Li, Chen, Trister,
  Krishnan, and Mahmood]{Chen22}
Richard~J. Chen, Chengkuan Chen, Yicong Li, Tiffany~Y. Chen, Andrew~D. Trister,
  Rahul~G. Krishnan, and Faisal Mahmood.
\newblock Scaling vision transformers to gigapixel images via hierarchical
  self-supervised learning.
\newblock In \emph{2022 IEEE/CVF Conference on Computer Vision and Pattern
  Recognition (CVPR)}, pages 16123--16134, New York, NY, USA,
  2022{\natexlab{a}}. IEEE.
\newblock \doi{10.1109/CVPR52688.2022.01567}.

\bibitem[Chen et~al.(2022{\natexlab{b}})Chen, Chen, Li, Chen, Trister,
  Krishnan, and Mahmood]{chen2023scaling}
Richard~J Chen, Chengkuan Chen, Yicong Li, Tiffany~Y Chen, Andrew~D Trister,
  Rahul~G Krishnan, and Faisal Mahmood.
\newblock Scaling vision transformers to gigapixel images via hierarchical
  self-supervised learning.
\newblock In \emph{Proceedings of the IEEE/CVF Conference on Computer Vision
  and Pattern Recognition (CVPR)}, pages 16144--16155, 2022{\natexlab{b}}.

\bibitem[Chen et~al.(2020)Chen, Kornblith, Norouzi, and Hinton]{chen2020simple}
Ting Chen, Simon Kornblith, Mohammad Norouzi, and Geoffrey Hinton.
\newblock A simple framework for contrastive learning of visual
  representations.
\newblock In \emph{International conference on machine learning}, pages
  1597--1607. PmLR, 2020.

\bibitem[Child et~al.(2019{\natexlab{a}})Child, Gray, Radford, and
  Sutskever]{child2019generating}
Rewon Child, Scott Gray, Alec Radford, and Ilya Sutskever.
\newblock Generating long sequences with sparse transformers.
\newblock \emph{arXiv preprint arXiv:1904.10509}, 2019{\natexlab{a}}.

\bibitem[Child et~al.(2019{\natexlab{b}})Child, Gray, Radford, and
  Sutskever]{sparse-transformer19}
Rewon Child, Scott Gray, Alec Radford, and Ilya Sutskever.
\newblock Generating long sequences with sparse transformers,
  2019{\natexlab{b}}.
\newblock URL \url{https://arxiv.org/abs/1904.10509}.

\bibitem[Choromanski et~al.(2020)Choromanski, Likhosherstov, Dohan, Song, Gane,
  Sarlos, Hawkins, Davis, Mohiuddin, Kaiser, et~al.]{choromanski2020rethinking}
Krzysztof Choromanski, Valerii Likhosherstov, David Dohan, Xingyou Song,
  Andreea Gane, Tamas Sarlos, Peter Hawkins, Jared Davis, Afroz Mohiuddin,
  Lukasz Kaiser, et~al.
\newblock Rethinking attention with performers.
\newblock \emph{arXiv preprint arXiv:2009.14794}, 2020.

\bibitem[Cnudde and Boone(2013)]{cnudde2013high}
Veerle Cnudde and Matthieu~Nicolaas Boone.
\newblock High-resolution x-ray computed tomography in geosciences: A review of
  the current technology and applications.
\newblock \emph{Earth-Science Reviews}, 123:\penalty0 1--17, 2013.

\bibitem[Dao(2023)]{dao2023flashattention2}
Tri Dao.
\newblock Flashattention-2: Faster attention with better parallelism and work
  partitioning, 2023.

\bibitem[Dao et~al.(2022{\natexlab{a}})Dao, Chen, Sohoni, Desai, Poli, Grogan,
  Liu, Rao, Rudra, and Re]{pmlr-v162-dao22a}
Tri Dao, Beidi Chen, Nimit~S Sohoni, Arjun Desai, Michael Poli, Jessica Grogan,
  Alexander Liu, Aniruddh Rao, Atri Rudra, and Christopher Re.
\newblock Monarch: Expressive structured matrices for efficient and accurate
  training.
\newblock In Kamalika Chaudhuri, Stefanie Jegelka, Le~Song, Csaba Szepesvari,
  Gang Niu, and Sivan Sabato, editors, \emph{Proceedings of the 39th
  International Conference on Machine Learning}, volume 162 of
  \emph{Proceedings of Machine Learning Research}, pages 4690--4721. PMLR,
  17--23 Jul 2022{\natexlab{a}}.
\newblock URL \url{https://proceedings.mlr.press/v162/dao22a.html}.

\bibitem[Dao et~al.(2022{\natexlab{b}})Dao, Fu, Ermon, Rudra, and
  Ré]{FlashAttention22}
Tri Dao, Daniel~Y. Fu, Stefano Ermon, Atri Rudra, and Christopher Ré.
\newblock Flashattention: Fast and memory-efficient exact attention with
  io-awareness.
\newblock In \emph{NeurIPS: Proceedings of the 35th Neural Information
  Processing Systems Conference}, New York, NY, USA, 2022{\natexlab{b}}.
  Association for Computing Machinery.
\newblock \doi{10.48550/ARXIV.2205.14135}.
\newblock URL \url{https://arxiv.org/abs/2205.14135}.

\bibitem[Devlin et~al.(2019)Devlin, Chang, Lee, and Toutanova]{devlin2019bert}
Jacob Devlin, Ming-Wei Chang, Kenton Lee, and Kristina Toutanova.
\newblock Bert: Pre-training of deep bidirectional transformers for language
  understanding.
\newblock In \emph{Proceedings of the 2019 conference of the North American
  chapter of the association for computational linguistics: human language
  technologies, volume 1 (long and short papers)}, pages 4171--4186, 2019.

\bibitem[Dosovitskiy et~al.(2021)Dosovitskiy, Beyer, Kolesnikov, Weissenborn,
  Zhai, Unterthiner, Dehghani, Minderer, Heigold, Gelly, Uszkoreit, and
  Houlsby]{dosovitskiy2021image}
Alexey Dosovitskiy, Lucas Beyer, Alexander Kolesnikov, Dirk Weissenborn,
  Xiaohua Zhai, Thomas Unterthiner, Mostafa Dehghani, Matthias Minderer, Georg
  Heigold, Sylvain Gelly, Jakob Uszkoreit, and Neil Houlsby.
\newblock An image is worth 16x16 words: Transformers for image recognition at
  scale.
\newblock In \emph{International Conference on Learning Representations
  (ICLR)}, 2021.

\bibitem[Hatamizadeh et~al.(2022)Hatamizadeh, Tang, Nath, Yang, Myronenko,
  Landman, Roth, and Xu]{hatamizadeh2022unetr}
Ali Hatamizadeh, Yucheng Tang, Vishwesh Nath, Dong Yang, Andriy Myronenko,
  Bennett Landman, Holger~R Roth, and Daguang Xu.
\newblock Unetr: Transformers for 3d medical image segmentation.
\newblock In \emph{Proceedings of the IEEE/CVF winter conference on
  applications of computer vision}, pages 574--584, 2022.

\bibitem[He et~al.(2020)He, Fan, Wu, Xie, and Girshick]{he2020momentum}
Kaiming He, Haoqi Fan, Yuxin Wu, Saining Xie, and Ross Girshick.
\newblock Momentum contrast for unsupervised visual representation learning.
\newblock In \emph{Proceedings of the IEEE/CVF conference on computer vision
  and pattern recognition}, pages 9729--9738, 2020.

\bibitem[He et~al.(2022)He, Chen, Xie, Li, Doll{\'a}r, and
  Girshick]{he2022masked}
Kaiming He, Xinlei Chen, Saining Xie, Yanghao Li, Piotr Doll{\'a}r, and Ross
  Girshick.
\newblock Masked autoencoders are scalable vision learners.
\newblock In \emph{Proceedings of the IEEE/CVF conference on computer vision
  and pattern recognition}, pages 16000--16009, 2022.

\bibitem[He et~al.(2023)He, Nath, Yang, Tang, Myronenko, and
  Xu]{he2023swinunetrv2}
Yufan He, Vishwesh Nath, Dong Yang, Yucheng Tang, Andriy Myronenko, and Daguang
  Xu.
\newblock {SwinUNETR-V2}: Stronger swin transformers with stagewise
  convolutions for {3D} medical image segmentation.
\newblock In \emph{Medical Image Computing and Computer Assisted Intervention
  -- MICCAI 2023}, volume 14223 of \emph{Lecture Notes in Computer Science},
  pages 416--426. Springer, 2023.
\newblock \doi{10.1007/978-3-031-43901-8\_40}.

\bibitem[Heller et~al.(2021)Heller, Isensee, Maier-Hein, Hou, Xie, Li, Nan, Mu,
  Lin, Han, et~al.]{Heller2019kits19}
Nicholas Heller, Fabian Isensee, Klaus~H Maier-Hein, Xiaoshuai Hou, Chunmei
  Xie, Fengyi Li, Yang Nan, Guangrui Mu, Zhiyong Lin, Miofei Han, et~al.
\newblock The state of the art in kidney and kidney tumor segmentation in
  contrast-enhanced {CT} imaging: Results of the {KiTS19} challenge.
\newblock \emph{Medical Image Analysis}, 67:\penalty0 101821, 2021.
\newblock \doi{10.1016/j.media.2020.101821}.

\bibitem[Ibing et~al.(2023)Ibing, Kobsik, and Kobbelt]{ibing2023octree}
Moritz Ibing, Gregor Kobsik, and Leif Kobbelt.
\newblock Octree transformer: Autoregressive 3d shape generation on
  hierarchically structured sequences.
\newblock In \emph{Proceedings of the IEEE/CVF Conference on Computer Vision
  and Pattern Recognition}, pages 2697--2706, 2023.

\bibitem[Jacobs et~al.(2023)Jacobs, Tanaka, Zhang, Zhang, Song, Rajbhandari,
  and He]{jacobs2023deepspeed}
Sam~Ade Jacobs, Masahiro Tanaka, Chengming Zhang, Minjia Zhang, Shuaiwen~Leon
  Song, Samyam Rajbhandari, and Yuxiong He.
\newblock Deepspeed ulysses: System optimizations for enabling training of
  extreme long sequence transformer models, 2023.

\bibitem[Jain(2022)]{jain2022hugging}
Shashank~Mohan Jain.
\newblock Hugging face.
\newblock In \emph{Introduction to transformers for NLP: With the hugging face
  library and models to solve problems}, pages 51--67. Springer, 2022.

\bibitem[Katharopoulos et~al.(2020{\natexlab{a}})Katharopoulos, Vyas, Pappas,
  and Fleuret]{Katharopoulos20}
Angelos Katharopoulos, Apoorv Vyas, Nikolaos Pappas, and Fran\c{c}ois Fleuret.
\newblock Transformers are rnns: Fast autoregressive transformers with linear
  attention.
\newblock In \emph{Proceedings of the 37th International Conference on Machine
  Learning}, ICML'20, New York, NY, USA, 2020{\natexlab{a}}. Association for
  Computing Machinery.

\bibitem[Katharopoulos et~al.(2020{\natexlab{b}})Katharopoulos, Vyas, Pappas,
  and Fleuret]{katharopoulos2020transformers}
Angelos Katharopoulos, Apoorv Vyas, Nikolaos Pappas, and Fran{\c{c}}ois
  Fleuret.
\newblock Transformers are rnns: Fast autoregressive transformers with linear
  attention.
\newblock In \emph{International conference on machine learning}, pages
  5156--5165. PMLR, 2020{\natexlab{b}}.

\bibitem[Kim et~al.(2021)Kim, Jang, Lee, Park, Min, Hong, Park, Lee, Kim, Hong,
  Jung, Liu, Rajkumar, Khened, Krishnamurthi, Yang, Wang, Han, Kwak, Ma, Tang,
  Marami, Zeineh, Zhao, Heng, Schmitz, Madesta, Rösch, Werner, Tian,
  Puybareau, Bovio, Zhang, Zhu, Chun, Jeong, Park, and Choi]{KIM2021101854}
Yoo~Jung Kim, Hyungjoon Jang, Kyoungbun Lee, Seongkeun Park, Sung-Gyu Min,
  Choyeon Hong, Jeong~Hwan Park, Kanggeun Lee, Jisoo Kim, Wonjae Hong, Hyun
  Jung, Yanling Liu, Haran Rajkumar, Mahendra Khened, Ganapathy Krishnamurthi,
  Sen Yang, Xiyue Wang, Chang~Hee Han, Jin~Tae Kwak, Jianqiang Ma, Zhe Tang,
  Bahram Marami, Jack Zeineh, Zixu Zhao, Pheng-Ann Heng, Rüdiger Schmitz,
  Frederic Madesta, Thomas Rösch, Rene Werner, Jie Tian, Elodie Puybareau,
  Matteo Bovio, Xiufeng Zhang, Yifeng Zhu, Se~Young Chun, Won-Ki Jeong, Peom
  Park, and Jinwook Choi.
\newblock Paip 2019: Liver cancer segmentation challenge.
\newblock \emph{Medical Image Analysis}, 67:\penalty0 101854, 2021.
\newblock ISSN 1361-8415.
\newblock \doi{https://doi.org/10.1016/j.media.2020.101854}.
\newblock URL
  \url{https://www.sciencedirect.com/science/article/pii/S1361841520302188}.

\bibitem[Kirillov et~al.(2023)Kirillov, Mintun, Ravi, Mao, Rolland, Gustafson,
  Xiao, Whitehead, Berg, Lo, et~al.]{kirillov2023segment}
Alexander Kirillov, Eric Mintun, Nikhila Ravi, Hanzi Mao, Chloe Rolland, Laura
  Gustafson, Tete Xiao, Spencer Whitehead, Alexander~C Berg, Wan-Yen Lo, et~al.
\newblock Segment anything.
\newblock In \emph{Proceedings of the IEEE/CVF International Conference on
  Computer Vision}, pages 4015--4026, 2023.

\bibitem[Kitaev et~al.(2020{\natexlab{a}})Kitaev, Kaiser, and
  Levskaya]{kitaev2020reformer}
Nikita Kitaev, {\L}ukasz Kaiser, and Anselm Levskaya.
\newblock Reformer: The efficient transformer.
\newblock \emph{arXiv preprint arXiv:2001.04451}, 2020{\natexlab{a}}.

\bibitem[Kitaev et~al.(2020{\natexlab{b}})Kitaev, Kaiser, and
  Levskaya]{reformer20}
Nikita Kitaev, Lukasz Kaiser, and Anselm Levskaya.
\newblock Reformer: The efficient transformer.
\newblock In \emph{The International Conference on Learning Representations
  (ICLR)}, New York, NY, USA, 2020{\natexlab{b}}. Association for Computing
  Machinery.
\newblock \doi{10.48550/ARXIV.2001.04451}.
\newblock URL \url{https://arxiv.org/abs/2001.04451}.

\bibitem[Kiyama et~al.(2022)Kiyama, Yoshida, Nonoyama, Sedla{\v{c}}{\'i}k,
  Jinnai, Kurokawa, Nakajima, and Gong]{Kiyama2022Nanoscale}
Ryuji Kiyama, Masahiro Yoshida, Takayuki Nonoyama, Tom{\'a}{\v{s}}
  Sedla{\v{c}}{\'i}k, Hiroshi Jinnai, Takayuki Kurokawa, Tasuku Nakajima, and
  Jian~Ping Gong.
\newblock Nanoscale {TEM} imaging of hydrogel network architecture.
\newblock \emph{Advanced Materials}, 35\penalty0 (1):\penalty0 2208902, 2022.
\newblock \doi{10.1002/adma.202208902}.

\bibitem[Kreuzer et~al.(2021)Kreuzer, Beaini, Hamilton, L\'{e}tourneau, and
  Tossou]{NEURIPS2021_b4fd1d2c}
Devin Kreuzer, Dominique Beaini, Will Hamilton, Vincent L\'{e}tourneau, and
  Prudencio Tossou.
\newblock Rethinking graph transformers with spectral attention.
\newblock In M.~Ranzato, A.~Beygelzimer, Y.~Dauphin, P.S. Liang, and J.~Wortman
  Vaughan, editors, \emph{Advances in Neural Information Processing Systems},
  volume~34, pages 21618--21629. Curran Associates, Inc., 2021.
\newblock URL
  \url{https://proceedings.neurips.cc/paper_files/paper/2021/file/b4fd1d2cb085390fbbadae65e07876a7-Paper.pdf}.

\bibitem[Landman et~al.(2015)Landman, Xu, Igelsias, Styner, Langerak, and
  Klein]{landman2015miccai}
B~Landman, Z~Xu, J~Igelsias, M~Styner, T~Langerak, and A~Klein.
\newblock Miccai multi-atlas labeling beyond the cranial vault--workshop and
  challenge.
\newblock In \emph{Proc. MICCAI Multi-Atlas Labeling Beyond Cranial
  Vault—Workshop Challenge}, 2015.

\bibitem[Li et~al.(2023)Li, Shao, Xie, Xing, Gonzalez, Stoica, Ma, and
  Zhang]{li2023lightseq}
Dacheng Li, Rulin Shao, Anze Xie, Eric~P. Xing, Joseph~E. Gonzalez, Ion Stoica,
  Xuezhe Ma, and Hao Zhang.
\newblock Lightseq: Sequence level parallelism for distributed training of long
  context transformers, 2023.

\bibitem[Liu et~al.(2023)Liu, Zaharia, and Abbeel]{liu2023ring}
Hao Liu, Matei Zaharia, and Pieter Abbeel.
\newblock Ring attention with blockwise transformers for near-infinite context,
  2023.

\bibitem[Liu et~al.(2021)Liu, Lin, Cao, Hu, Wei, Zhang, Lin, and
  Guo]{liu2021swin}
Ze~Liu, Yutong Lin, Yue Cao, Han Hu, Yixuan Wei, Zheng Zhang, Stephen Lin, and
  Baining Guo.
\newblock Swin transformer: Hierarchical vision transformer using shifted
  windows.
\newblock In \emph{Proceedings of the IEEE/CVF international conference on
  computer vision}, pages 10012--10022, 2021.

\bibitem[Ma et~al.(2024)Ma, Li, and Wang]{ma2024u}
Jun Ma, Feifei Li, and Bo~Wang.
\newblock U-mamba: Enhancing long-range dependency for biomedical image
  segmentation.
\newblock \emph{arXiv preprint arXiv:2401.04722}, 2024.

\bibitem[Mei et~al.(2024)Mei, Chen, Yuille, and Xie]{mei2024spformer}
Jieru Mei, Liang-Chieh Chen, Alan Yuille, and Cihang Xie.
\newblock Spformer: Enhancing vision transformer with superpixel
  representation.
\newblock \emph{arXiv preprint arXiv:2401.02931}, 2024.

\bibitem[Noguchi et~al.(2024)Noguchi, Kiyama, Yoshida, Marsudi, Kashimura,
  Tadanaga, Gong, and Nonoyama]{noguchi2024real}
Shinji Noguchi, Ryuji Kiyama, Masahiro Yoshida, Maradhana~Agung Marsudi,
  Naohiro Kashimura, Kiyoharu Tadanaga, Jian~Ping Gong, and Takayuki Nonoyama.
\newblock Real-space visualization of charged polymer network of hydrogel by
  double network strategy and mineral staining.
\newblock \emph{Nano Letters}, 24\penalty0 (29):\penalty0 9088--9095, 2024.

\bibitem[Rabe and Staats(2022)]{rabe2022selfattention}
Markus~N. Rabe and Charles Staats.
\newblock Self-attention does not need $o(n^2)$ memory, 2022.

\bibitem[Ravi et~al.(2024)Ravi, Gabeur, Hu, Hu, Ryali, Ma, Khedr, R{\"a}dle,
  Rolland, Gustafson, Mintun, Pan, Alwala, Carion, Wu, Girshick, Doll{\'a}r,
  and Feichtenhofer]{ravi2024sam2}
Nikhila Ravi, Valentin Gabeur, Yuan-Ting Hu, Ronghang Hu, Chaitanya Ryali,
  Tengyu Ma, Haitham Khedr, Roman R{\"a}dle, Chloe Rolland, Laura Gustafson,
  Eric Mintun, Junting Pan, Kalyan~Vasudev Alwala, Nicolas Carion, Chao-Yuan
  Wu, Ross Girshick, Piotr Doll{\'a}r, and Christoph Feichtenhofer.
\newblock Sam 2: Segment anything in images and videos.
\newblock \emph{arXiv preprint arXiv:2408.00714}, 2024.
\newblock URL \url{https://arxiv.org/abs/2408.00714}.

\bibitem[Rong et~al.(2014)Rong, Li, Zhang, and Sun]{rong2014improved}
Weibin Rong, Zhanjing Li, Wei Zhang, and Lining Sun.
\newblock An improved canny edge detection algorithm.
\newblock In \emph{2014 IEEE international conference on mechatronics and
  automation}, pages 577--582. IEEE, 2014.

\bibitem[Ronneberger et~al.(2015)Ronneberger, Fischer, and
  Brox]{ronneberger2015u}
Olaf Ronneberger, Philipp Fischer, and Thomas Brox.
\newblock U-net: Convolutional networks for biomedical image segmentation.
\newblock In \emph{Medical image computing and computer-assisted
  intervention--MICCAI 2015: 18th international conference, Munich, Germany,
  October 5-9, 2015, proceedings, part III 18}, pages 234--241. Springer, 2015.

\bibitem[Roy et~al.(2021)Roy, Saffar, Vaswani, and Grangier]{Roy21}
Aurko Roy, Mohammad Saffar, Ashish Vaswani, and David Grangier.
\newblock {Efficient Content-Based Sparse Attention with Routing Transformers}.
\newblock \emph{Transactions of the Association for Computational Linguistics},
  9:\penalty0 53--68, 02 2021.
\newblock ISSN 2307-387X.
\newblock \doi{10.1162/tacl_a_00353}.
\newblock URL \url{https://doi.org/10.1162/tacl\_a\_00353}.

\bibitem[Ryali et~al.(2023)Ryali, Hu, Bolya, Wei, Fan, Huang, Aggarwal,
  Chowdhury, Poursaeed, Hoffman, et~al.]{ryali2023hiera}
Chaitanya Ryali, Yuan-Ting Hu, Daniel Bolya, Chen Wei, Haoqi Fan, Po-Yao Huang,
  Vaibhav Aggarwal, Arkabandhu Chowdhury, Omid Poursaeed, Judy Hoffman, et~al.
\newblock Hiera: A hierarchical vision transformer without the
  bells-and-whistles.
\newblock In \emph{International Conference on Machine Learning}, pages
  29441--29454. PMLR, 2023.

\bibitem[Shepp and Logan(1974)]{shepp1974fourier}
Lawrence~A Shepp and Benjamin~F Logan.
\newblock The fourier reconstruction of a head section.
\newblock \emph{IEEE Transactions on nuclear science}, 21\penalty0
  (3):\penalty0 21--43, 1974.

\bibitem[Shi et~al.(2020)Shi, Subramanian, Cao, Demehri, Siewerdsen, and
  Zbijewski]{shi2020application}
Gengxin Shi, Shalini Subramanian, Qian Cao, Shadpour Demehri, Jeffrey~H
  Siewerdsen, and Wojciech Zbijewski.
\newblock Application of a novel ultra-high resolution multi-detector ct in
  quantitative imaging of trabecular microstructure.
\newblock In \emph{Medical Imaging 2020: Biomedical Applications in Molecular,
  Structural, and Functional Imaging}, volume 11317, pages 356--362. SPIE,
  2020.

\bibitem[Shi et~al.(2021)Shi, Gao, Ren, Xu, Liang, Li, and Kwok]{Shi21}
Han Shi, Jiahui Gao, Xiaozhe Ren, Hang Xu, Xiaodan Liang, Zhenguo Li, and
  James~T. Kwok.
\newblock Sparsebert: Rethinking the importance analysis in self-attention.
\newblock In \emph{Proceedings of the 38th International Conference on Machine
  Learning}, volume 139 of \emph{Proceedings of Machine Learning Research},
  pages 9547--9557, New York, NY, USA, 2021. {PMLR}.
\newblock URL \url{http://proceedings.mlr.press/v139/shi21a.html}.

\bibitem[Si and Roberts(2021)]{Si21}
Yuqi Si and Kirk Roberts.
\newblock Three-level hierarchical transformer networks for long-sequence and
  multiple clinical documents classification, 2021.
\newblock URL \url{https://arxiv.org/abs/2104.08444}.

\bibitem[Takeuchi et~al.(2024)Takeuchi, Uesugi, Sada, and
  Uesugi]{10.1117/12.3027913}
Akihisa Takeuchi, Masayuki Uesugi, Yuki Sada, and Kentaro Uesugi.
\newblock {Hierarchical three-dimensional imaging using x-ray
  micro-/nano-tomography at BL20XU of SPring-8}.
\newblock In Bert M{\"u}ller and Ge~Wang, editors, \emph{Developments in X-Ray
  Tomography XV}, volume 13152, page 131521R. International Society for Optics
  and Photonics, SPIE, 2024.
\newblock \doi{10.1117/12.3027913}.
\newblock URL \url{https://doi.org/10.1117/12.3027913}.

\bibitem[Tang et~al.(2022{\natexlab{a}})Tang, Zhang, Zhu, and
  Tan]{tang2022quadtree}
Shitao Tang, Jiahui Zhang, Siyu Zhu, and Ping Tan.
\newblock Quadtree attention for vision transformers.
\newblock \emph{arXiv preprint arXiv:2201.02767}, 2022{\natexlab{a}}.

\bibitem[Tang et~al.(2022{\natexlab{b}})Tang, Yang, Li, Roth, Landman, Xu,
  Nath, and Hatamizadeh]{DBLP:conf/cvpr/TangY0RLXNH22}
Yucheng Tang, Dong Yang, Wenqi Li, Holger~R. Roth, Bennett~A. Landman, Daguang
  Xu, Vishwesh Nath, and Ali Hatamizadeh.
\newblock Self-supervised pre-training of swin transformers for 3d medical
  image analysis.
\newblock In \emph{{CVPR}}, pages 20698--20708. {IEEE}, 2022{\natexlab{b}}.

\bibitem[Van~Aarle et~al.(2016)Van~Aarle, Palenstijn, Cant, Janssens,
  Bleichrodt, Dabravolski, De~Beenhouwer, Joost~Batenburg, and
  Sijbers]{van2016fast}
Wim Van~Aarle, Willem~Jan Palenstijn, Jeroen Cant, Eline Janssens, Folkert
  Bleichrodt, Andrei Dabravolski, Jan De~Beenhouwer, K~Joost~Batenburg, and Jan
  Sijbers.
\newblock Fast and flexible x-ray tomography using the astra toolbox.
\newblock \emph{Optics express}, 24\penalty0 (22):\penalty0 25129--25147, 2016.

\bibitem[Wang et~al.(2023)Wang, Lyngaas, Tsaris, Chen, Dash, Shekar, Luo, Yoon,
  Wahib, and Gouley]{wang2023ultralong}
Xiao Wang, Isaac Lyngaas, Aristeidis Tsaris, Peng Chen, Sajal Dash,
  Mayanka~Chandra Shekar, Tao Luo, Hong-Jun Yoon, Mohamed Wahib, and John
  Gouley.
\newblock Ultra-long sequence distributed transformer, 2023.
\newblock URL \url{https://arxiv.org/abs/2311.02382}.

\bibitem[Wang et~al.(2025)Wang, Choi, Kurihaya, Lyngaas, Yoon, Xiao, Pugmire,
  Fan, Nafi, Tsaris, et~al.]{wang2025orbit}
Xiao Wang, Jong-Youl Choi, Takuya Kurihaya, Isaac Lyngaas, Hong-Jun Yoon,
  Xi~Xiao, David Pugmire, Ming Fan, Nasik~M Nafi, Aristeidis Tsaris, et~al.
\newblock Orbit-2: Scaling exascale vision foundation models for weather and
  climate downscaling.
\newblock \emph{arXiv preprint arXiv:2505.04802}, 2025.

\bibitem[Wei et~al.(2025)Wei, Sun, and Li]{wei2025deepseek}
Haoran Wei, Yaofeng Sun, and Yukun Li.
\newblock Deepseek-ocr: Contexts optical compression.
\newblock \emph{arXiv preprint arXiv:2510.18234}, 2025.

\bibitem[Xie et~al.(2021)Xie, Zhang, Shen, and Xia]{xie2021cotr}
Yutong Xie, Jianpeng Zhang, Chunhua Shen, and Yong Xia.
\newblock {CoTr}: Efficiently bridging {CNN} and transformer for {3D} medical
  image segmentation.
\newblock In \emph{Medical Image Computing and Computer Assisted Intervention
  -- MICCAI 2021}, pages 171--180. Springer, 2021.
\newblock \doi{10.1007/978-3-030-87199-4\_16}.

\bibitem[Ying et~al.(2021)Ying, Ke, He, and Liu]{ying2021lazyformer}
Chengxuan Ying, Guolin Ke, Di~He, and Tie-Yan Liu.
\newblock Lazyformer: Self attention with lazy update, 2021.

\bibitem[Yu et~al.(2023)Yu, Simig, Flaherty, Aghajanyan, Zettlemoyer, and
  Lewis]{yu2023megabyte}
Lili Yu, Dániel Simig, Colin Flaherty, Armen Aghajanyan, Luke Zettlemoyer, and
  Mike Lewis.
\newblock Megabyte: Predicting million-byte sequences with multiscale
  transformers, 2023.

\bibitem[Zaheer et~al.(2020{\natexlab{a}})Zaheer, Guruganesh, Dubey, Ainslie,
  Alberti, Ontanon, Pham, Ravula, Wang, Yang, et~al.]{bigbird20}
Manzil Zaheer, Guru Guruganesh, Kumar~Avinava Dubey, Joshua Ainslie, Chris
  Alberti, Santiago Ontanon, Philip Pham, Anirudh Ravula, Qifan Wang, Li~Yang,
  et~al.
\newblock Big bird: Transformers for longer sequences.
\newblock \emph{Advances in Neural Information Processing Systems},
  33:\penalty0 17283--17297, 2020{\natexlab{a}}.

\bibitem[Zaheer et~al.(2020{\natexlab{b}})Zaheer, Guruganesh, Dubey, Ainslie,
  Alberti, Ontanon, Pham, Ravula, Wang, Yang, et~al.]{zaheer2020big}
Manzil Zaheer, Guru Guruganesh, Kumar~Avinava Dubey, Joshua Ainslie, Chris
  Alberti, Santiago Ontanon, Philip Pham, Anirudh Ravula, Qifan Wang, Li~Yang,
  et~al.
\newblock Big bird: Transformers for longer sequences.
\newblock \emph{Advances in neural information processing systems},
  33:\penalty0 17283--17297, 2020{\natexlab{b}}.

\bibitem[Zhang et~al.(2024)Zhang, Lyngaas, Chen, Wang, Igarashi, Huo, Munetomo,
  and Wahib]{zhang2024adaptive}
Enzhi Zhang, Isaac Lyngaas, Peng Chen, Xiao Wang, Jun Igarashi, Yuankai Huo,
  Masaharu Munetomo, and Mohamed Wahib.
\newblock Adaptive patching for high-resolution image segmentation with
  transformers.
\newblock In \emph{SC24: International Conference for High Performance
  Computing, Networking, Storage and Analysis}, pages 1--16. IEEE, 2024.

\bibitem[Zhang et~al.(2025)Zhang, Chen, Zhong, Wu, Igarashi, Lyngaas, Wang,
  Munetomo, and Wahib]{zhang2025shf}
Enzhi Zhang, Peng Chen, Rui Zhong, Du~Wu, Jun Igarashi, Isaac Lyngaas, Xiao
  Wang, Masaharu Munetomo, and Mohamed Wahib.
\newblock Shf: Symmetrical hierarchical forest with pretrained vision
  transformer encoder for high-resolution medical segmentation.
\newblock In \emph{Advances in Neural Information Processing Systems}, 2025.

\bibitem[Zhang and
  Lui(2022)]{zhang2022topologypreservingsegmentationnetworkdeep}
Han Zhang and Lok~Ming Lui.
\newblock Topology-preserving segmentation network: A deep learning
  segmentation framework for connected component, 2022.
\newblock URL \url{https://arxiv.org/abs/2202.13331}.

\bibitem[Zhou et~al.(2023)Zhou, Guo, Zhang, Han, Yu, Wang, and
  Yu]{zhou2021nnformer}
Hong-Yu Zhou, Jiansen Guo, Yinghao Zhang, Xiaoguang Han, Lequan Yu, Liansheng
  Wang, and Yizhou Yu.
\newblock {nnFormer}: Volumetric medical image segmentation via a {3D}
  transformer.
\newblock \emph{IEEE Transactions on Image Processing}, 32:\penalty0
  4036--4045, 2023.
\newblock \doi{10.1109/TIP.2023.3293771}.

\bibitem[Ziabari and Alnaggar(2025)]{ziabari2025pycmg}
Amir Ziabari and Mohammed Alnaggar.
\newblock Pycmg-based simulation of volumetric concrete microstructure.
\newblock Technical report, Oak Ridge National Laboratory (ORNL), Oak Ridge, TN
  (United States). Oak~…, 2025.

\end{thebibliography}
}

\newpage
\appendix
\section{Supplementary Material}
\label{sec:supp}

\subsection{Broader Impacts}
\label{sec:supp_impact}
\method{} is a self-supervised pre-training framework targeted at ultra-high-resolution scientific images (electron microscopy, X-ray CT, and WSI pathology). On the positive side, it lowers the compute and annotation cost of analyzing giga-pixel scientific data, which can accelerate discovery in materials science, biomedical imaging, and structural biology, and reduce reliance on expensive expert annotation. We do not foresee direct negative societal impacts from the method itself: the framework operates on scientific images rather than personal or web-scraped content, does not generate synthetic media, and is not a generative or language model. The one downstream consideration we want to flag is medical: pathology datasets such as PAIP-32K contain images of patient tissue, and any clinical deployment of \method{}-derived models would require additional validation, regulatory clearance, and safeguards against over-reliance on automated segmentation. We release model weights and code (where applicable) under standard research-use terms to support reproducibility while keeping clinical decision-making with qualified practitioners.

\subsection{Hyperparameters and Implementation Details}
\label{sec:supp_hparams}

This section provides the full set of hyperparameters used in pre-training and fine-tuning. All experiments were conducted with PyTorch and DeepSpeed on NVIDIA H100 nodes.

\begin{figure}[h]
  \centering
  \begin{subfigure}[b]{0.24\linewidth}
    \includegraphics[width=\linewidth]{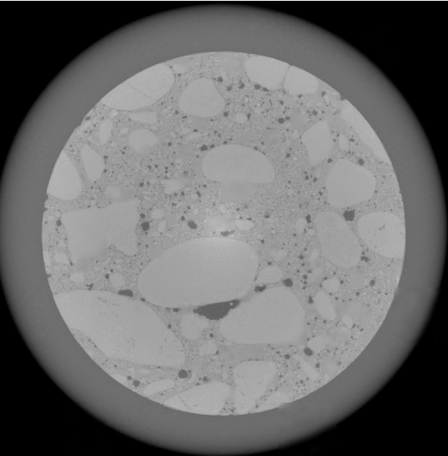}
    \caption*{Input}
  \end{subfigure}\hfill
  \begin{subfigure}[b]{0.24\linewidth}
    \includegraphics[width=\linewidth]{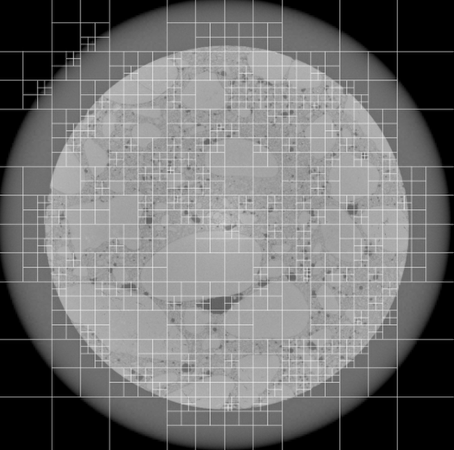}
    \caption*{Quadtree}
  \end{subfigure}\hfill
  \begin{subfigure}[b]{0.24\linewidth}
    \includegraphics[width=\linewidth]{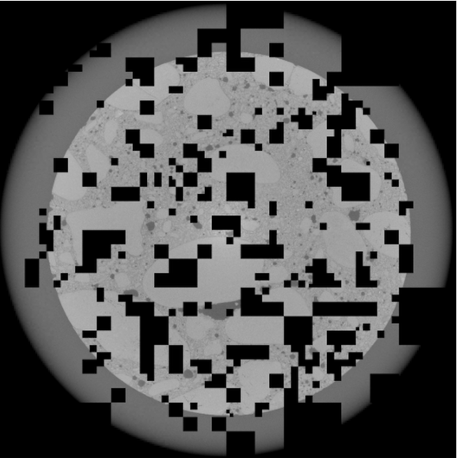}
    \caption*{Masked ($r{=}0.25$)}
  \end{subfigure}\hfill
  \begin{subfigure}[b]{0.24\linewidth}
    \includegraphics[width=\linewidth]{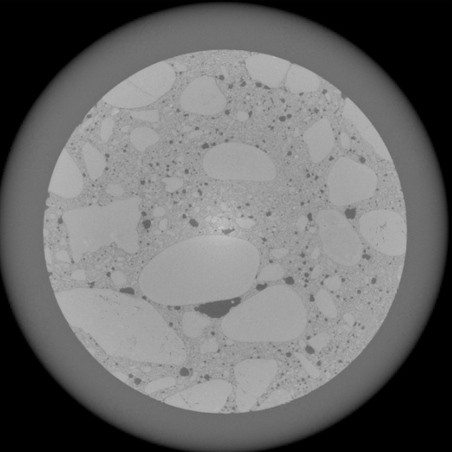}
    \caption*{Reconstruction}
  \end{subfigure}\\[2pt]
  \begin{subfigure}[b]{0.24\linewidth}
    \includegraphics[width=\linewidth]{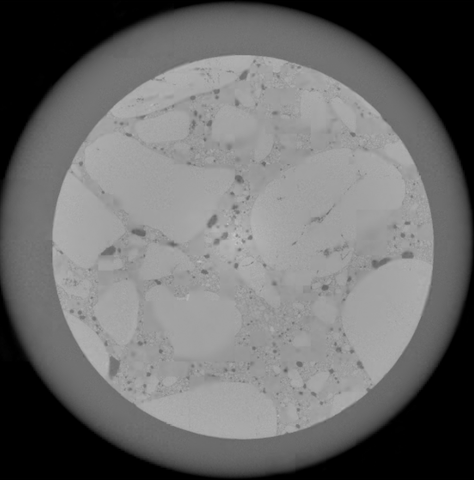}
    \caption*{Input}
  \end{subfigure}\hfill
  \begin{subfigure}[b]{0.24\linewidth}
    \includegraphics[width=\linewidth]{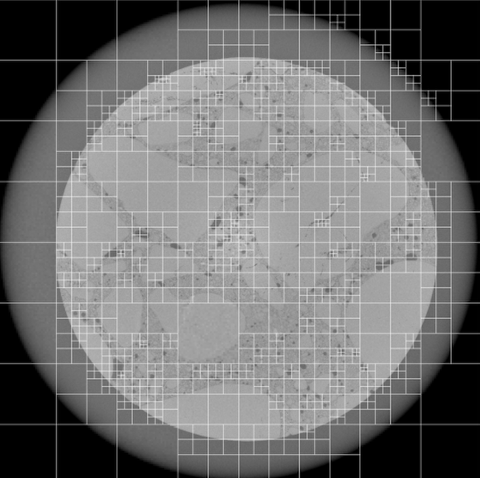}
    \caption*{Quadtree}
  \end{subfigure}\hfill
  \begin{subfigure}[b]{0.24\linewidth}
    \includegraphics[width=\linewidth]{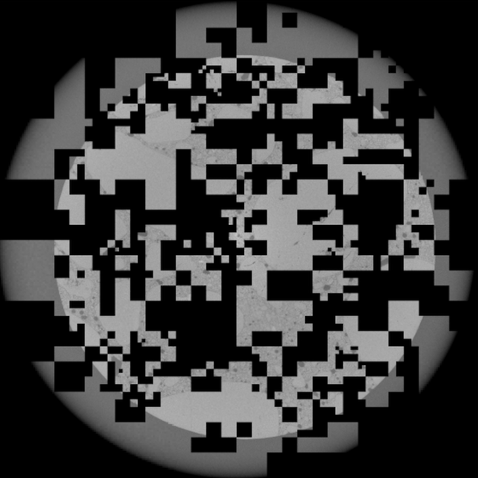}
    \caption*{Masked ($r{=}0.50$)}
  \end{subfigure}\hfill
  \begin{subfigure}[b]{0.24\linewidth}
    \includegraphics[width=\linewidth]{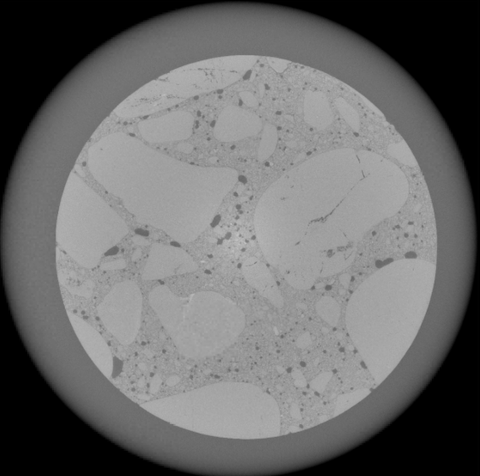}
    \caption*{Reconstruction}
  \end{subfigure}\\[2pt]
  \begin{subfigure}[b]{0.24\linewidth}
    \includegraphics[width=\linewidth]{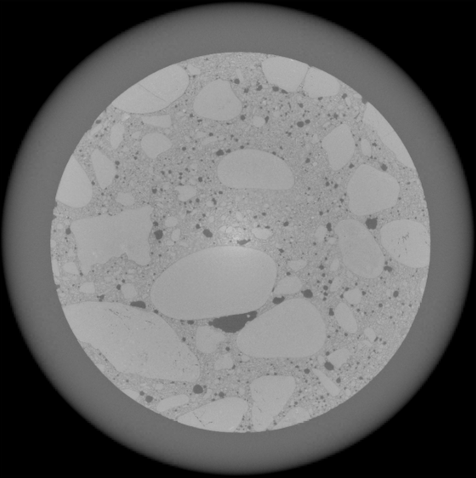}
    \caption*{Input}
  \end{subfigure}\hfill
  \begin{subfigure}[b]{0.24\linewidth}
    \includegraphics[width=\linewidth]{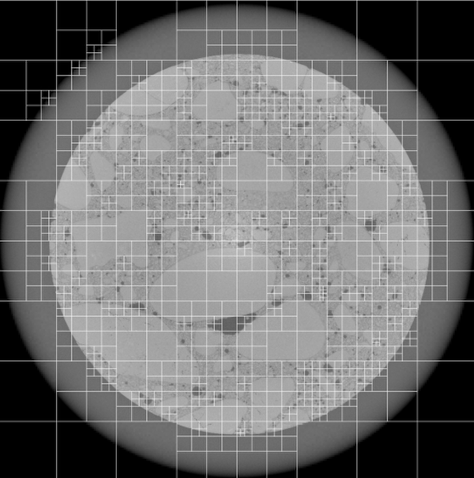}
    \caption*{Quadtree}
  \end{subfigure}\hfill
  \begin{subfigure}[b]{0.24\linewidth}
    \includegraphics[width=\linewidth]{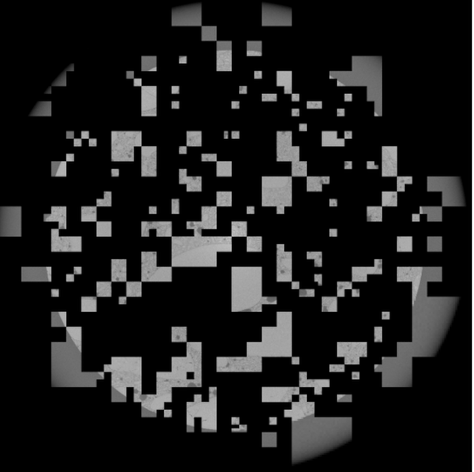}
    \caption*{Masked ($r{=}0.75$)}
  \end{subfigure}\hfill
  \begin{subfigure}[b]{0.24\linewidth}
    \includegraphics[width=\linewidth]{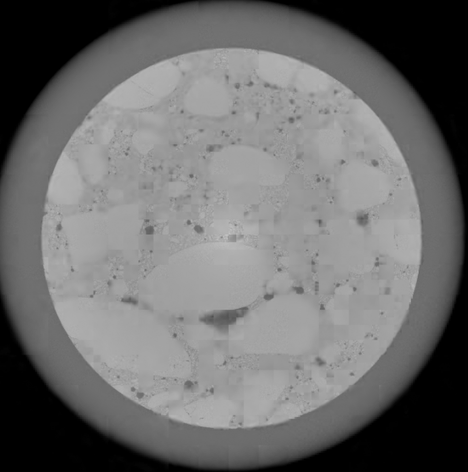}
    \caption*{Reconstruction}
  \end{subfigure}
  \caption{Masking ratio ablation. Each row fixes a ratio $r\in\{0.25, 0.50, 0.75\}$ and shows the input, the quadtree partition, the visible patches after structure-guided masking, and the \method{} reconstruction. Higher masking ratios yield a harder pre-training task while the adaptive quadtree keeps structurally important regions visible.}
  \label{fig:supp_ratio}
\end{figure}

\paragraph{Masking Ratio.}
For all \method{} pre-training experiments, we adopt a fixed masking ratio of $r=0.75$ over the adaptive structural patch sequence, matching the ratio recommended in MAE~\cite{he2022masked}. We additionally explored $r\in\{0.50, 0.60, 0.75, 0.85, 0.90\}$ on HydrogelTEM-1K and found $r=0.75$ to give the best Dice on the validation split. Unlike random MAE masking, our masks are sampled per-quadtree-level so that the ratio is preserved within each scale. Figure~\ref{fig:supp_ratio} visualizes the effect of sweeping $r\in\{0.25,0.50,0.75\}$ at the same sequence length: the adaptive quadtree produces consistent structural coverage, and the reconstruction quality degrades gracefully as the ratio grows.

\begin{figure}[h]
  \centering
  \begin{subfigure}[b]{0.24\linewidth}
    \includegraphics[width=\linewidth]{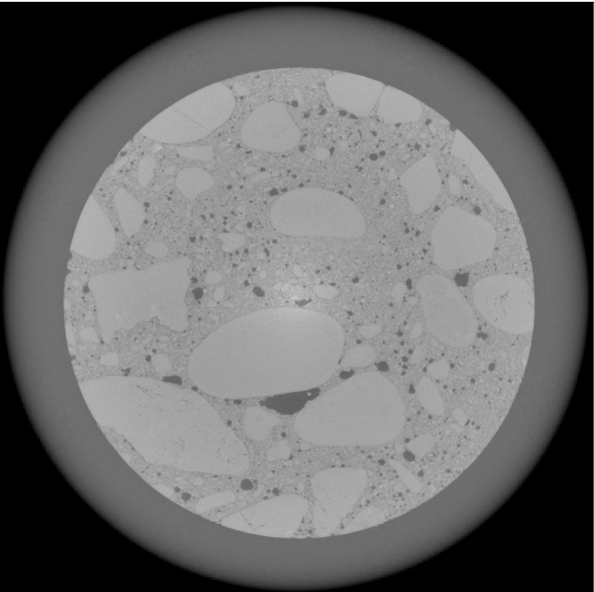}
    \caption*{Input}
  \end{subfigure}\hfill
  \begin{subfigure}[b]{0.24\linewidth}
    \includegraphics[width=\linewidth]{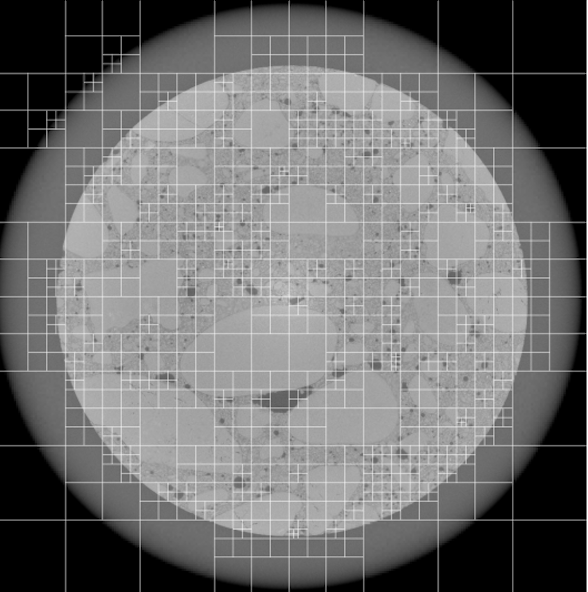}
    \caption*{$N{=}1024$}
  \end{subfigure}\hfill
  \begin{subfigure}[b]{0.24\linewidth}
    \includegraphics[width=\linewidth]{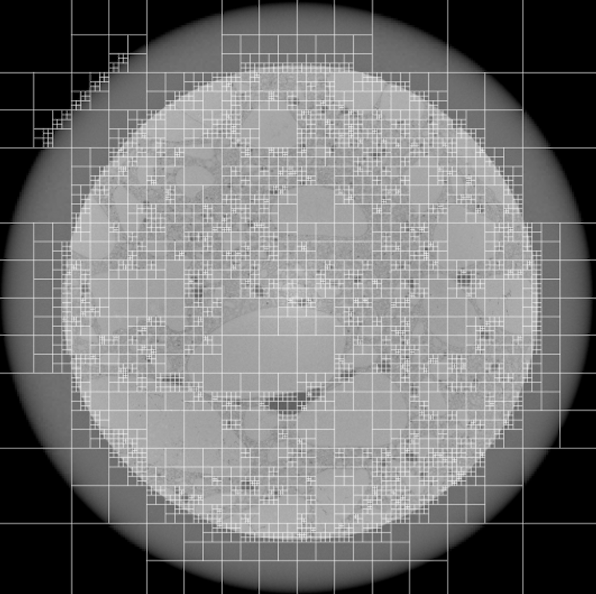}
    \caption*{$N{=}4096$}
  \end{subfigure}\hfill
  \begin{subfigure}[b]{0.24\linewidth}
    \includegraphics[width=\linewidth]{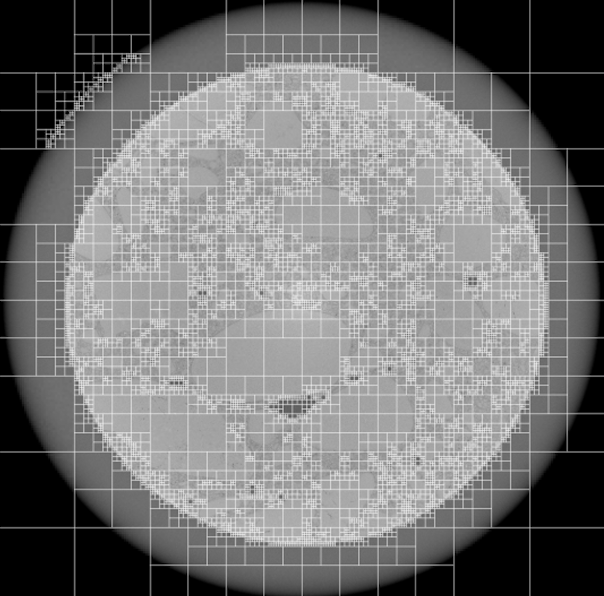}
    \caption*{$N{=}8192$}
  \end{subfigure}
  \caption{Sequence length ablation. The SGFT tokenizer concentrates more tokens in detail-rich regions as the budget $N$ grows from $1024$ to $8192$, while keeping uniform background coarsely partitioned.}
  \label{fig:supp_seq}
\end{figure}

\paragraph{Sequence Length.}
The fixed sequence length $N$ produced by the Structure-Guided Feature Tokenizer (SGFT) is the only hyperparameter that bounds the $O(N)$ tokenization budget. We use the following defaults per dataset:
\begin{itemize}[leftmargin=2.5mm]
    \item HydrogelTEM-1K ($1024^2$): $N=8{,}194$ for fine-tuning; $N=16{,}384$ for the high-fidelity \method{}-SAM/SAM~2 variants.
    \item SpringXCT-8K ($8192^2$): $N=1{,}024$ for the speed-optimized run; $N=8{,}194$ and $N=16{,}384$ for the highest-accuracy runs.
    \item PAIP-32K ($32{,}768^2$): $N=2{,}048$ for the speed-optimized run; up to $N=16{,}384$ for the best Dice runs.
\end{itemize}
The SGFT temperature is fixed at $\tau=1.0$. We did not observe meaningful changes for $\tau\in[0.5, 2.0]$. Figure~\ref{fig:supp_seq} visualizes the quadtree tokenization at $N\in\{1024, 4096, 8192\}$ on the same input: as $N$ grows, the tree spends more tokens on detail-rich regions, while uniform regions remain coarsely tokenized.

\begin{figure}[h]
  \centering
  \begin{subfigure}[b]{0.24\linewidth}
    \includegraphics[width=\linewidth]{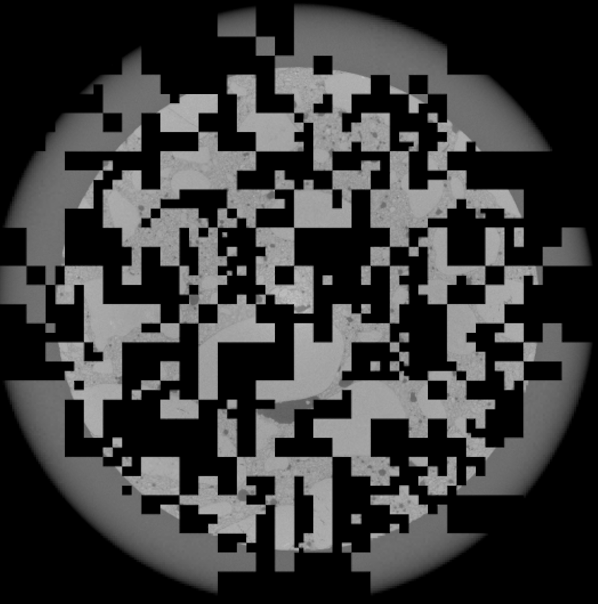}
    \caption*{Input}
  \end{subfigure}\hfill
  \begin{subfigure}[b]{0.24\linewidth}
    \includegraphics[width=\linewidth]{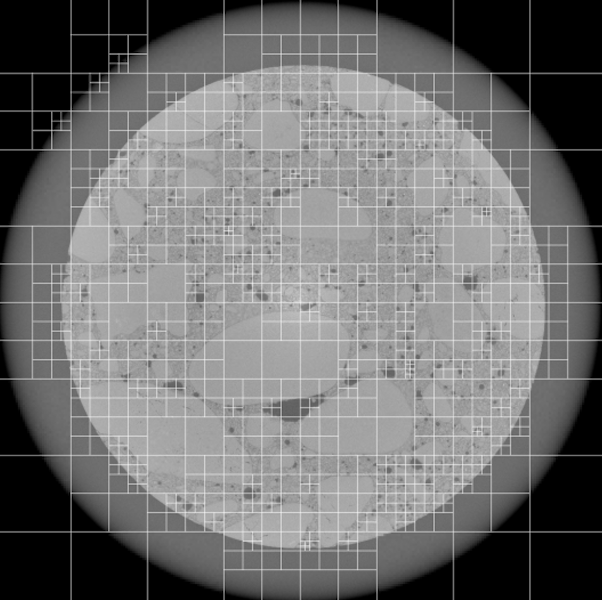}
    \caption*{Quadtree grid}
  \end{subfigure}\hfill
  \begin{subfigure}[b]{0.24\linewidth}
    \includegraphics[width=\linewidth]{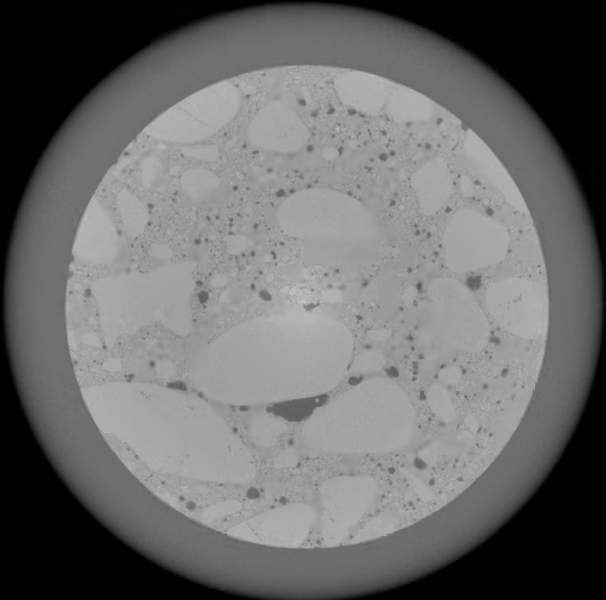}
    \caption*{Reconstruction (ViT-L)}
  \end{subfigure}\hfill
  \begin{subfigure}[b]{0.24\linewidth}
    \includegraphics[width=\linewidth]{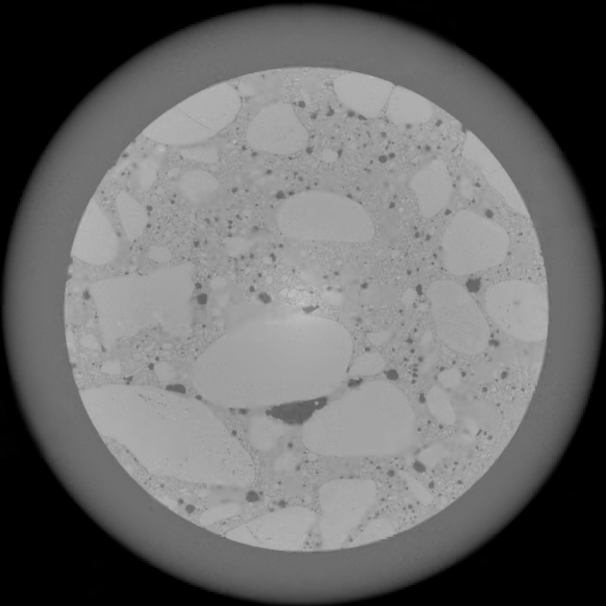}
    \caption*{Reconstruction (ViT-XL)}
  \end{subfigure}
  \caption{Model size ablation. Reconstruction quality improves from ViT-L to ViT-XL on a HydrogelTEM-1K patch under the same structure-guided masking and quadtree tokenization. The larger backbone recovers finer network topology, which motivates our use of larger backbones on higher-resolution datasets.}
  \label{fig:supp_model}
\end{figure}

\paragraph{Model Size.}
\method{} reuses the standard ViT backbones to keep comparisons fair:
\begin{itemize}[leftmargin=2.5mm]
    \item \emph{Encoder}: ViT-B/16 (12 layers, 768 hidden, 12 heads, 86M params) for HydrogelTEM-1K and PAIP-32K; ViT-L/16 (24 layers, 1024 hidden, 16 heads, 304M params) for SpringXCT-8K.
    \item \emph{Decoder} (pre-training only): 8 transformer blocks, 512 hidden, 16 heads (same as MAE~\cite{he2022masked}), discarded after pre-training.
    \item \emph{Segmentation heads}: SAM/SAM~2 default decoders for the SAM variants; the symmetric depatching head from~\cite{zhang2025shf} for our SGMA-SAM variants; the standard UNETR~\cite{hatamizadeh2022unetr} head for UNETR variants.
\end{itemize}
Figure~\ref{fig:supp_model} shows MAE-style reconstructions on a HydrogelTEM-1K patch using ViT-L vs.\ ViT-XL encoders, alongside the corresponding quadtree grid. The larger backbone recovers fine-grained network topology that the smaller variant blurs, motivating our use of ViT-L for the highest-resolution datasets.

\paragraph{Optimization.}
AdamW ($\beta_1{=}0.9$, $\beta_2{=}0.95$, weight decay $0.05$) with a cosine schedule, $40$ warmup epochs, base learning rate $1.5\!\times\!10^{-4}$ scaled linearly with the effective batch size. Pre-training runs $400$ epochs; fine-tuning runs $100$ epochs with a $10\times$ smaller learning rate. Mixed-precision (bf16) is used throughout.

\subsection{Zero-shot Downstream Tasks Evaluation}
\label{sec:supp_zeroshot}

\begin{figure}[h]
  \begin{center}
    \begin{subfigure}[t]{0.19\linewidth}
        \includegraphics[width=\linewidth,height=\linewidth]{figs/seg/gd.png}
        \caption*{\centering Ground Truth\\(Dice $100\%$)}
    \end{subfigure}\hfill
    \begin{subfigure}[t]{0.19\linewidth}
        \includegraphics[width=\linewidth,height=\linewidth]{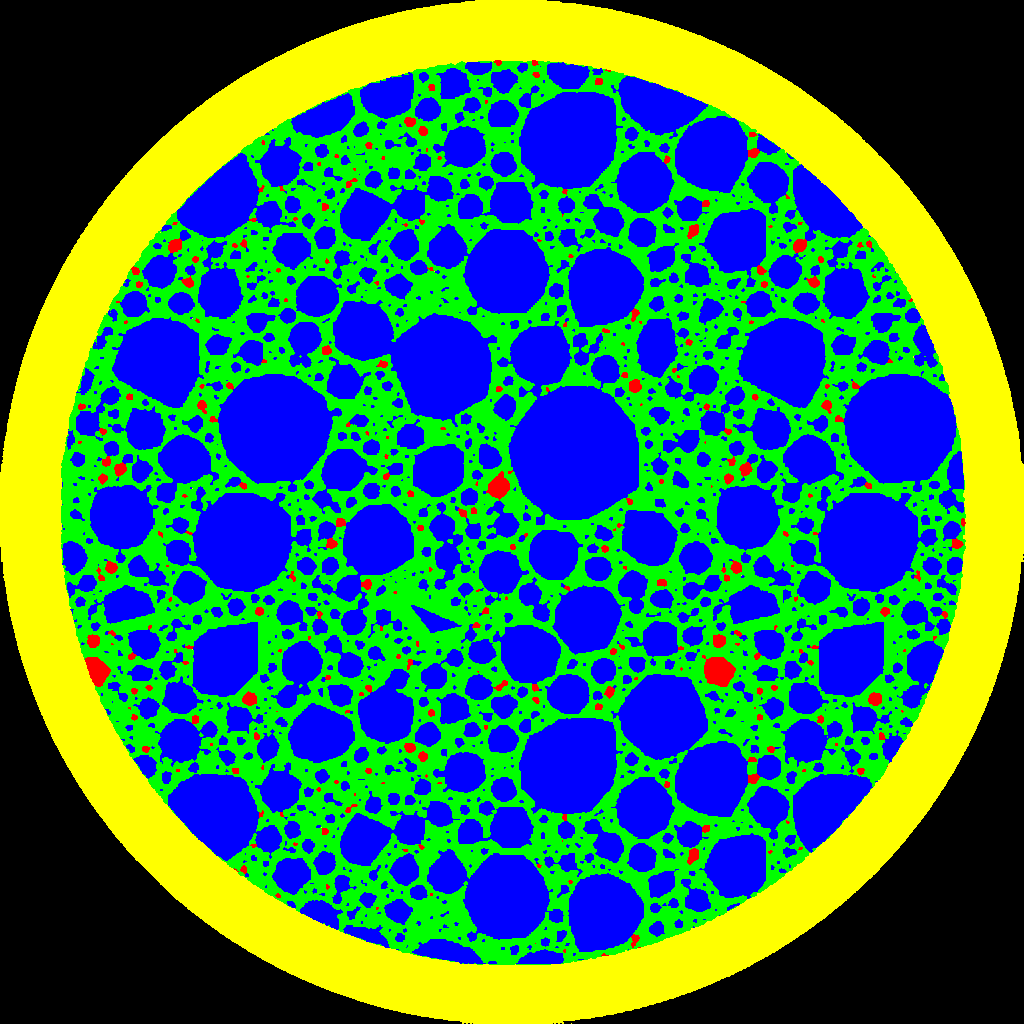}
        \caption*{\centering UNet~\cite{ronneberger2015u}\\(Dice $58.38\%$)}
    \end{subfigure}\hfill
    \begin{subfigure}[t]{0.19\linewidth}
        \includegraphics[width=\linewidth,height=\linewidth]{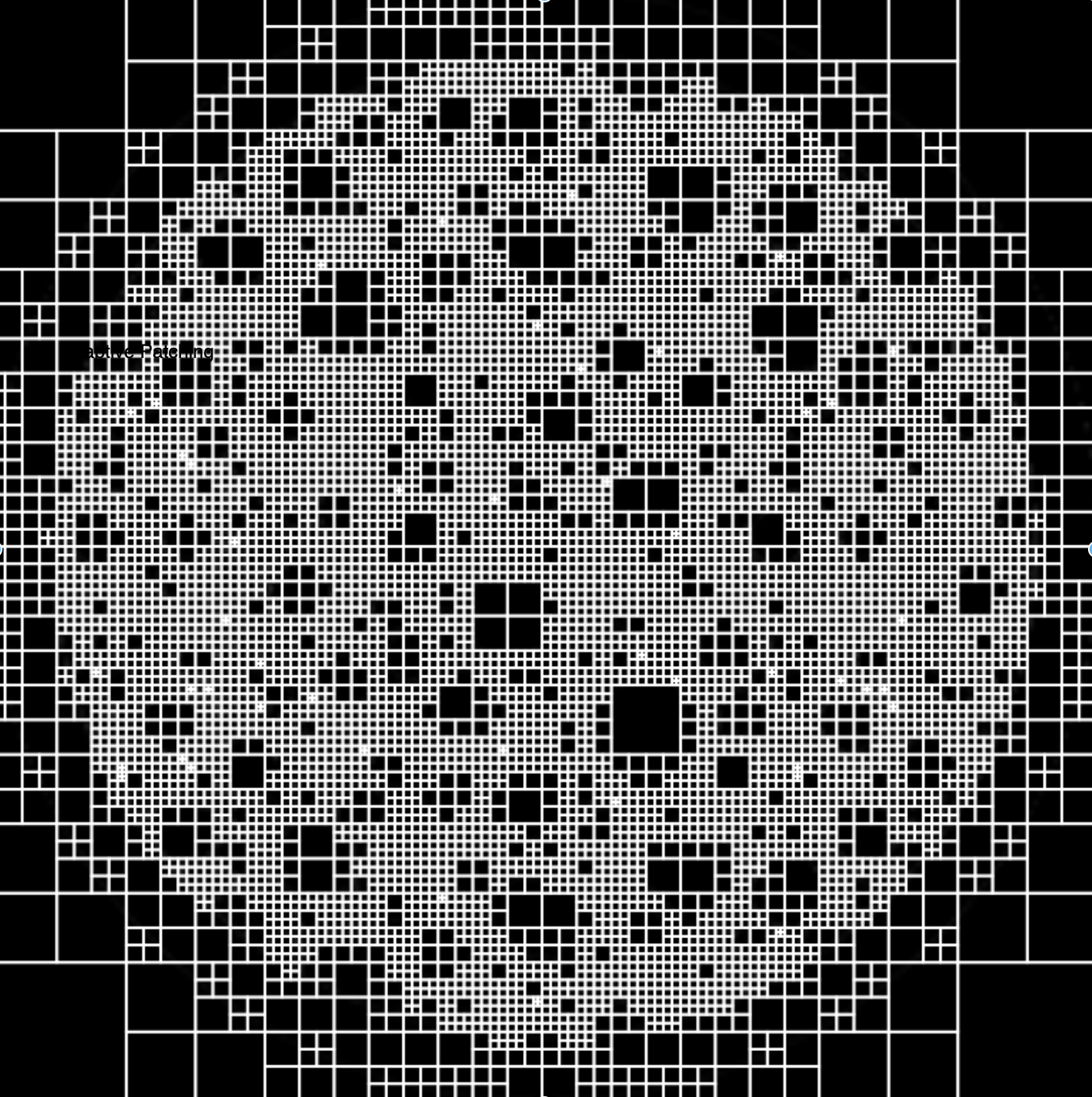}
        \caption*{\centering AP~\cite{zhang2024adaptive}\\($N{=}8{,}194$)}
    \end{subfigure}\hfill
    \begin{subfigure}[t]{0.19\linewidth}
        \includegraphics[width=\linewidth,height=\linewidth]{figs/seg/sam2.png}
        \caption*{\centering MAE-SAM~2~\cite{ravi2024sam2}\\(Dice $85.98\%$)}
    \end{subfigure}\hfill
    \begin{subfigure}[t]{0.19\linewidth}
        \includegraphics[width=\linewidth,height=\linewidth]{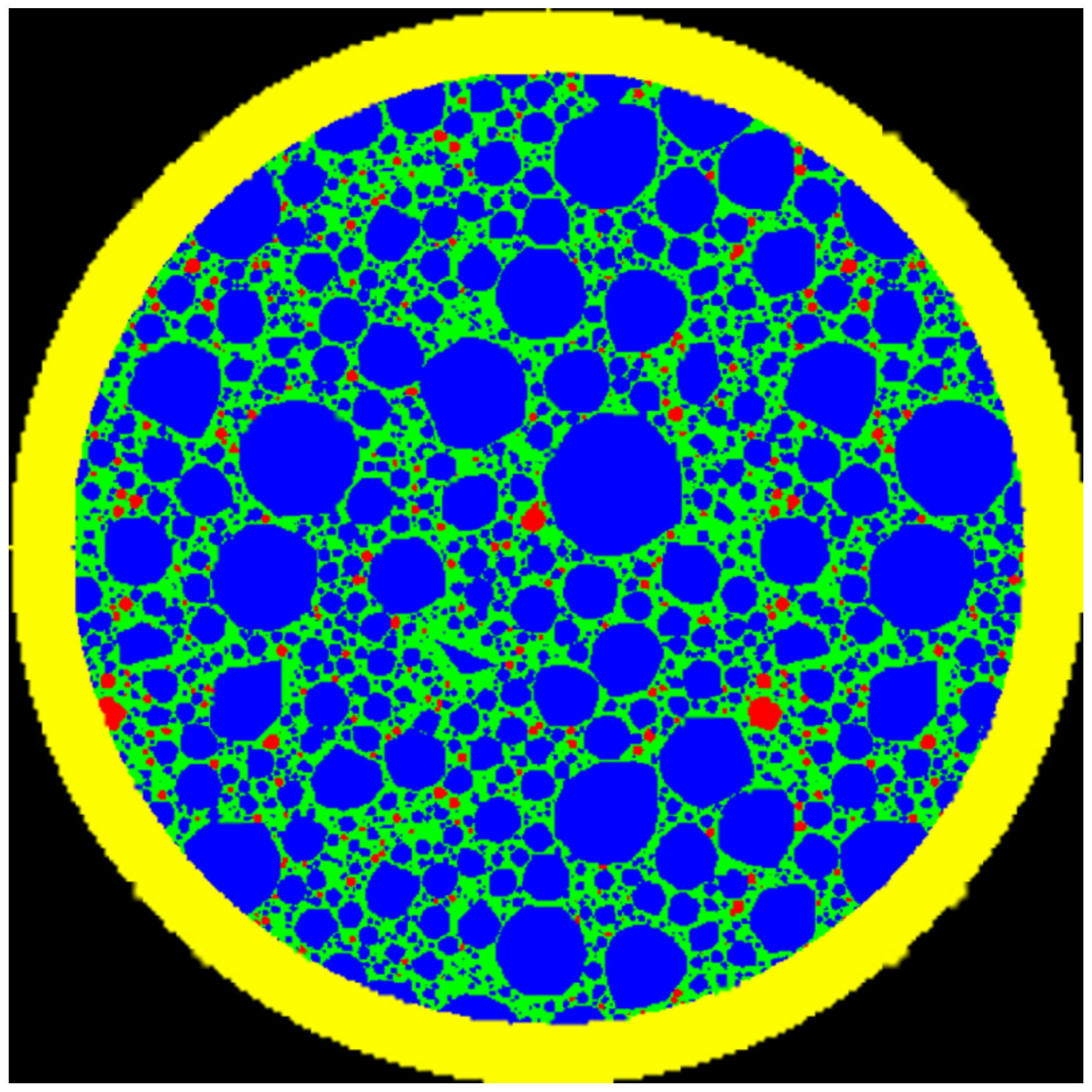}
        \caption*{\centering \textbf{\method{}-SAM~2 (ours)}\\($N{=}8{,}194$, Dice $94.79\%$)}
    \end{subfigure}
    \caption{Segmentation comparison on a SpringXCT-8K simulated slice (per-sample Dice reported). UNet~\cite{ronneberger2015u} loses small-pore topology; the Adaptive Patching baseline~\cite{zhang2024adaptive} at $N{=}8{,}194$ tokens improves but still drops thin channels; MAE-pretrained SAM~2~\cite{ravi2024sam2} at the smallest feasible $128^2$ patch on an 8K image captures coarse structure only; our \method{}-SAM~2 at a $2{\times}2$ effective patch size with $N{=}8{,}194$ SGFT tokens recovers the fine pore topology. The aggregate test-set Dice for each model is reported in Table~\ref{tab:combined_results}; the numbers here are for this specific sample.}
    \label{fig:seg_sim_s8d}
  \end{center}
\end{figure}

We use the zero-shot segmentation masks from the SpringXCT-8K and HydrogelTEM-1K datasets to conduct a statistical analysis of the components and connections. Before presenting the downstream statistics, we first visualize the per-sample segmentation quality that drives them. Figure~\ref{fig:seg_sim_s8d} compares four segmentation models on an identical simulated SpringXCT-8K slice (ground-truth mask available): the vanilla UNet~\cite{ronneberger2015u} collapses on small pores ($58.38\%$ Dice), the Adaptive Patching baseline~\cite{zhang2024adaptive} at $N{=}8{,}194$ tokens improves but still misses thin channels, the MAE-pretrained SAM~2~\cite{ravi2024sam2} at a $128{\times}128$ patch is forced to operate at coarse resolution by GPU memory ($85.98\%$ Dice), and our \method{}-SAM~2 at a $2{\times}2$ effective patch with $N{=}8{,}194$ SGFT tokens recovers the full pore topology ($94.79\%$ Dice). This side-by-side view is what our Table~\ref{tab:combined_results} Dice gains look like at the pixel level on a single sample, and it explains why the downstream statistics in the next paragraphs are only reliable for \method{}-derived masks.

\begin{figure}[h]
  \begin{center}
    \begin{subfigure}[b]{0.48\linewidth}
        \includegraphics[width=\textwidth]{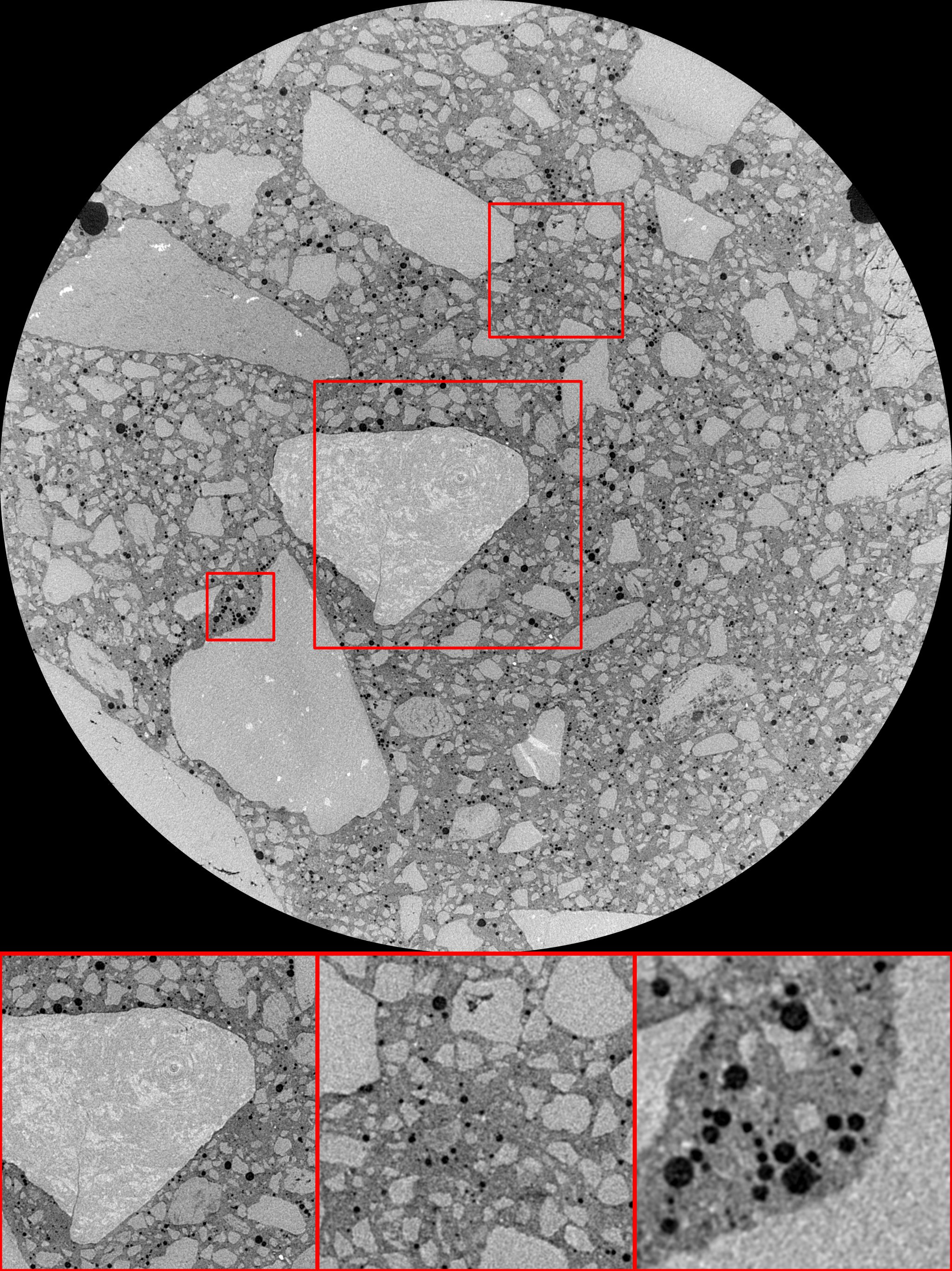}
        \caption*{Real Sample Slice.}
    \end{subfigure}
    \hfill
    \begin{subfigure}[b]{0.48\linewidth}
        \includegraphics[width=\textwidth]{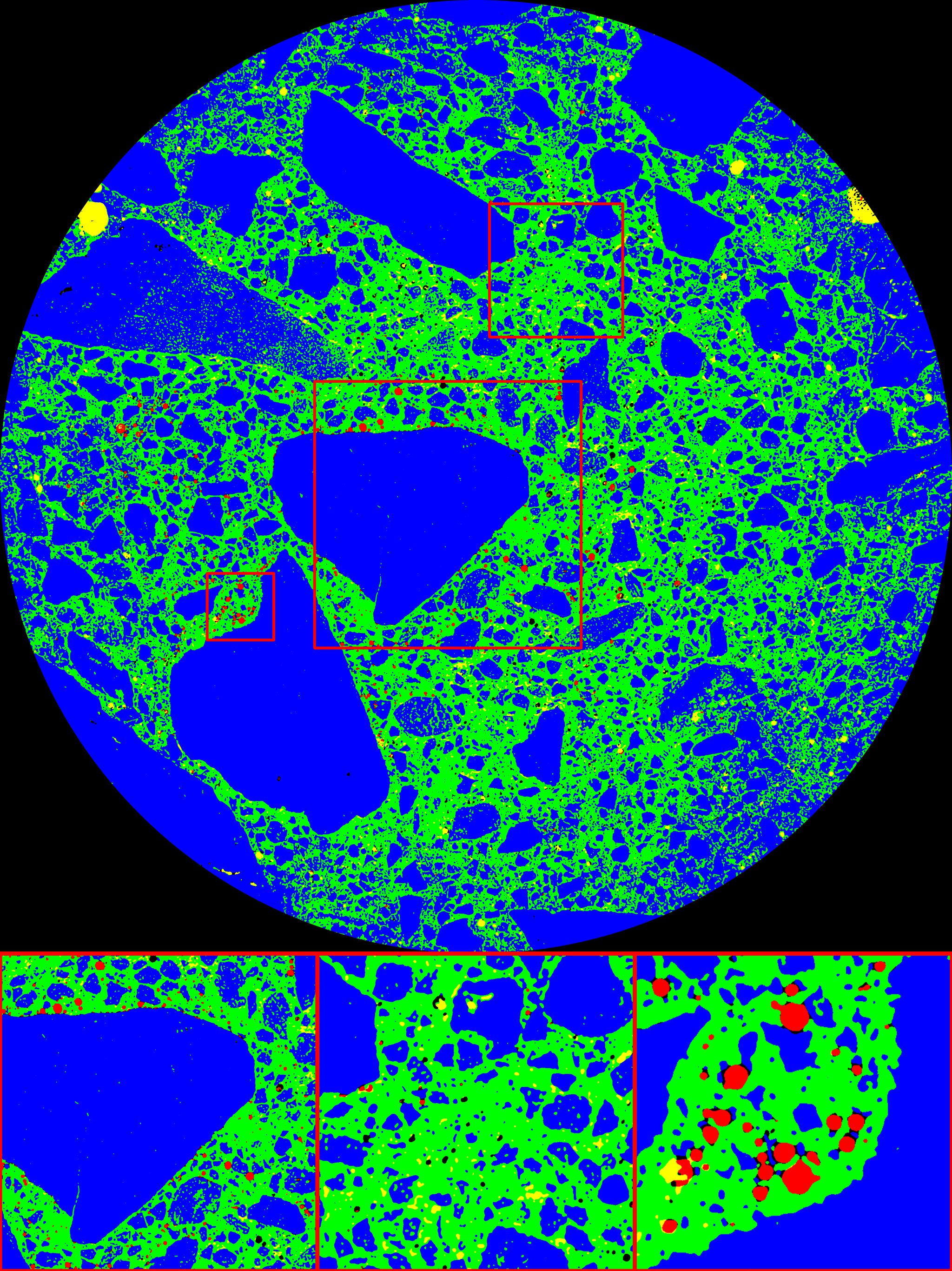}
        \caption*{Zero-shot Prediction}
    \end{subfigure}
    \caption{Segmentation on a real SpringXCT-8K sample with zero-shot inference. The pixel-level microstructure (e.g.\ void area) can be precisely extracted, which is hard for human experts.}
    \label{fig:seg_real}
  \end{center}
\end{figure}

\begin{figure}[h]
  \begin{center}
    \includegraphics[clip,width=0.95\linewidth]{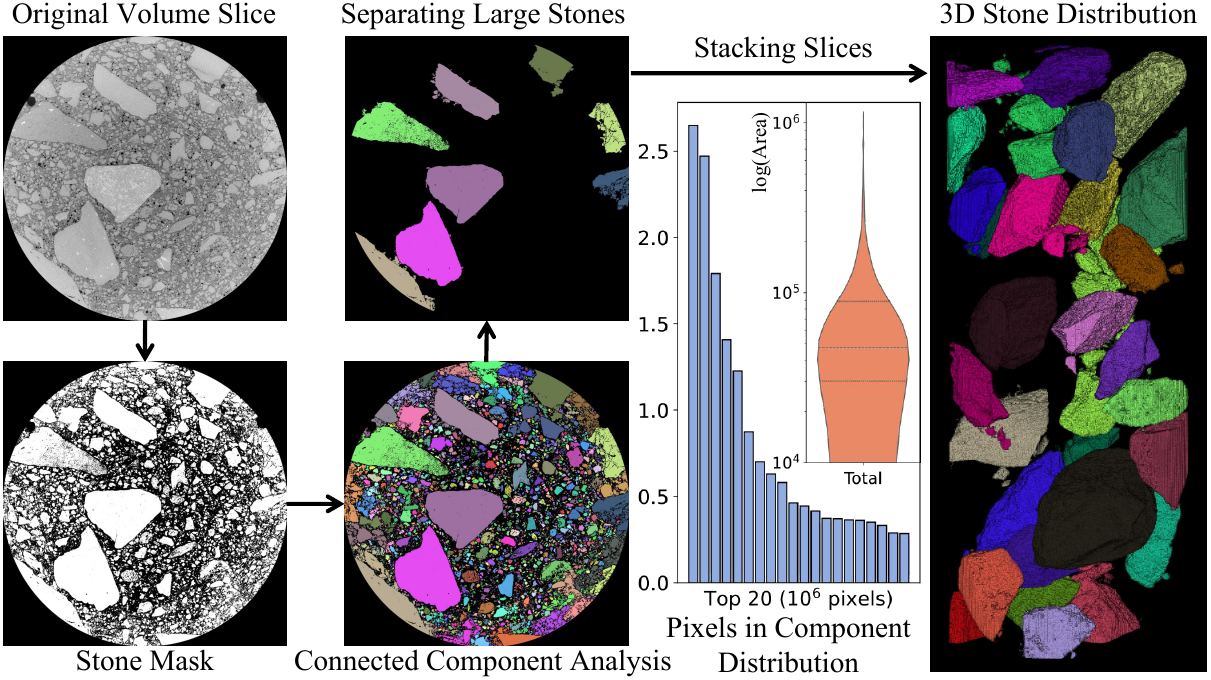}
    \caption{Example of SpringXCT-8K downstream task: connected component analysis~\cite{zhang2022topologypreservingsegmentationnetworkdeep}, separation, and mapping the stones distribution in the 3D image.}
    \label{fig:downstream-s8d}
  \end{center}
\end{figure}

\begin{figure}[h]
  \begin{center}
    \includegraphics[clip,width=0.95\linewidth]{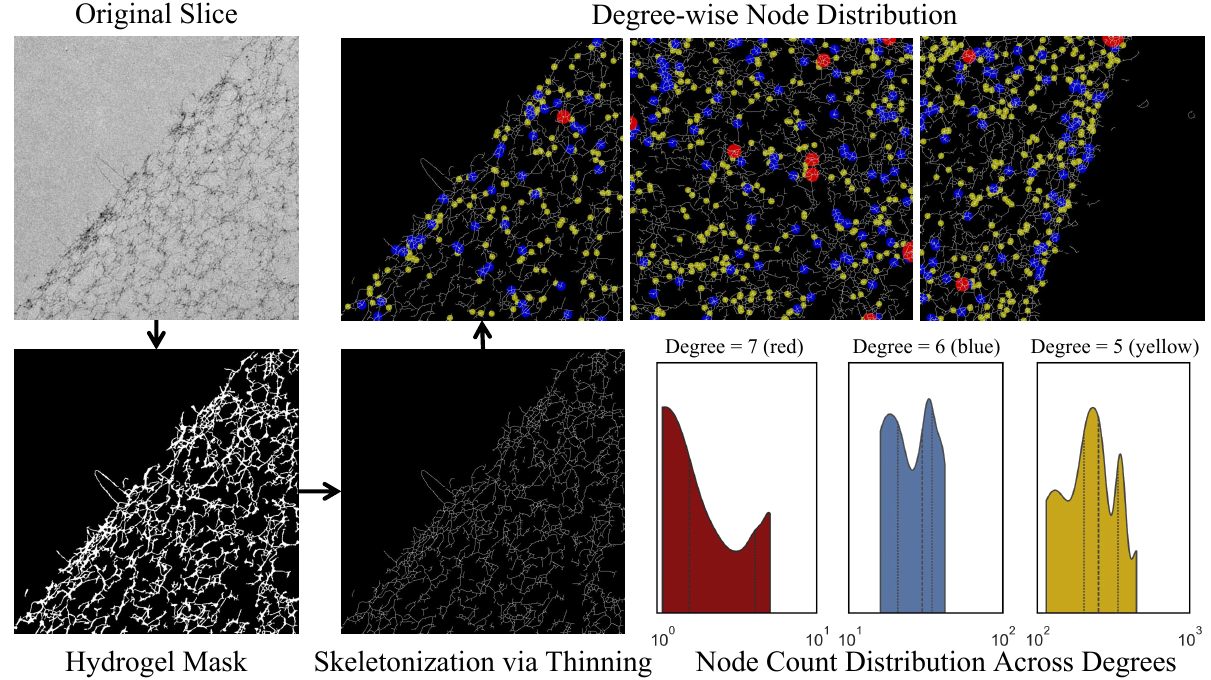}
    \caption{Example of HydrogelTEM-1K downstream task: a hydrogel sample processed through zero-shot segmentation and skeletonization, followed by junction detection and degree-wise node annotation, illustrating the structural complexity of the hydrogel network.}
    \label{fig:hydrogel-simulate}
  \end{center}
\end{figure}

As shown in Figure~\ref{fig:seg_real} and Figure~\ref{fig:downstream-s8d}, the components derived from our SGMA-UNETR segmentations yield a near-perfect segmentation on real samples. In contrast, the statistics derived from the MAE-SAM~2 masks are noisy and unreliable, systematically underestimating the volume of small pores that its large-patch pre-training failed to capture. For HydrogelTEM-1K in Figure~\ref{fig:hydrogel-simulate}, we also performed zero-shot segmentation and extracted the skeleton for degree analysis. These examples demonstrate that \method{} is not just a marginal improvement in segmentation quality; it is an enabling technology that provides the high-precision results necessary to replace manual, error-prone, and expensive lab-based analysis in core scientific applications.

\subsection{Downstream Task Visualizations}
\label{sec:supp_downstream}

\subsection{Comparison with Long-Sequence and Hierarchical Transformer Methods}
\label{sec:supp_longseq}

Table~\ref{tab:long_sequence_summary} summarizes how \method{} compares against prior long-sequence and hierarchical Transformer methods along four axes: best-case complexity, attention mechanism invariance (i.e.\ whether the approach requires modifying self-attention itself), model-architecture coupling (whether the method is tied to a specific forked backbone), and implementation locus (attention kernel vs.\ model vs.\ input-preprocessing). Most prior work either \emph{(i)} approximates attention with sparsity, low-rank, or locality assumptions, which trades off accuracy for speed and often requires forked PyTorch kernels; or \emph{(ii)} designs hierarchical backbones (CrossViT, HIPT, MEGABYTE) that need multiple trained models and impose specific inductive biases. In contrast, \method{} operates entirely at the \emph{input-preprocessing} stage: it leaves the attention mechanism and ViT architecture untouched, and instead produces a fixed-length structure-conditioned token sequence via the Structure-Guided Feature Tokenizer (SGFT). This makes \method{} a drop-in replacement for MAE pre-training on any standard ViT, yet reduces the effective complexity to $O(\log^2 N)$ in the input resolution $N$ in the best case.

\begingroup
\footnotesize
\setlength{\tabcolsep}{4pt}
\renewcommand{\arraystretch}{1.2}
\begin{xltabular}{\textwidth}{%
    @{}>{\raggedright\arraybackslash}p{1.4cm}%
       >{\raggedright\arraybackslash}p{2.6cm}%
       Y%
       >{\raggedright\arraybackslash}p{1.5cm}%
       >{\raggedright\arraybackslash}p{1.9cm}%
       >{\raggedright\arraybackslash}p{1.6cm}@{}%
}
    \caption{A summary of relevant long-sequence training methods that attack quadratic attention by reducing the work done. Here $N$ = sequence length.}
    \label{tab:long_sequence_summary}\\
    \toprule
    \textbf{Approach} & \textbf{Method} & \textbf{Merits \& Demerits} & \textbf{Complexity (Best)} & \textbf{Model} & \textbf{Implementation} \\
    \midrule
    \endfirsthead

    \multicolumn{6}{r}{\textit{Table~\thetable{} --- continued from previous page}}\\
    \toprule
    \textbf{Approach} & \textbf{Method} & \textbf{Merits \& Demerits} & \textbf{Complexity (Best)} & \textbf{Model} & \textbf{Implementation} \\
    \midrule
    \endhead

    \midrule
    \multicolumn{6}{r}{\textit{continued on next page}}\\
    \endfoot

    \bottomrule
    \endlastfoot

    \multirow{5}{=}{Attention Approximation}
      & Longformer~\cite{beltagy2020longformer}, ETC~\cite{ainslie2020etc}
      & \textbf{(+)} Better time complexity vs.\ Transformer. \newline \textbf{(-)} Sparsity levels insufficient for gains to materialize.
      & $O(N)$\newline$O(N\sqrt{N})$
      & Some models, forked PyTorch & Self-attention kernel \\
    \cmidrule(l){2-6}
      & BigBird~\cite{zaheer2020big}, Reformer~\cite{kitaev2020reformer}
      & \textbf{(+)} Theoretically proven time complexity. \newline \textbf{(-)} High-order derivatives.
      & $O(N\log{N})$ & Some models, forked PyTorch & Self-attention kernel \\
    \cmidrule(l){2-6}
      & Sparse Attention~\cite{child2019generating}
      & \textbf{(+)} Introduced sparse factorizations of attention. \newline \textbf{(-)} Higher time complexity.
      & $O(N\sqrt{N})$ & Some models, forked PyTorch & Self-attention kernel \\
    \cmidrule(l){2-6}
      & Linformer~\cite{katharopoulos2020transformers}, Performer~\cite{choromanski2020rethinking}
      & \textbf{(+)} Fast adaptation. \newline \textbf{(-)} Assumes self-attention is low rank.
      & $O(N)$ & Some models, forked PyTorch & Self-attention kernel \\
    \cmidrule(l){2-6}
      & SPFormer~\cite{mei2024spformer} (Prediction)
      & \textbf{(+)} Irregular tokens. \newline \textbf{(-)} No adaptation to high resolution.
      & $O(P^2)$\newline($P$: \#regions)
      & Model, plain PyTorch & Model Implementation \\
    \midrule

    \multirow{4}{=}{Hierarchical}
      & Hier.\ Transformer~\cite{Si21} (Text Class.)
      & \textbf{(+)} Independent hyperparameter tuning of hierarchical levels. \newline \textbf{(-)} No support for ViT.
      & $O(N\log{N})$ & Model, plain PyTorch & Model Implementation \\
    \cmidrule(l){2-6}
      & CrossViT~\cite{chen2021crossvit} (Class.)
      & \textbf{(+)} Better time complexity vs.\ standard ViT. \newline \textbf{(-)} Complex token fusion in dual-branch ViTs.
      & $O(N)$ & Model, plain PyTorch & Model Implementation \\
    \cmidrule(l){2-6}
      & HIPT~\cite{Chen22} (Class.)
      & \textbf{(+)} Models inductive biases of features in the hierarchy. \newline \textbf{(-)} High cost for training multiple models.
      & $O(N\log{N})$ & Model, plain PyTorch & Model Implementation \\
    \cmidrule(l){2-6}
      & MEGABYTE~\cite{yu2023megabyte} (Prediction)
      & \textbf{(+)} Supports multi-modality. \newline \textbf{(-)} High cost for training multiple models.
      & $O(N^{4/3})$ & Model, plain PyTorch & Model Implementation \\
    \midrule

    State-Space Model
      & U-Mamba~\cite{ma2024u} (Segm.)
      & \textbf{(+)} Linear-time long-range dependency via selective SSM. \newline \textbf{(-)} Replaces attention entirely; not compatible with MAE / standard ViT pre-training.
      & $O(N)$ & Model, plain PyTorch & Model Implementation \\
    \midrule

    \multirow{3}{=}{High-resolution}
      & Swin-UNETR~\cite{cao2022swin} (Segm.)
      & \textbf{(+)} Shifted windows preserve locality at scale. \newline \textbf{(-)} Window partitioning causes inter-layer load imbalance; fixed window size limits adaptivity.
      & $O(N)$ & Model, plain PyTorch & Model Implementation \\
    \cmidrule(l){2-6}
      & Hiera~\cite{ryali2023hiera} (Segm.\ / Class.)
      & \textbf{(+)} Simple hierarchical ViT; strong downstream accuracy. \newline \textbf{(-)} Requires a bespoke backbone; no adaptive tokenization on giga-pixel inputs.
      & $O(N)$ & Model, plain PyTorch & Model Implementation \\
    \cmidrule(l){2-6}
      & QuadTree Attn.~\cite{tang2022quadtree} (Detect.\ / Match.)
      & \textbf{(+)} Coarse-to-fine attention via quadtree; reduces attention cost. \newline \textbf{(-)} Modifies the attention computation itself; incompatible with standard MAE.
      & $O(N)$ & Model, forked PyTorch & Self-attention kernel \\
    \midrule

    \textbf{Ours}
      & \textbf{Structure-Guided Masked AutoEncoder (Segm.\ \& Class.)}
      & \textbf{(+) Attention mechanism intact.} \newline \textbf{(+) Largely reduces computation cost; maintains quality.} \newline \textbf{(+) Efficiency scales with image detail.} \newline \textbf{(-) Task semantics independent of edge information.}
      & $\boldsymbol{O(\log^2 N)}$
      & \textbf{Any model, plain PyTorch} & \textbf{Image pre-processing} \\
\end{xltabular}
\endgroup

\subsection{SGFT Tokenizer Construction Efficiency}
\label{sec:supp_hat_construction}

A natural concern with any adaptive tokenization scheme is that the tokenizer itself becomes a new bottleneck: prior adaptive-patching methods such as Adaptive Patching~\cite{zhang2024adaptive} and SHF~\cite{zhang2025shf} construct the quadtree by a strictly sequential $\arg\max$ greedy split over the active token set, whose cost grows roughly linearly with the target sequence length $N$. Because SGFT is invoked once per sample at every epoch, an $O(N)$ tokenizer would erase the $O(\log^2 N)$ attention savings on giga-pixel images.

We therefore deploy SGFT's softmax-batched token selection (Section~\ref{sec:hierarchical_patching}, Eq.~\ref{eq:softmax_sampling}) as the production path, and benchmark it against two stricter $\arg\max$ baselines: the sequential greedy split used in AP~\cite{zhang2024adaptive} / SHF~\cite{zhang2025shf} (``\textbf{Naive (SHF/AP)}''), and a heap-accelerated variant of the same rule (``\textbf{Heap (intermediate)}'') that we developed as an intermediate optimization. All three produce a quadtree of $N$ tokens; we additionally report the pixel-level MSE between the quadtree-reconstructed image (each token rendered at its mean color) and the original as a tokenization-quality measure --- lower MSE means the tokens spend their budget on more informative regions.

\begin{table}[h]
    \centering
    \caption{SGFT tokenizer construction efficiency. Three quadtree-construction algorithms producing the same number of tokens $N$ at three image resolutions. \emph{Naive (SHF/AP)} is the sequential $\arg\max$ greedy split used in prior adaptive-patching work~\cite{zhang2025shf}; \emph{Heap (intermediate)} is our heap-accelerated variant of the same rule; \emph{SGFT (softmax batch, ours)} is the production tokenizer used throughout the paper. The Naive baseline is run only at the $1{,}024^2$ scale and reported as ``--'' at higher resolutions, where its sequential cost becomes prohibitive. MSE between the three algorithms is within $\sim$3\% on the small bench (SGFT occasionally lower because the softmax injects a small amount of exploration), so the speedup comes essentially for free on tokenization quality.}
    \label{tab:hat_construction}
    \resizebox{\linewidth}{!}{
    \begin{tabular}{l|r|r|r|r|r|r}
        \toprule
        \thead{\textbf{Image Res.}} & \thead{\textbf{Leaves $N$}} &
        \thead{\textbf{Naive (SHF/AP)}\\\cite{zhang2024adaptive,zhang2025shf}} &
        \thead{\textbf{Heap (intermediate)}\\(ours)} &
        \thead{\textbf{SGFT}\\(ours)} &
        \thead{\textbf{Speedup}\\\textbf{vs.\ Naive}} & \thead{\textbf{Speedup}\\\textbf{vs.\ Heap}} \\
        \midrule
        \multirow{3}{*}{$1{,}024^2$}
            & 4{,}096   & 38.8 ms  & 32.3 ms  & \textbf{12.1 ms}  & \textbf{3.2$\times$} & \textbf{2.7$\times$} \\
            & 8{,}192   & 74.0 ms  & 52.3 ms  & \textbf{13.1 ms}  & \textbf{5.7$\times$} & \textbf{4.0$\times$} \\
            & 16{,}384  & 176.9 ms & 97.1 ms  & \textbf{18.6 ms}  & \textbf{9.5$\times$} & \textbf{5.2$\times$} \\
        \midrule
        \multirow{3}{*}{$4{,}096^2$}
            & 65{,}536      & --       & 1.85 s   & \textbf{0.33 s}   & --                    & \textbf{5.6$\times$} \\
            & 262{,}144     & --       & 6.97 s   & \textbf{0.63 s}   & --                    & \textbf{11$\times$}  \\
            & 1{,}024{,}000 & --       & 26.81 s  & \textbf{1.31 s}   & --                    & \textbf{20$\times$}  \\
        \midrule
        \multirow{3}{*}{$8{,}192^2$}
            & 65{,}536      & --       & 2.41 s   & \textbf{1.01 s}   & --                    & \textbf{2.4$\times$} \\
            & 262{,}144     & --       & 7.51 s   & \textbf{1.39 s}   & --                    & \textbf{5.4$\times$} \\
            & 1{,}024{,}000 & --       & 27.49 s  & \textbf{2.98 s}   & --                    & \textbf{9.2$\times$} \\
        \bottomrule
    \end{tabular}
    }
\end{table}

Two observations are worth highlighting. \emph{First}, the tokenization-quality gap is negligible: across the small-scale runs the per-pixel MSE between the three algorithms agrees to within $\sim$3\% (and SGFT is sometimes \emph{lower} than the deterministic argmax variants because the softmax sampling occasionally routes tokens into regions the greedy rule misses), so the speedup comes essentially for free on quality. \emph{Second}, the speedup widens with $N$: on a $4{,}096^2$ image at $N=10^6$ tokens the heap-based $\arg\max$ takes $\sim$27\,s per image while SGFT finishes in $\sim$1.3\,s, a $20\times$ gap; on $8{,}192^2$ the same configuration is $9.2\times$ faster. In other words, SGFT is not only cheap relative to attention --- it also consistently beats the SHF/AP-style adaptive tokenizer in raw wall-clock construction time, which is what allows the per-sample preprocessing cost of \method{} to stay well under a second on the giga-pixel datasets used in the main experiments.

\subsection{Additional 3D Medical Segmentation Baselines}
\label{sec:supp_extra_baselines}

Although ultra-high-resolution 2D imaging is the primary focus of the paper, the long-sequence problem that motivates \method{} also arises in \emph{3D volumetric data} at moderate spatial resolution: a $512^3$ CT volume with $4^3$ patches still produces a token sequence of length $N=2{,}097{,}152$, which is well beyond what standard self-attention can handle. We therefore additionally evaluate \method{}-SAM~2 on two standard 3D medical segmentation benchmarks, BTCV~\cite{landman2015miccai} and KiTS19~\cite{Heller2019kits19}, to test whether the SGFT-based sequence compression remains effective in the 3D setting. We compare against widely used 3D medical baselines (U-Net, U-Mamba, TransUNet, UNETR, Swin UNETR, Swin UNETR-V2, CoTr, nnFormer) and the SAM~2 backbone with the same fine-tuning recipe. We omit a separate MAE-SAM~2 row here because the MAE-vs-\method{} pre-training comparison is already established at length on matched architectures in Section~\ref{sec:seg_performance} (Table~\ref{tab:combined_results}); the goal of Table~\ref{tab:btcv_kits} is to position \method{}-SAM~2 against the broader 3D medical landscape.

The takeaway from Table~\ref{tab:btcv_kits} is that, even at this lower spatial resolution where the segmentation accuracy of strong specialized 3D baselines is already saturated, \method{}-SAM~2 with $2^3$ patches matches or marginally beats the best baseline on Dice (\textbf{89.71\%} vs.\ Swin UNETR's $89.5\%$ on BTCV; \textbf{87.25\%} vs.\ UNETR's $86.45\%$ on KiTS19), while delivering a substantial speedup driven by SGFT's sequence compression (\textbf{7.15$\times$} faster than SAM~2 on BTCV; \textbf{2.58$\times$} faster on KiTS19). In other words, on standard-resolution 3D data the headline contribution of \method{} is \emph{efficiency at parity accuracy}, complementing the much larger accuracy gains we observe on the ultra-high-resolution 2D datasets in the main text.

\begin{table}[t]
    \centering
    \caption{Segmentation of BTCV~\cite{landman2015miccai} for multi-organ segmentation and KiTS19~\cite{Heller2019kits19} for Kidney Tumor Segmentation on a single GPU. \emph{Time} indicates the end-to-end runtime to achieve the corresponding Dice Score.}
    \label{tab:btcv_kits}
    \resizebox{\linewidth}{!}{
	\begin{tabular}{c|l|c|r|c|c}
		\toprule
        \thead{\textbf{Dataset}} &
		\thead{\textbf{Model}} & \thead{\textbf{Patch Size}} & \thead{\textbf{Time}}  & \thead{\textbf{Speedup ($\times$)}} & \thead{\textbf{Dice Score (\%)}} \\
            \midrule
            \multirow{7}{*}{\shortstack{BTCV~\cite{landman2015miccai}}}
            & U-Net~\cite{ronneberger2015u} & N/A & 843.90 Seconds    & 9.04$\times$ &  80.2  \\
            \cline{2-6}
            & U-Mamba~\cite{ma2024u} & N/A & 8,016.24 Seconds  & 0.95$\times$ & 83.51 \\
            \cline{2-6}
            & TransUNet~\cite{chen2021transunet} & N/A & 3115.25 Seconds  & 2.45$\times$ & 83.8 \\
            \cline{2-6}
            & UNETR~\cite{hatamizadeh2022unetr} &  $4^3$ & 8386.56 Seconds  & 0.91$\times$ &  89.1  \\
            \cline{2-6}
            & Swin UNETR~\cite{DBLP:conf/cvpr/TangY0RLXNH22} & $4^3$ & 5861.93 Seconds  & 1.30$\times$ &  89.5  \\
            \cline{2-6}
            & SAM~2~\cite{ravi2024sam2} & $4^3$ & 7637.28 Seconds  & 1.0$\times$ &  82.77  \\
            \cline{2-6}
            & \textbf{SGMA-SAM~2} & $\mathbf{2^3}$ & \textbf{1067.88 Seconds}  & \textbf{7.15$\times$} &  \textbf{89.71}  \\
            \midrule
            \multirow{9}{*}{\shortstack{KiTS~\cite{Heller2019kits19}}}
            & U-Net~\cite{ronneberger2015u} & N/A & 243.7 Minutes    & 1.99$\times$ &  83.23  \\
            \cline{2-6}
            & U-Mamba~\cite{ma2024u} & N/A & 969.0 Minutes  & 0.5$\times$ & 86.22 \\
            \cline{2-6}
            & CoTr~\cite{xie2021cotr} & N/A & 488.7 Minutes  & 0.99$\times$ & 84.59 \\
            \cline{2-6}
            & UNETR~\cite{hatamizadeh2022unetr} &  $8^3$ & 513.6 Minutes  & 0.94$\times$ &  86.45  \\
            \cline{2-6}
            & nnFormer~\cite{zhou2021nnformer} & $8^3$ & 876.5 Minutes  & 0.55$\times$ &  75.85  \\
            \cline{2-6}
            & Swin UNETR~\cite{DBLP:conf/cvpr/TangY0RLXNH22} & $8^3$ & 748.3 Minutes  & 0.65$\times$ &  81.27  \\
            \cline{2-6}
            & Swin UNETR-V2~\cite{he2023swinunetrv2} & $8^3$ & 766.4 Minutes  & 0.63$\times$ &  84.14  \\
            \cline{2-6}
            & SAM~2~\cite{ravi2024sam2} & $8^3$ & 483.1 Minutes  & 1.0$\times$ &  81.35  \\
            \cline{2-6}
            & \textbf{SGMA-SAM~2} & $\mathbf{2^3}$ & \textbf{187.3 Minutes}  & \textbf{2.58$\times$} &  \textbf{87.25}  \\
	    \bottomrule
	\end{tabular}
    }
\end{table}

\subsection{Additional Task: Classification on PAIP-16K and a DA Ablation}
\label{sec:supp_cls_ablation}

To test whether \method{} transfers beyond segmentation, we additionally evaluate it on a whole-slide classification task on PAIP downsampled to $16{,}384^2$ (``PAIP-16K''). We compare against (i) vanilla ViT~\cite{dosovitskiy2021image}, (ii) HIPT~\cite{Chen22} --- a hierarchical giga-pixel ViT baseline --- and (iii) SHF-ViT~\cite{zhang2025shf}, which uses the same symmetric hierarchical forest tokenizer as our \method{}-SAM head but without SGMA pre-training. For each architecture we report two patch-size settings: \emph{coarse} ($4{,}096^2$) and \emph{fine} ($2^2$); GPU counts match what the model can fit at that setting. On the classification-only baselines (ViT / HIPT / SHF-ViT) we leave the Dice column empty (``--'') because they do not produce a segmentation head under this protocol. For \method{} we also report a \textbf{-NoDA} ablation in which the Damped Accumulation (DA) signal path is removed, so the difference between \textbf{SGMA-ViT-2-NoDA} and \textbf{SGMA-ViT-2} isolates the contribution of DA on top of SGFT tokenization.

As shown in Table~\ref{tab:classification_paip16k}, \method{}-ViT-2 achieves the best Top-1 accuracy (\textbf{83.77\%}) and the best Dice (\textbf{81.67\%}), improving over SHF-ViT-2 by $+3.63$ accuracy points and over HIPT by $+11.08$ points. More revealingly, the NoDA ablation shows a clear \emph{task-dependent} pattern: removing DA costs only $+0.88$ Top-1 (within run-to-run noise for classification) but $+3.00$ Dice on segmentation. In other words, \textbf{DA contributes essentially nothing to classification and a substantial margin to segmentation}, which is consistent with its design --- DA injects multi-scale spatial structure along the quadtree path, and this kind of hierarchical spatial signal is valuable for dense pixel-level prediction but largely redundant when the task only needs a single global category label. SGFT alone already matches or beats SHF-ViT-2 on both tasks, and DA is precisely the component that lifts the segmentation half.

\begin{table}[htbp]
    \centering
    \caption{Classification (Top-1 accuracy, \%) and Segmentation (Dice, \%) of vanilla ViT, HIPT~\cite{Chen22}, SHF-ViT~\cite{zhang2025shf}, and \method{}-ViT on PAIP-16K ($16{,}384^2$ res.). ``--'' denotes classification-only baselines. ``-NoDA'' ablates the Damped Accumulation signal path to isolate the contribution of DA on top of SGFT.}
    \label{tab:classification_paip16k}
    \resizebox{\linewidth}{!}{
    \begin{tabular}{l|c|c|c|c}
        \toprule
        \thead{\textbf{Model}} & \thead{\textbf{GPUs}} & \thead{\textbf{Patch Size}} & \thead{\textbf{Accuracy (\%)}} & \thead{\textbf{Dice (\%)}} \\
        \midrule
        ViT~\cite{dosovitskiy2021image}  & 128 & $4{,}096^2$                 & 68.97 & -- \\
        HIPT~\cite{Chen22}               & 128 & $\{16{,}256^2,\, 4{,}096^2\}$ & 72.69 & -- \\
        \midrule
        SHF-ViT-4096~\cite{zhang2025shf} & 8   & $4{,}096^2$ & 69.11 & -- \\
        SHF-ViT-2~\cite{zhang2025shf}    & 128 & $2^2$       & 80.14 & -- \\
        \midrule
        \method{}-ViT-4096               & 8   & $4{,}096^2$ & 70.37 & 72.83 \\
        \method{}-ViT-2-NoDA (SGFT only)             & 128 & $2^2$       & 82.89 & 78.67 \\
        \method{}-ViT-2                  & 128 & $2^2$       & \textbf{83.77} & \textbf{81.67} \\
        \bottomrule
    \end{tabular}
    }
\end{table}

\end{document}